\documentclass[manuscript,screen,nonacm]{acmart}
\usepackage{graphicx}
\usepackage{multirow}
\usepackage{amsmath}
\usepackage{amsfonts}
\usepackage{amsthm}
\usepackage{mathrsfs}
\usepackage[title]{appendix}
\usepackage{xcolor}
\usepackage{textcomp}
\usepackage{manyfoot}
\usepackage{booktabs}
\usepackage{algorithm}
\usepackage{algorithmicx}
\usepackage{algpseudocode}
\usepackage{listings}
\usepackage{multirow}
\usepackage{graphicx}  
\usepackage{lmodern}
\usepackage{tabularray}
\usepackage{longtable}
\usepackage{censor}
\usepackage{pifont}
\usepackage{booktabs}
\usepackage{float}
\usepackage{xcolor}
\usepackage{array}
\usepackage{lscape}
\usepackage{url}
\usepackage{hyperref}
\usepackage{adjustbox}
\hypersetup{breaklinks=true}
\usepackage{blindtext}
\usepackage{xpatch}

\AtBeginDocument{
  }

\renewcommand\footnotetextcopyrightpermission[1]{}

\begin{document}

\title{A Comprehensive Survey of Deep Research: Systems, Methodologies, and Applications}

\author{Renjun Xu}
\authornote{Corresponding author: rux@zju.edu.cn}
\author{Jingwen Peng}
\affiliation{
  \institution{Zhejiang University}
  \country{China}
}

\renewcommand{\shortauthors}{Xu et al.}

\begin{abstract}
This survey examines the rapidly evolving field of Deep Research systems---AI-powered applications that automate complex research workflows through the integration of large language models, advanced information retrieval, and autonomous reasoning capabilities. We analyze more than 80 commercial and non-commercial implementations that have emerged since 2023, including \path{OpenAI/Deep Research}, \path{Gemini/Deep Research}, \path{Perplexity/Deep Research}, and numerous open-source alternatives. Through comprehensive examination, we propose a novel hierarchical taxonomy that categorizes systems according to four fundamental technical dimensions: foundation models and reasoning engines, tool utilization and environmental interaction, task planning and execution control, and knowledge synthesis and output generation. We explore the architectural patterns, implementation approaches, and domain-specific adaptations that characterize these systems across academic, scientific, business, and educational applications. Our analysis reveals both the significant capabilities of current implementations and the technical and ethical challenges they present regarding information accuracy, privacy, intellectual property, and accessibility. The survey concludes by identifying promising research directions in advanced reasoning architectures, multimodal integration, domain specialization, human-AI collaboration, and ecosystem standardization that will likely shape the future evolution of this transformative technology. By providing a comprehensive framework for understanding Deep Research systems, this survey contributes to both the theoretical understanding of AI-augmented knowledge work and the practical development of more capable, responsible, and accessible research technologies. The paper resources can be viewed at \href{https://github.com/scienceaix/deepresearch}{https://github.com/scienceaix/deepresearch}.
\end{abstract}



\maketitle

\clearpage
\tableofcontents
\clearpage

\section{Introduction}

Rapid advancement of artificial intelligence has precipitated a paradigm shift in how knowledge is discovered, validated, and utilized across academic and industrial domains. Traditional research methodologies, reliant on manual literature reviews, experimental design, and data analysis, are increasingly supplemented—and in some cases supplanted—by intelligent systems capable of automating end-to-end research workflows. This evolution has given rise to a novel domain we term ``Deep Research'', which signifies the convergence of large language models (LLMs), advanced information retrieval systems, and automated reasoning frameworks to redefine the boundaries of scholarly inquiry and practical problem-solving.

\subsection{Definition and Scope of Deep Research}\label{sec:definition_scope}
Deep Research refers to the systematic application of AI technologies to automate and enhance research processes through three core dimensions:

\begin{enumerate}
    \item \textbf{Intelligent Knowledge Discovery:} Automating literature search, hypothesis generation, and pattern recognition across heterogeneous data sources
    \item \textbf{End-to-End Workflow Automation:} Integrating experimental design, data collection, analysis, and result interpretation into unified AI-driven pipelines
    \item \textbf{Collaborative Intelligence Enhancement:} Facilitating human-AI collaboration through natural language interfaces, visualizations, and dynamic knowledge representation
\end{enumerate}

To clearly delineate the boundaries of Deep Research, we distinguish it from adjacent AI systems as follows:

\begin{itemize}
    \item \textbf{Differentiating from General AI Assistants:} While general AI assistants like ChatGPT can answer research questions, they lack the autonomous workflow capabilities, specialized research tools, and end-to-end research orchestration that define Deep Research systems. Recent surveys have highlighted this crucial distinction between specialized research systems and general AI capabilities~\cite{Transforming_Science_with_Large_Language_Models, A_Survey_on_RAG_Meeting_LLMs}, with particular emphasis on how domain-specific tools fundamentally transform research workflows compared to general-purpose assistants~\cite{zenil2023futurefundamentalscienceled, pournaras2023scienceerachatgptlarge}.

    \item \textbf{Differentiating from Single-Function Research Tools:} Specialized tools like citation managers, literature search engines, or statistical analysis packages address isolated research functions but lack the integrated reasoning and cross-functional orchestration of Deep Research systems. Tools like \path{scispace}~\cite{scispace} and \path{You.com}~\cite{you_com} represent earlier attempts at research assistance but lack the end-to-end capabilities that define true Deep Research systems.

    \item \textbf{Differentiating from Pure LLM Applications:} Applications that simply wrap LLMs with research-oriented prompts lack the environmental interaction, tool integration, and workflow automation capabilities that characterize true Deep Research systems.
\end{itemize}

This survey specifically examines systems that exhibit at least two of the three core dimensions, with a focus on those incorporating large language models as their foundational reasoning engine. Our scope encompasses commercial offerings such as \path{OpenAI/Deep Research} \cite{openai2025}, Google's \path{Gemini/Deep Research} \cite{google2024}, and \path{Perplexity/Deep Research} \cite{perplexity2025}, alongside open-source implementations including \path{dzhng/deep-research} \cite{dzhng2024}, \path{HKUDS/Auto-Deep-Research} \cite{hkuds2024}, and numerous others detailed in subsequent sections. We exclude purely bibliometric tools or single-stage automation systems lacking integrated cognitive capabilities, such as research assistance tools like \path{Elicit} \cite{elicit}, \path{ResearchRabbit} \cite{researchrabbit}, \path{Consensus} \cite{Consensus}, or citation tools like \path{Scite} \cite{scite}. Additional specialized tools like \path{STORM} \cite{STORM}, which focuses on scientific text retrieval and organization, are valuable but lack the end-to-end deep research capabilities central to our survey scope.

\subsection{Historical Context and Technical Evolution}
The trajectory of Deep Research can be mapped through three evolutionary stages that reflect both technological advancements and implementation approaches:

\subsubsection{Origin and Early Exploration (2023 - February 2025)} 
It should be noted that workflow automation frameworks like \path{n8n} \cite{n8n2024}, \path{QwenLM/Qwen-Agent} \cite{qwen2025}, etc. had already been in existence long before the boom of deep research. Their early establishment demonstrated the pre-existing groundwork in related technological domains, highlighting that the development landscape was not solely shaped by the emergence of deep research, but had a more diverse and earlier-rooted origin. 
The concept of Deep Research emerged from the shift of AI assistants towards intelligent agents. In December 2024, Google Gemini pioneered this functionality with its initial Deep Research implementation, focusing on basic multi-step reasoning and knowledge integration \cite{deep_research_now_available_gemini}. This phase laid the groundwork for subsequent advancements, setting the stage for more sophisticated AI-driven research tools. Many of these advances built upon earlier workflow automation tools like \path{n8n} \cite{n8n2024} and agent frameworks such as \path{AutoGPT} \cite{autogpt} and \path{BabyAGI} \cite{babyagi} that had already established foundations for autonomous task execution. 
Other early contributions to this ecosystem include \path{cline2024} \cite{cline2024}, which pioneered integrated research workflows, and \path{open_operator} \cite{open_operator}, which developed foundational browser automation capabilities essential for web-based research.

\subsubsection{Technological Breakthrough and Competitive Rivalry (February - March 2025)}
The rise of DeepSeek's open-source models \cite{deepseek} revolutionized the market with efficient reasoning and cost-effective solutions. In February 2025, OpenAI's release of Deep Research, marked a significant leap forward \cite{openai2025}. Powered by the o3 model, it demonstrated advanced capabilities such as autonomous research planning, cross-domain analysis, and high-quality report generation, achieving accuracy rates exceeding previous benchmarks in complex tasks. Concurrently, Perplexity launched its free-to-use Deep Research in February 2025 \cite{perplexity2025}, emphasizing rapid response and accessibility to capture the mass market. Open-source projects such as \path{nickscamara/open-deep-research} \cite{nickscamara2024}, \path{mshumer/OpenDeepResearcher} \cite{mshumer2024}, \path{btahir_open_deep_research} \cite{btahir_open_deep_research}, and \path{GPT-researcher} \cite{gpt_researcher} emerged as community-driven alternatives to commercial platforms. 
The ecosystem continued to expand with lightweight implementations like \path{Automated-AI-Web-Researcher-Ollama} \cite{Automated-AI-Web-Researcher-Ollama}, designed for local execution with limited resources, and modular frameworks such as \path{Langchain-AI/Open_deep_research} \cite{langchain2024} that provided composable components for custom research workflows.

\subsubsection{Ecosystem Expansion and Multi-modal Integration (March 2025 - Present)}
The third stage is characterized by the maturation of a diverse ecosystem. Open-source projects like \path{Jina-AI/node-DeepResearch} \cite{jina2025} enable localized deployment and customization, while commercial closed-source versions from OpenAI and Google continue to push boundaries with multi-modal support and multi-agent collaboration capabilities. The integration of advanced search technologies and report generation frameworks further enhances the tool's utility across academic research, financial analysis, and other fields. Meanwhile, platforms like \path{Manus} \cite{manus2025} and \path{AutoGLM-Research} \cite{autoglm_research2025}, \path{MGX} \cite{mgx2025}, and \path{Devin} \cite{devin2025} are incorporating advanced AI research capabilities to enhance their services. Concurrently, Anthropic launched \path{Claude/Research} \cite{ClaudeResearch} in April 2025, introducing agentic search capabilities that systematically explore multiple angles of queries and deliver comprehensive answers with verifiable citations. Agent frameworks such as \path{OpenManus} \cite{openmanus2025}, \path{Camel-AI/OWL} \cite{camel2025}, and \path{TARS} \cite{tars2025} further expand the ecosystem with specialized capabilities and domain-specific optimizations.

\begin{figure}[ht]
    \centering
    \includegraphics[width=1.0\linewidth]{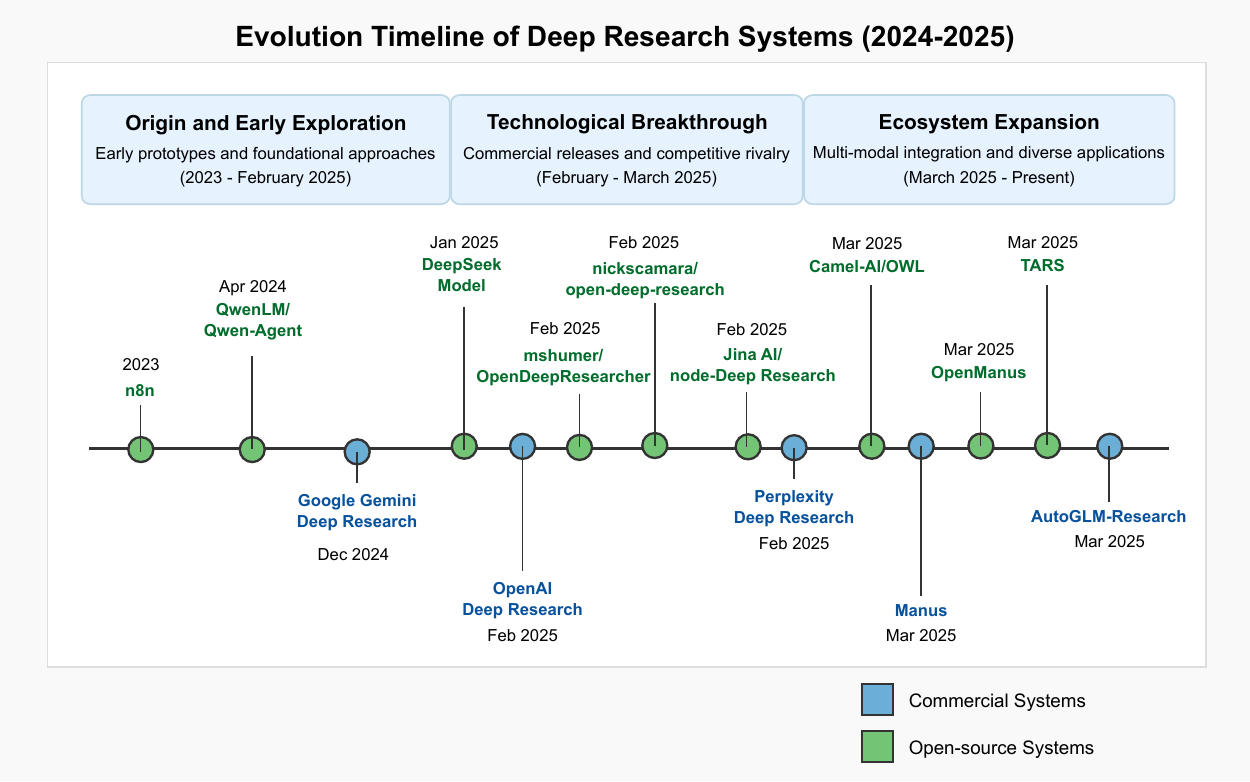}
    \caption{Evolution Timeline of Deep Research Systems}
    \label{fig:evolution}
\end{figure}

\subsection{Significance and Practical Implications}
Deep Research demonstrates transformative potential across multiple domains:
\begin{enumerate}
    \item \textbf{Academic Innovation:} Accelerating hypothesis validation through automated literature synthesis (e.g., HotpotQA \cite{hotpotQA} performance benchmarks) and enabling researchers to explore broader interdisciplinary connections that might otherwise remain undiscovered. 
    The transformative potential of Deep Research extends beyond individual applications to fundamentally reshape scientific discovery processes. As Sourati and Evans \cite{sourati2023acceleratingsciencehumanawareartificial} argue, human-aware artificial intelligence can significantly accelerate science by augmenting researchers' capabilities while adapting to their conceptual frameworks and methodological approaches. This human-AI synergy represents a fundamental shift from traditional automation toward collaborative intelligence that respects and enhances human scientific intuition. Complementary work by Khalili and Bouchachia \cite{khalili2022buildingsciencediscoverymachines} further demonstrates how systematic approaches to building science discovery machines can transform hypothesis generation, experimental design, and theory refinement through integrated AI-driven research workflows.
    
    \item \textbf{Enterprise Transformation:} Enabling data-driven decision-making at scale through systems like \path{Agent-RL/ReSearch} \cite{agentrl2024} and \path{smolagents/open_deep_research} \cite{smolagents2024} that can analyze market trends, competitive landscapes, and strategic opportunities with unprecedented depth and efficiency.
    
    \item \textbf{Democratization of Knowledge:} Reducing barriers to entry through open-source implementations like \path{grapeot/deep_research_agent} \cite{grapeot2024} and \path{OpenManus} \cite{openmanus2025}, making sophisticated research capabilities accessible to individuals and organizations regardless of technical expertise or resource constraints.
\end{enumerate}

\subsection{Research Questions and Contribution of this Survey}
This survey addresses three fundamental questions:
\begin{enumerate}
    \item How do architectural choices (system architecture, implementation approach, functional capabilities) impact Deep Research effectiveness?
    
    \item What technical innovations have emerged in LLM fine-tuning, retrieval mechanisms, and workflow orchestration across the spectrum of Deep Research implementations?
    
    \item How do existing systems balance performance, usability, and ethical considerations, and what patterns emerge from comparing approaches like those of \path{n8n} \cite{n8n2024} and \path{OpenAI/AgentsSDK} \cite{openaiagents2025}?
\end{enumerate}

Our contributions manifest in three dimensions:

\begin{enumerate}
    \item \textbf{Methodological:} Proposing a novel taxonomy categorizing systems by their technical architecture, from foundation models to knowledge synthesis capabilities
    
    \item \textbf{Analytical:} Conducting comparative analysis of representative systems across evaluation metrics, highlighting the strengths and limitations of different approaches
    
    \item \textbf{Practical:} Identifying key challenges and formulating a roadmap for future development, with specific attention to emerging architectures and integration opportunities
\end{enumerate}

The remainder of this paper follows a structured exploration beginning with conceptual frameworks (Section 2), technical innovations and comparative analysis (Sections 3-4), implementation technologies (Section 5), evaluation methodologies (Section 6), applications and use cases (Section 7), ethical considerations (Section 8), and future directions (Section 9).

\section{The Evolution and Technical Framework of Deep Research}

This section presents a comprehensive technical taxonomy for understanding Deep Research systems, organized around four fundamental technological capabilities that define these systems. For each capability, we examine the evolutionary trajectory and technical innovations while highlighting representative implementations that exemplify each approach.

\begin{figure}[ht]
    \centering
    \includegraphics[width=1.0\linewidth]{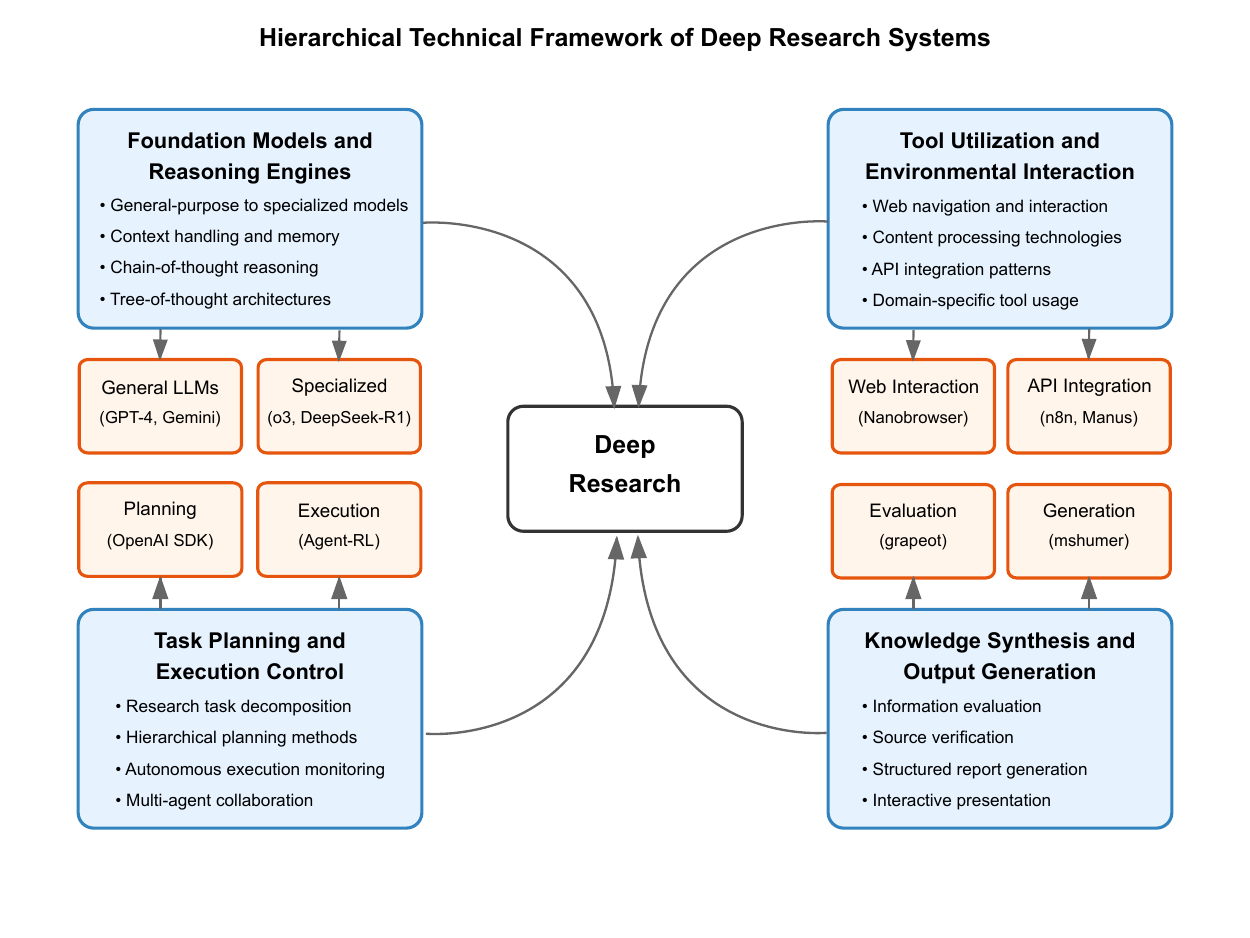}
    \caption{Hierarchical Technical Framework of Deep Research Systems}
    \label{fig:framework}
\end{figure}

\subsection{Foundation Models and Reasoning Engines: Evolution and Advances}

The foundation of Deep Research systems lies in their underlying AI models and reasoning capabilities, which have evolved from general-purpose language models to specialized research-oriented architectures.

\subsubsection{From General-Purpose LLMs to Specialized Research Models}

The progression from general LLMs to research-specialized models represents a fundamental shift in deep research capabilities:

\paragraph{Technical Evolution Trajectory}
Early implementations relied on general-purpose LLMs with minimal task-specific optimization. Current systems feature models specifically enhanced for research tasks through architectural modifications, specialized training corpora, and fine-tuning regimes focused on analytical and reasoning capabilities. The transition from models like GPT-4 to OpenAI's o3 demonstrates significant improvements in abstraction, multi-step reasoning, and knowledge integration capabilities essential for complex research tasks \cite{introducing_OpenAI_o3_and_o4mini, o3ando4mini}.

\paragraph{Representative Systems}
\path{OpenAI/Deep Research} \cite{openai2025} exemplifies this evolution with its o3-based model optimized specifically for web browsing and data analysis. The system leverages chain-of-thought and tree-of-thought reasoning techniques to navigate complex information landscapes. Google's \path{Gemini/Deep Research} \cite{deep_research_now_available_gemini} similarly employs Gemini 2.5 Pro with enhanced reasoning capabilities and a million-token context window to process extensive information. These approaches build upon foundational work in reasoning enhancement techniques like chain-of-thought prompting~\cite{wei2023chainofthoughtpromptingelicitsreasoning}, self-consistency~\cite{wang2023selfconsistencyimproveschainthought}, and human preference alignment~\cite{ouyang2022traininglanguagemodelsfollow} that have been adapted specifically for research-intensive tasks.
In the open-source domain, \path{AutoGLM-Research} \cite{autoglm_research2025} demonstrates how specialized training regimes can optimize existing models like ChatGLM for research-intensive tasks, achieving significant performance gains through targeted enhancements to reasoning components.

\subsubsection{Context Understanding and Memory Mechanisms}

The ability to process, retain, and utilize extensive contextual information represents a crucial advancement in Deep Research systems:

\paragraph{Technical Evolution Trajectory}
Early systems struggled with limited context windows, hampering their ability to synthesize information from multiple sources. Contemporary implementations employ sophisticated memory management techniques including episodic buffers, hierarchical compression, and attention-based retrieval mechanisms that extend effective context far beyond model limitations. The million-token context windows of models like Grok 3 \cite{Grok_3_Beta} and Gemini 2.5 Pro \cite{deep_research_now_available_gemini}, along with the context optimization in OpenAI's o3 model \cite{compare_models_OpenAI_API}, have dramatically expanded the information processing capabilities of these systems.
Advanced systems now distinguish between working memory (active reasoning context) and long-term memory (knowledge repository), allowing for more human-like research processes.

\paragraph{Representative Systems}
\path{Perplexity/Deep Research} \cite{perplexity2025} has pioneered efficient context processing by leveraging DeepSeek-R1's capabilities while implementing proprietary mechanisms for structured information management. The system can analyze hundreds of sources while maintaining coherent reasoning threads. Similarly, \path{Camel-AI/OWL} \cite{camel2025} employs an innovative open-weight approach to memory management, allowing for dynamic allocation of attention resources based on information relevance and task requirements. Both systems demonstrate how effective memory architectures can significantly enhance research performance even with comparable base model capabilities.

\subsubsection{Enhancements in Reasoning Capabilities}

Advanced reasoning mechanisms distinguish modern Deep Research systems from conventional LLM applications:

\paragraph{Technical Evolution Trajectory}
Early implementations relied primarily on zero-shot or few-shot prompting for reasoning tasks. 
Current systems integrate explicit reasoning frameworks including chain-of-thought, tree-of-thought, and graph-based reasoning architectures.  
Recent work by Lang et al. \cite{lang2025debatehelpsweaktostronggeneralization} demonstrates how debate-driven reasoning can facilitate weak-to-strong generalization, enabling more robust performance on complex research tasks through structured argumentative processes. These approaches implement reasoning patterns that more closely mirror human scientific discourse, with explicit representation of alternative viewpoints and structured evaluation of competing hypotheses.
Advanced implementations like OpenAI's o3 incorporate self-critique, uncertainty estimation, and recursive reasoning refinement \cite{o3ando4mini, introducing_OpenAI_o3_and_o4mini}. This evolution enables increasingly sophisticated forms of evidence evaluation, hypothesis testing, and knowledge synthesis essential for high-quality research outputs.

\paragraph{Representative Systems}
\path{QwenLM/Qwen-Agent} \cite{qwen2025} exemplifies advanced reasoning capabilities through its specialized toolkit integration and modular reasoning framework. The system employs a multi-stage reasoning process with explicit planning, information gathering, analysis, and synthesis phases optimized for research workflows. Similar capabilities are evident in \path{smolagents/open_deep_research} \cite{smolagents2024}, which implements a flexible reasoning architecture that can adapt to different research domains and methodologies. Systems like CycleResearcher~\cite{weng2025cycleresearcherimprovingautomatedresearch} demonstrate how integrating automated review processes into research workflows can enhance accuracy through structured feedback loops. These approaches implement explicit verification steps that identify potential errors and inconsistencies before generating final research outputs. The application of AI to complex domains like mathematics further illustrates this progress, where models are increasingly viewed from a cognitive science perspective to enhance their reasoning abilities~\cite{zhang2023aimathematicscognitivescience}, achieving notable milestones such as silver-medal standards in solving International Mathematical Olympiad problems~\cite{AI_achieves_silver-medal_standard}. These systems highlight how reasoning enhancements can dramatically improve research quality even without requiring the largest or most computationally intensive base models.

\subsection{Tool Utilization and Environmental Interaction: Evolution and Advances}

Deep Research systems must effectively interact with external environments to gather and process information, representing a fundamental capability beyond core language model functions\cite{TORL}.

\subsubsection{Web Interaction Technology Development}

The ability to navigate and extract information from the web represents a foundational capability for deep research:

\paragraph{Technical Evolution Trajectory}
Initial implementations relied on simple API-based search queries with limited interaction capabilities. Current systems employ sophisticated web navigation including dynamic content handling, authentication management, and interactive element manipulation. Advanced implementations feature semantic understanding of web structures, allowing for adaptive information extraction and multi-page navigation flows. This evolution has dramatically expanded access to web-based information sources and the ability to extract insights from complex web environments.

\paragraph{Representative Systems}
\path{Nanobrowser} \cite{nanobrowser2024} represents a purpose-built browser environment designed specifically for AI agent use, offering optimized rendering and interaction capabilities for research tasks. It enables fine-grained control of web navigation while maintaining security and performance. Similarly, \path{AutoGLM} \cite{autoglm_research2025} demonstrates sophisticated GUI interaction capabilities across both web and mobile interfaces, allowing it to access information through interfaces designed for human use. These systems showcase how specialized web interaction technologies can significantly expand the information gathering capabilities of Deep Research systems.

\subsubsection{Content Processing Technology Advancements}

Beyond basic navigation, the ability to process diverse content formats is crucial for comprehensive research:

\paragraph{Technical Evolution Trajectory}
Early systems were limited primarily to text extraction from HTML sources. 
Modern implementations support multi-modal content processing including structured data tables, embedded visualizations, PDF documents, and interactive applications. Advanced systems like those built on OpenAI's o3 can extract semantic structure from unstructured content, identify key information from diverse formats, and integrate insights across modalities \cite{thinking_with_images}. This evolution has dramatically expanded the range of information sources that can be incorporated into research processes.

\paragraph{Representative Systems}
The \path{dzhng/deep-research} \cite{dzhng2024} project exemplifies advanced content processing through its specialized modules for different document types and formats. It implements custom extraction logic for academic papers, technical documentation, and structured data sources. Similarly, \path{nickscamara/open-deep-research} \cite{nickscamara2024} features sophisticated content normalization pipelines that transform diverse formats into consistent knowledge representations suitable for analysis. Both systems demonstrate how specialized content processing can significantly enhance the quality and comprehensiveness of research outputs.

\subsubsection{Specialized Tool Integration Progress}

Integration with domain-specific tools extends Deep Research capabilities beyond general information processing:

\paragraph{Technical Evolution Trajectory}
Initial systems relied on general-purpose web search and basic API integrations. The integration of diverse tools has been dramatically advanced by frameworks like ToolLLM \cite{qin2023toolllmfacilitatinglargelanguage}, which enables large language models to master over 16,000 real-world APIs, significantly expanding the interaction capabilities of research systems. Similarly, AssistGPT \cite{gao2023assistgptgeneralmultimodalassistant} demonstrates how general multi-modal assistants can plan, execute, inspect, and learn across diverse environments, creating unified research experiences that seamlessly incorporate varied information sources and interaction modalities. LLaVA-Plus \cite{liu2023llavapluslearningusetools} further extends these capabilities through explicit tool learning mechanisms, enabling research assistants to adaptively incorporate specialized tools within multimodal workflows. 
Current implementations feature complex toolchains including specialized databases, analytical frameworks, and domain-specific services. 
Advanced systems dynamically select and orchestrate tools based on research requirements, effectively composing custom research workflows from available capabilities. Some implementations like those leveraging OpenAI's \path{Codex} \cite{openai_codex} can even generate custom code to process research data or implement analytical models on demand, further extending analytical capabilities. This evolution has enabled increasingly sophisticated analysis and domain-specific research applications.

\paragraph{Representative Systems}
\path{Manus} \cite{manus2025} exemplifies sophisticated tool orchestration through its extensive API integration framework and tool selection mechanisms. The system can incorporate domain-specific research tools and services into unified workflows, significantly expanding its analytical capabilities. Similarly, \path{n8n} \cite{n8n2024} provides a flexible workflow automation platform that can be configured for research tasks, allowing for integration with specialized data sources and analytical services. Steward extends web interaction capabilities by implementing natural language-driven navigation and operation across websites, overcoming scalability limitations of traditional automation frameworks while maintaining low operational costs \cite{tang2024stewardnaturallanguageweb}. These systems highlight how tool integration can extend Deep Research capabilities into specialized domains and complex analytical workflows.

\subsection{Task Planning and Execution Control: Evolution and Advances}

Effective research requires sophisticated planning and execution mechanisms to coordinate complex, multi-stage workflows.

\subsubsection{Research Task Planning Development}

The ability to decompose research objectives into manageable tasks represents a fundamental advancement:

\paragraph{Technical Evolution Trajectory}
Early approaches employed simple task decomposition with linear execution flows, similar to those found in early agent frameworks like \path{MetaGPT} \cite{metagpt} and \path{AgentGPT} \cite{agentGPT}. Modern systems implement hierarchical planning with dynamic refinement based on intermediate results and discoveries. Advanced planning approaches increasingly incorporate structured exploration methodologies to navigate complex solution spaces efficiently. AIDE~\cite{jiang2025aideaidrivenexplorationspace} demonstrates how tree search algorithms can effectively explore the space of potential code solutions for machine learning engineering, trading computational resources for enhanced performance through strategic reuse and refinement of promising pathways. 
Advanced implementations incorporate resource-aware planning, considering time constraints, computational limitations, and information availability. However, incorporating AI tools for tasks like automated code review has been observed to increase pull request closure durations despite benefits, as evidenced in studies such as Cihan et al.~\cite{cihan2024automatedcodereviewpractice}, highlighting the critical need to account for temporal impacts in such resource-aware systems. This evolution has enabled increasingly sophisticated research strategies adaptive to both task requirements and available resources.

\paragraph{Representative Systems}
The \path{OpenAI/AgentsSDK} \cite{openaiagents2025} provides a comprehensive framework for research task planning, with explicit support for goal decomposition, execution tracking, and adaptive refinement. It enables the development of applications with sophisticated planning capabilities for research workflows. Similarly, \path{Flowith/OracleMode} \cite{flowith2025} implements specialized planning mechanisms optimized for research tasks, with particular emphasis on information quality assessment and source prioritization. These systems demonstrate how advanced planning capabilities can significantly improve research efficiency and effectiveness.

\subsubsection{Autonomous Execution and Monitoring Advances}

Reliable execution of research plans requires sophisticated control and monitoring mechanisms:

\paragraph{Technical Evolution Trajectory}
Initial systems employed basic sequential execution with limited error handling. Current implementations feature concurrent execution paths, comprehensive monitoring, and dynamic response to execution challenges. Advanced systems implement self-supervision with explicit success criteria, failure detection, and autonomous recovery strategies. This evolution has dramatically improved the reliability and autonomy of Deep Research systems across complex tasks.

\paragraph{Representative Systems}
\path{Agent-RL/ReSearch} \cite{agentrl2024} exemplifies advanced execution control through its reinforcement learning-based approach to research execution. The system learns effective execution strategies from experience, continuously improving its ability to navigate complex research workflows. Its adaptive execution mechanisms can recover from failures and adjust strategies based on intermediate results, highlighting how sophisticated control mechanisms can enhance research reliability and effectiveness.

\subsubsection{Multi-Agent Collaboration Framework Development}

Complex research often benefits from specialized agent roles and collaborative approaches:

\paragraph{Technical Evolution Trajectory}
Early systems relied on monolithic agents with undifferentiated capabilities. Modern implementations employ specialized agent roles with explicit coordination mechanisms and information sharing protocols. Advanced systems feature dynamic role allocation, consensus-building mechanisms, and sophisticated conflict resolution strategies. This evolution has enabled increasingly complex collaborative research workflows and improved performance on challenging tasks\cite{From_Persona_to_Personalization}. For instance, frameworks employing multi-agent debate have been shown to improve evaluation consistency~\cite{chan2023chatevalbetterllmbasedevaluators}, while research into generative AI voting demonstrates resilience to model biases in collective decision-making~\cite{majumdar2025generativeaivotingfair}.

\paragraph{Representative Systems}
The \path{smolagents/open_deep_research} \cite{smolagents2024} framework demonstrates effective multi-agent collaboration through its modular agent architecture and explicit coordination mechanisms. It enables the composition of specialized research teams with complementary capabilities and shared objectives. Similarly, \path{TARS} \cite{tars2025} implements a sophisticated agent collaboration framework within its desktop environment, allowing multiple specialized agents to contribute to unified research workflows. These systems highlight how multi-agent approaches can enhance research capabilities through specialization and collaboration.

\subsection{Knowledge Synthesis and Output Generation: Evolution and Advances}

The ultimate value of Deep Research systems lies in their ability to synthesize disparate information into coherent, actionable insights.

\subsubsection{Information Evaluation Technology Development}

Critical assessment of information quality represents a crucial capability for reliable research:

\paragraph{Technical Evolution Trajectory}
Early systems relied primarily on source reputation heuristics with limited content-based assessment. Modern implementations employ sophisticated evaluation frameworks considering source characteristics, content features, and consistency with established knowledge. Advanced systems implement explicit uncertainty modeling, contradiction detection, and evidential reasoning approaches. This evolution has dramatically improved the reliability and trustworthiness of research outputs. Advances in knowledge retrieval based on generative AI enhance the ability to source and verify information~\cite{yang2025knowledgeretrievalbasedgenerative}.

\paragraph{Representative Systems}
The \path{grapeot/deep_research_agent} \cite{grapeot2024} implements sophisticated information evaluation mechanisms with explicit quality scoring for diverse source types. It can assess information reliability based on both intrinsic content features and extrinsic source characteristics, enabling more discerning information utilization. These capabilities highlight how advanced evaluation mechanisms can significantly enhance research quality and reliability.

\subsubsection{Report Generation Technology Advances}

Effective communication of research findings requires sophisticated content organization and presentation:

\paragraph{Technical Evolution Trajectory}
Initial systems produced simple text summaries with limited structure or coherence. Current implementations generate comprehensive reports with hierarchical organization, evidence integration, and coherent argumentation. Advanced systems produce adaptive outputs tailored to audience expertise, information needs, and presentation contexts. This evolution has dramatically improved the usability and impact of Deep Research outputs.

\paragraph{Representative Systems}
The \path{mshumer/OpenDeepResearcher} \cite{mshumer2024} project exemplifies advanced report generation through its structured output framework and evidence integration mechanisms. It produces comprehensive research reports with explicit attribution, structured arguments, and integrated supporting evidence. These capabilities demonstrate how sophisticated report generation can enhance the utility and trustworthiness of Deep Research outputs. Additionally, the MegaWika dataset \cite{barham2023megawikamillionsreportssources} offers a large-scale multilingual resource consisting of millions of articles and referenced sources, enabling collaborative AI report generation.

\subsubsection{Interactive Presentation Technology Development}

Beyond static reports, interactive result exploration enhances insight discovery and utilization:

\paragraph{Technical Evolution Trajectory}
Early systems produced fixed textual outputs with minimal user interaction. Modern implementations support dynamic exploration including drill-down capabilities, source verification, and alternative viewpoint examination. Advanced systems enable collaborative refinement through iterative feedback incorporation and adaptive response to user queries. This evolution has dramatically enhanced the utility and flexibility of Deep Research interfaces.

\paragraph{Representative Systems}
\path{HKUDS/Auto-Deep-Research} \cite{hkuds2024} implements sophisticated interactive presentation capabilities, allowing users to explore research findings through dynamic interfaces, examine supporting evidence, and refine analysis through iterative interaction. These features highlight how interactive presentation technologies can enhance the utility and accessibility of Deep Research outputs, facilitating more effective knowledge transfer and utilization.

This technical framework provides a comprehensive foundation for understanding the capabilities and evolution of Deep Research systems. The subsequent sections will build on this framework to analyze implementation approaches, evaluate system performance, and explore applications across diverse domains.

\section{Comparative Analysis and Evaluation of Deep Research Systems}

Building upon the technical framework established in Section 2, this section provides a comprehensive comparative analysis of existing Deep Research systems across multiple dimensions. We examine how different implementations balance technical capabilities, application suitability, and performance characteristics to address diverse research needs.

\subsection{Cross-Dimensional Technical Comparison}

Deep Research systems demonstrate varying strengths across the four key technical dimensions identified in our framework. This section analyzes how different implementations balance these capabilities and the resulting performance implications.

\subsubsection{Foundation Model and Reasoning Efficiency Comparison}

The underlying reasoning capabilities of Deep Research systems significantly impact their overall effectiveness:

\begin{table}[ht]
    \centering
    \caption{Comparison of Foundation Model Characteristics}
    \label{tab:foundation-models}
    \begin{tabular}{lcccc}
        \toprule
        \textbf{System} & \textbf{Base Model} & \textbf{Context Length} & \textbf{Reasoning Approach} \\
        \midrule
        \path{OpenAI/Deep Research} \cite{openai2025} & o3 & may up to 200k tokens \cite{compare_models_OpenAI_API} & Multi-step reasoning \\
        \path{Gemini/Deep Research} \cite{deep_research_now_available_gemini} & Gemini 2.5 Pro & 1M tokens \cite{openai_vs_gemini} & Chain-of-thought \\
        \path{Perplexity/Deep Research} \cite{perplexity2025} & DeepSeek-R1 & 128K tokens \cite{sonar_perplexity_research_models} & Iterative reasoning  \\
        \path{Grok 3 Beta} \cite{Grok_3_Beta} & Grok 3 & 1M tokens \cite{Grok_3_Beta} & Chain-of-thought \\ 
        \path{AutoGLM-Research} \cite{autoglm_research2025} & ChatGLM & DOM & Step-by-step planning \\
        \bottomrule
    \small{DOM: Depends On the Model}
    \end{tabular}
\end{table}

Commercial systems from OpenAI and Google leverage proprietary models with extensive context windows and sophisticated reasoning mechanisms, enabling them to process larger volumes of information with greater coherence. OpenAI's o3 model demonstrates particular strength in complex reasoning tasks, while Gemini 2.5 Pro excels in information integration across diverse sources. In contrast, \path{Perplexity/Deep Research} achieves competitive performance with the open-source DeepSeek-R1 model through optimized implementation and focused use cases.

Open-source implementations like \path{Camel-AI/OWL} \cite{camel2025} and \path{QwenLM/Qwen-Agent} \cite{qwen2025} demonstrate that effective deep research capabilities can be achieved with more accessible models through specialized optimization. The open-weight approach of \path{Camel-AI/OWL} \cite{camel2025} enables flexible deployment across computing environments, while \path{QwenLM/Qwen-Agent} \cite{qwen2025} leverages modular reasoning to compensate for more limited base model capabilities.

\subsubsection{Tool Integration and Environmental Adaptability Comparison}

The ability to interact with diverse information environments varies significantly across implementations:

\begin{table}[ht]
    \centering
    \caption{Environmental Interaction Capabilities of Deep Research Systems}
    \label{tab:tool-integration}
    \begin{adjustbox}{width=\textwidth}
    \begin{tabular}{lccccc}
        \toprule
        \textbf{System} & \textbf{Web Interaction} & \textbf{API Integration} & \textbf{Document Processing} & \textbf{GUI Navigation} \\
        \midrule
        \path{Nanobrowser} \cite{nanobrowser2024} & Headless browsing, JavaScript execution, dynamic content rendering & REST API connectors & Basic HTML parsing & Not implemented \\
        \path{AutoGLM} \cite{autoglm_research2025} & Full browser automation, form interaction & RESTful and GraphQL support & PDF, Office formats, JSON & Element identification, click/input automation \\
        \path{dzhng/deep-research} \cite{dzhng2024} & Multi-page navigation, cookie handling & OAuth authentication support & Academic paper extraction, table parsing & Not implemented \\
        \path{Manus} \cite{manus2025} & JavaScript rendering, session management & 150+ service integrations, webhook support & PDF with layout preservation, CSV processing & Basic element interaction \\
        \path{n8n} \cite{n8n2024} & Limited, via HTTP requests & 200+ integration nodes, custom webhook endpoints & CSV/XML processing & Not implemented \\
        \path{TARS} \cite{tars2025} & Viewport management, scroll handling & REST/SOAP support & Standard formats processing & Desktop application control, UI element recognition \\
        \bottomrule
    \end{tabular}
    \end{adjustbox}
    \scriptsize
    \textit{Note: Capabilities documented based on system repositories, technical documentation, and published demonstrations as of April 2025.}
\end{table}


Specialized tools like \path{Nanobrowser} \cite{nanobrowser2024} excel in web interaction capabilities, providing sophisticated navigation and content extraction optimized for research workflows. Systems like \path{dzhng/deep-research} \cite{dzhng2024} and \path{nickscamara/open-deep-research} \cite{nickscamara2024} complement these capabilities with advanced document processing features that can extract structured information from diverse formats.

Comprehensive platforms like \path{Manus} \cite{manus2025} and \path{AutoGLM} \cite{autoglm_research2025} offer broader environmental interaction capabilities, balancing web browsing, API integration, and document processing. These systems can adapt to diverse research scenarios but may not match the specialized performance of more focused tools in specific domains. The workflow automation capabilities of \path{n8n} \cite{n8n2024} provide exceptional flexibility for API integration but offer more limited direct interaction with web and document environments.

\subsubsection{Task Planning and Execution Stability Comparison}

Effective research requires reliable task planning and execution capabilities:

\begin{table}[ht]
    \centering
    \caption{Planning and Execution Capabilities of Deep Research Systems}
    \label{tab:planning-execution}
    \begin{adjustbox}{width=\textwidth}
    \begin{tabular}{lccc}
        \toprule
        \textbf{System} & \textbf{Task Planning Mechanisms} & \textbf{Error Handling Features} & \textbf{Collaboration Infrastructure} \\
        \midrule
        \path{OpenAI/AgentsSDK} \cite{openaiagents2025} & Hierarchical task decomposition, goal-oriented planning & Automated retry logic, exception handling & Supervisor-worker architecture \\
        \path{Flowith/OracleMode} \cite{flowith2025} & Constraint-based planning, information quality prioritization & Checkpoint-based recovery & Limited role-based workflow \\
        \path{Agent-RL/ReSearch} \cite{agentrl2024} & Reinforcement learning planning, adaptive task ordering & Progressive fallback strategies, state restoration & Standard agent messaging protocol \\
        \path{smolagents/open_deep_research} \cite{smolagents2024} & Task queue management, priority-based scheduling & Basic retry mechanisms & Multi-agent configuration, specialized role definitions \\
        \path{TARS} \cite{tars2025} & Process template architecture, event-driven coordination & State persistence, interruption handling & Team-based agent organization, shared memory \\
        \path{grapeot/deep_research_agent} \cite{grapeot2024} & Linear task execution, sequential processing & Timeout handling & Single-agent architecture \\
        \bottomrule
    \end{tabular}
    \end{adjustbox}
    \scriptsize
    \textit{Note: Capabilities documented based on system repositories, technical documentation, and published implementations as of April 2025.}
\end{table}


The \path{OpenAI/AgentsSDK} \cite{openaiagents2025} demonstrates sophisticated planning capabilities with hierarchical task decomposition and adaptive execution, enabling complex research workflows with reliable completion rates. Similarly, \path{Flowith/OracleMode} \cite{flowith2025} offers advanced planning mechanisms optimized for research tasks, though with more limited error recovery capabilities.

\path{Agent-RL/ReSearch} \cite{agentrl2024} employs reinforcement learning techniques to develop robust execution strategies, enabling exceptional error recovery capabilities that can adapt to unexpected challenges during research workflows. In contrast, \path{smolagents/open_deep_research} \cite{smolagents2024} and \path{TARS} \cite{tars2025} focus on multi-agent collaboration, distributing complex tasks across specialized agents to enhance overall research effectiveness.

Simpler implementations like \path{grapeot/deep_research_agent} \cite{grapeot2024} offer more limited planning and execution capabilities but may provide sufficient reliability for less complex research tasks, demonstrating the range of complexity available across the ecosystem.

\subsubsection{Knowledge Synthesis and Output Quality Comparison}

The ability to synthesize findings into coherent, reliable outputs varies significantly:

\begin{table}[ht]
    \centering
    \caption{Knowledge Synthesis Capabilities of Deep Research Systems}
    \label{tab:knowledge-synthesis}
    \begin{adjustbox}{width=\textwidth}
    \begin{tabular}{lccc}
        \toprule
        \textbf{System} & \textbf{Source Evaluation Mechanisms} & \textbf{Output Structuring} & \textbf{User Interaction Features} \\
        \midrule
        \path{OpenAI/Deep Research} \cite{openai2025} & Source corroboration, authority ranking algorithms & Hierarchical report generation, section organization & Query clarification dialogue, result expansion \\
        \path{Perplexity/Deep Research} \cite{perplexity2025} & Source diversity metrics, publication date filtering & Citation-based organization, inline attribution & Source exploration interface, follow-up questioning \\
        \path{mshumer/OpenDeepResearcher} \cite{mshumer2024} & Publication venue filtering, citation count tracking & Template-based document generation, section templating & Minimal interaction, batch processing focus \\
        \path{HKUDS/Auto-Deep-Research} \cite{hkuds2024} & Basic source categorization, recency filtering & Standard academic format, heading organization & Interactive result exploration, citation navigation \\
        \path{grapeot/deep_research_agent} \cite{grapeot2024} & Evidence classification algorithms, contradictory claim detection & Minimal formatting, raw data presentation & Command-line interface, non-interactive \\
        \path{OpenManus} \cite{openmanus2025} & Source type categorization, basic metadata filtering & Markdown formatting, hierarchy-based organization & Basic query refinement, result browsing \\
        \bottomrule
    \end{tabular}
    \end{adjustbox}
    \scriptsize
    \textit{Note: Capabilities documented based on system repositories, technical documentation, and published implementations as of April 2025.}
\end{table}


Commercial platforms like \path{OpenAI/Deep Research} \cite{openai2025} and \path{Perplexity/Deep Research} \cite{perplexity2025} demonstrate sophisticated information evaluation capabilities, effectively assessing source credibility and content reliability to produce high-quality syntheses. OpenAI's implementation excels in report structure and organization, while Perplexity offers particularly strong citation practices for source attribution and verification.

Open-source implementations like \path{mshumer/OpenDeepResearcher} \cite{mshumer2024} focus on report structure and organization, producing well-formatted outputs that effectively communicate research findings. \path{HKUDS/Auto-Deep-Research} \cite{hkuds2024} emphasizes interactive exploration, allowing users to examine evidence and refine analyses through iterative interaction. Specialized tools like \path{grapeot/deep_research_agent} \cite{grapeot2024} prioritize information evaluation over presentation, focusing on reliable content assessment rather than sophisticated output formatting.

\subsection{Application-Based System Suitability Analysis}

Beyond technical capabilities, Deep Research systems demonstrate varying suitability for different application contexts. This section examines how system characteristics align with key application domains.

\subsubsection{Academic Research Scenario Adaptability Assessment}

Academic research requires particular emphasis on comprehensive literature review, methodological rigor, and citation quality. Systems like \path{OpenAI/Deep Research} \cite{openai2025} excel in this domain through their ability to access academic databases, comprehensively analyze research methodologies, and generate properly formatted citations. Other specialized academic research tools like \path{PaperQA} \cite{paper_qa} and \path{Scite} \cite{scite} offer complementary capabilities focused specifically on scientific literature processing, while Google's \path{NotebookLm} \cite{notebooklm} provides structured knowledge workspaces for academic exploration.

\begin{table}[ht]
    \centering
    \caption{Academic Research Application Features of Deep Research Systems}
    \label{tab:academic-research}
    \begin{adjustbox}{width=\textwidth}
    \begin{tabular}{lccc}
        \toprule
        \textbf{System} & \textbf{Academic Database Integration} & \textbf{Methodology Analysis Features} & \textbf{Citation Management} \\
        \midrule
        \path{OpenAI/Deep Research} \cite{openai2025} & ArXiv, IEEE Xplore, PubMed, Google Scholar & Statistical method identification, study design classification & IEEE, APA, MLA, Chicago format support \\
        \path{Perplexity/Deep Research} \cite{perplexity2025} & ArXiv, PubMed, JSTOR, ACM Digital Library & Experimental design analysis, sample size assessment & Automated citation generation, DOI resolution \\
        \path{dzhng/deep-research} \cite{dzhng2024} & ArXiv, Semantic Scholar, limited database access & Basic methodology extraction & BibTeX export, standard format support \\
        \path{Camel-AI/OWL} \cite{camel2025} & Custom corpus integration, specialized domain databases & Research design pattern recognition, methodology comparison & Domain-specific citation formatting \\
        \path{mshumer/OpenDeepResearcher} \cite{mshumer2024} & Open access databases, PDF repository processing & Methodology summary extraction & Standard citation format generation \\
        \path{HKUDS/Auto-Deep-Research} \cite{hkuds2024} & University library integration, institutional repository access & Research approach categorization & Reference management, bibliography generation \\
        \bottomrule
    \end{tabular}
    \end{adjustbox}
    \scriptsize
    \textit{Note: Features documented based on system repositories, technical documentation, and published use cases as of April 2025.}
\end{table}


\path{OpenAI/Deep Research} \cite{openai2025} demonstrates exceptional suitability for academic research through its comprehensive literature coverage, methodological rigor, and high-quality citation practices. The system can effectively navigate academic databases, understand research methodologies, and produce well-structured literature reviews with appropriate attribution. \path{Perplexity/Deep Research} \cite{perplexity2025} offers similarly strong performance for literature coverage and citation quality, though with somewhat less methodological sophistication.

Open-source alternatives like \path{Camel-AI/OWL} \cite{camel2025} provide competitive capabilities for specific academic domains, 
particular strength in methodological understanding for specific domains. Systems like \path{dzhng/deep-research} \cite{dzhng2024}, \path{mshumer/OpenDeepResearcher} \cite{mshumer2024}, and \path{HKUDS/Auto-Deep-Research} \cite{hkuds2024} offer moderate capabilities across all dimensions, making them suitable for less demanding academic research applications or preliminary literature exploration.

\subsubsection{Enterprise Decision-Making Scenario Adaptability Assessment}

Business intelligence and strategic decision-making emphasize information currency, analytical depth, and actionable insights:

\begin{table}[ht]
    \centering
    \caption{Enterprise Decision-Making Application Features of Deep Research Systems}
    \label{tab:enterprise-decision}
    \begin{adjustbox}{width=\textwidth}
    \begin{tabular}{lccc}
        \toprule
        \textbf{System} & \textbf{Market Information Sources} & \textbf{Analytical Frameworks} & \textbf{Decision Support Features} \\
        \midrule
        \path{Gemini/Deep Research} \cite{deep_research_now_available_gemini} & News API integration, SEC filings access, market data feeds & Competitor analysis templates, trend detection algorithms & Executive summary generation, recommendation formatting \\
        \path{Manus} \cite{manus2025} & Financial data integrations, news aggregation, industry reports & Market sizing frameworks, SWOT analysis templates & Strategic options presentation, decision matrix generation \\
        \path{n8n} \cite{n8n2024} & CRM integration, marketing platform connectivity, custom data sources & Custom analytics workflow creation, data pipeline automation & Dashboard generation, notification systems \\
        \path{Agent-RL/ReSearch} \cite{agentrl2024} & Configurable information source adapters, custom data inputs & Pattern recognition algorithms, causal analysis frameworks & Scenario planning tools, impact assessment matrices \\
        \path{Flowith/OracleMode} \cite{flowith2025} & Real-time data feeds, specialized industry sources & Industry-specific analytical templates, framework application & Strategic briefing generation, insight prioritization \\
        \path{TARS} \cite{tars2025} & Enterprise system integration, desktop application data access & Basic analytical template application & Standardized reporting, data visualization \\
        \bottomrule
    \end{tabular}
    \end{adjustbox}
    \scriptsize
    \textit{Note: Features documented based on system repositories, technical documentation, and published use cases as of April 2025.}
\end{table}


\path{Gemini/Deep Research} \cite{deep_research_now_available_gemini} demonstrates exceptional suitability for enterprise decision-making through its strong information currency, analytical capabilities, and actionable output formats. The system effectively navigates business information sources, analyzes market trends, and produces insights directly relevant to decision processes. \path{Manus} \cite{manus2025} offers similarly strong performance for information acquisition and analysis, though with somewhat less emphasis on actionable recommendation formatting. Microsoft Copilot \cite{copilotmicrosoft} empowers organizations with powerful generative AI, enterprise-grade security and privacy, and is trusted by companies around the world. Similarly, the Adobe Experience Platform AI Assistant \cite{mukherjee2025documentsdialoguebuildingkgrag} employs knowledge graph-enhanced retrieval-augmented generation to accurately respond over private enterprise documents, significantly enhancing response relevance while maintaining provenance tracking.

Workflow automation platforms like \path{n8n} \cite{n8n2024} provide particular strengths in information currency and actionability through their integration with enterprise data sources and business intelligence tools. Research-focused systems like \path{Agent-RL/ReSearch} \cite{agentrl2024} and \path{Flowith/OracleMode} \cite{flowith2025} offer competitive analytical capabilities but may require additional processing to translate findings into actionable business recommendations.

\subsubsection{Personal Knowledge Management Adaptability Assessment}

Individual knowledge management emphasizes accessibility, personalization, and integration with existing workflows:

\begin{table}[ht]
    \centering
    \caption{Personal Knowledge Management Features of Deep Research Systems}
    \label{tab:personal-knowledge}
    \begin{adjustbox}{width=\textwidth}
    \begin{tabular}{lccc}
        \toprule
        \textbf{System} & \textbf{User Interface Design} & \textbf{Customization Options} & \textbf{Existing Tool Integration} \\
        \midrule
        \path{Perplexity/Deep Research} \cite{perplexity2025} & Web-based interface, mobile application support & Topic preference settings, information filtering options & Browser extension, sharing functionality \\
        \path{nickscamara/open-deep-research} \cite{nickscamara2024} & Command-line interface, web interface option & Modular configuration, source priority adjustment & Local file system integration, note-taking exports \\
        \path{OpenManus} \cite{openmanus2025} & Desktop application, local web interface & Template customization, workflow configuration & Note application exports, knowledge base connections \\
        \path{Nanobrowser} \cite{nanobrowser2024} & Programmatic interface, developer-focused API & Full configuration access, component-level customization & Browser automation framework compatibility \\
        \path{smolagents/open_deep_research} \cite{smolagents2024} & Technical interface, Python library integration & Architecture-level customization, agent behavior configuration & Python ecosystem integration, custom adapter support \\
        \path{Jina-AI/node-DeepResearch} \cite{jina2025} & Node.js integration, API-driven interface & Component-level configuration, pipeline customization & Node.js application ecosystem, JavaScript framework support \\
        \bottomrule
    \end{tabular}
    \end{adjustbox}
    \scriptsize
    \textit{Note: Features documented based on system repositories, technical documentation, and published implementations as of April 2025.}
\end{table}


\path{Perplexity/Deep Research} \cite{perplexity2025} offers strong accessibility for personal knowledge management through its consumer-friendly interface and free access tier, though with more limited personalization capabilities. Open-source implementations like \path{nickscamara/open-deep-research} \cite{nickscamara2024} and \path{OpenManus} \cite{openmanus2025} provide greater personalization possibilities through local deployment and customization, enabling adaptation to individual information management preferences.

Infrastructure tools like \path{Nanobrowser} \cite{nanobrowser2024} and \path{Jina-AI/node-DeepResearch} \cite{jina2025} offer particular strengths in workflow integration, allowing seamless incorporation into existing personal knowledge management systems and processes. More complex frameworks like \path{smolagents/open_deep_research} \cite{smolagents2024} provide sophisticated capabilities but may present accessibility challenges for non-technical users.

\subsection{Performance Metrics and Benchmarking}\label{Sec3.3:Performance}

Beyond qualitative comparisons, quantitative performance metrics provide objective assessment of Deep Research capabilities across systems.

\subsubsection{Quantitative Evaluation Metrics}

Standard benchmarks enable comparative evaluation of core research capabilities:

\begin{table}[ht]
    \centering
    \caption{Performance on Standard Evaluation Benchmarks}
    \label{tab:benchmarks}
    \begin{adjustbox}{width=\textwidth}
    \begin{tabular}{lccccc}
        \toprule
        \textbf{System} & \textbf{HLE Score*} \cite{HLE} & \textbf{MMLU** Score} \cite{MMLU} & \textbf{HotpotQA Score} \cite{hotpotQA} & \textbf{GAIA Score(pass@1)*** \cite{GAIA}} \\
        \midrule
        \path{OpenAI/Deep Research} \cite{openai2025} & 26.6\% & - & - & 67.36\% \\
        \path{Gemini-2.5} \cite{deep_research_now_available_gemini,gemini2.5HLE} & 18.8\% & - & - & - \\
        \path{Gemini-2.0-Flash} \cite{google2024,gemini2.0flash} & -  & 77.9\% & - & - \\
        \path{Perplexity/Deep Research} \cite{perplexity2025} & 21.1\% & - & - & - \\
        \path{Grok 3 Beta} \cite{Grok_3_Beta} & - & 79.9\%  & - & - \\
        \path{Manus} \cite{manus2025} & - & - & - & 86.5\% \\
        \path{Agent-RL/ReSearch} \cite{agentrl2024} & - & - & 37.51\% & - \\        
        \bottomrule
    \end{tabular}
    \end{adjustbox}
    \small{*Humanity's Last Exam: Tests frontier research capabilities}\\
    \small{**Massive Multitask Language Understanding: Tests general knowledge} \\
    \small{***GAIA Score(pass@1): Average score}
\end{table}

\begin{table}[ht]
    \centering
    \caption{Documented Performance Metrics from Deep Research Systems}
    \label{tab:documented-performance}
    \begin{adjustbox}{width=\textwidth}
    \begin{tabular}{lcccc}
        \toprule
        \textbf{System} & \textbf{Benchmark} & \textbf{Reported Score} & \textbf{Evaluation Context} & \textbf{Source} \\
        \midrule
        \path{OpenAI/Deep Research} & HLE & 26.6\% & Humanity's Last Exam & \cite{openai2025} \\
        \path{OpenAI/Deep Research} & GAIA (pass@1) & 67.36\% & General AI assistant tasks & \cite{openai2025} \\
        \path{Perplexity/Deep Research} & HLE & 21.1\% & Humanity's Last Exam & \cite{perplexity2025} \\
        \path{Perplexity/Deep Research} & SimpleQA & 93.9\% & Factual question answering & \cite{perplexity2025} \\
        \path{Grok 3 Beta} & MMLU & 92.7\% & Multitask language understanding & \cite{Grok_3_Beta} \\
        \path{Manus} & GAIA (pass@1) & 86.5\% & General AI assistant tasks & \cite{manus2025} \\
        \path{Agent-RL/ReSearch} & HotpotQA & 37.51\% & Multi-hop question answering & \cite{agentrl2024} \\
        \path{AutoGLM} & WebArena-Lite & 55.2\% (59.1\% retry) & Web navigation tasks & \cite{autoglm_research2025} \\
        \path{AutoGLM} & OpenTable & 96.2\% & Restaurant booking tasks & \cite{autoglm_research2025} \\
        \bottomrule
    \end{tabular}
    \end{adjustbox}
    \scriptsize
    \textit{Note: Scores reflect performance on specific benchmarks as reported in cited publications. Direct comparison requires consideration of evaluation methodologies and task specifications.}
\end{table}

\path{OpenAI/Deep Research} \cite{openai2025,openais_deep_research_tool_is_it_useful,openai_vs_perplexity} demonstrates leading performance across various benchmark categories, particularly excelling in Humanity's Last Exam (HLE) \cite{HLE} hich measures advanced research and reasoning capabilities. \path{Gemini/Deep Research} \cite{deep_research_now_available_gemini} shows comparable performance.  According to the introduction of Google Deep Research with Gemini 2.5 Pro Experimental \cite{deep_research_now_available_gemini,gemini_strong_lead_over_openai}, the new model demonstrated superior user preference over \path{OpenAI/Deep Research} across four key metrics: instruction following (60.6\% vs. 39.4\%), Comprehensiveness (76.9\% vs. 23.1\%), Completeness (73.3\% vs. 26.7\%), and Writing quality (58.2\% vs. 41.8\%). These results suggest Gemini 2.5 Pro’s enhanced capability in synthesizing structured, high-fidelity research outputs. This capability is further amplified in fullstack applications, where the integration of Gemini models with frameworks like LangGraph facilitates research-augmented conversational AI for comprehensive query handling, as demonstrated in \path{Google-Gemini/Gemini-Fullstack-Langgraph-Quickstart} \cite{GoogleGeminiFullstack}. \path{Perplexity/Deep Research} \cite{perplexity2025} achieves competitive results despite utilizing the open-source DeepSeek-R1 model, highlighting the importance of implementation quality beyond raw model capabilities.

Open-source implementations show progressively lower benchmark scores, though many still achieve respectable performance suitable for practical applications. Systems like \path{AutoGLM-Research} \cite{autoglm_research2025}, \path{HKUDS/Auto-Deep-Research} \cite{hkuds2024}, and \path{Camel-AI/OWL} \cite{camel2025} demonstrate that effective research capabilities can be achieved with more accessible models and frameworks, though with some performance trade-offs compared to leading commercial implementations.

Recent benchmark development has expanded evaluation to more specialized aspects of research assistance. The AAAR-1.0 benchmark~\cite{lou2025aaar10assessingaispotential} specifically evaluates AI's potential to assist research through 150 multi-domain tasks designed to test both retrieval and reasoning capabilities. Domain-specific approaches include DSBench~\cite{jing2025dsbenchfardatascience}, which evaluates data science agent capabilities across 20 real-world tasks\cite{Mysore_2023,Wang_2022}, SciCode~\cite{tian2024scicoderesearchcodingbenchmark} for scientific code generation, MASSW~\cite{zhang2024masswnewdatasetbenchmark} for scientific workflow assistance, and MMSci~\cite{li2025mmscidatasetgraduatelevelmultidiscipline} for multimodal scientific understanding across graduate-level materials. ScienceQA\cite{lu2022learnexplainmultimodalreasoning} offers a comprehensive multimodal science benchmark with chain-of-thought explanations for evaluating reasoning capabilities. Domain-specific benchmarks like TPBench~\cite{Theoretical_Physics_Benchmark} for theoretical physics and AAAR-1.0~\cite{lou2025aaar10assessingaispotential} for research assistance capabilities offer additional targeted evaluation approaches for specialized research applications. Multi-domain code generation benchmark like DomainCodeBench\cite{zheng2025generalperformancedomain} is designed to systematically assess large language models across 12 software application domains and 15 programming languages. Interactive evaluation frameworks like LatEval \cite{huang2024latevalinteractivellmsevaluation} specifically assess systems' capabilities in handling incomplete information through lateral thinking puzzles, providing insight into research abilities under uncertainty and ambiguity. Complementary approaches like Mask-DPO \cite{gu2025maskdpogeneralizablefinegrainedfactuality} focus on generalizable fine-grained factuality alignment, addressing a critical requirement for reliable research outputs. Domain-specific benchmarks such as GMAI-MMBench \cite{chen2024gmaimmbenchcomprehensivemultimodalevaluation} provide comprehensive multimodal evaluation frameworks specifically designed for medical AI applications, while AutoBench \cite{chen2025autobenchautomatedbenchmarkscientific} offers automated evaluation of scientific discovery capabilities, providing standardized assessment of core research functions.
Other broad evaluation frameworks including HELM~\cite{Holistic_Evaluation_of_Language_Models}, BIG-bench~\cite{big_bench}, and AGIEval~\cite{AGIEval}, provide complementary assessment dimensions. Specialized multimodal benchmarks like INQUIRE \cite{vendrow2024inquirenaturalworldtexttoimage} extend this landscape to ecological challenges, rigorously evaluating expert-level text-to-image retrieval tasks critical for accelerating biodiversity research.

\begin{table}[ht]
    \centering
    \caption{Specialized Deep Research Benchmarks}
    \label{tab:specialized-benchmarks}
    \begin{adjustbox}{width=\textwidth}
    \begin{tabular}{lccc}
        \toprule
        \textbf{Benchmark} & \textbf{Focus Area} & \textbf{Evaluation Approach} & \textbf{Key Metrics} \\
        \midrule
        AAAR-1.0 \cite{lou2025aaar10assessingaispotential} & Research assistance & 150 multi-domain tasks & Retrieval and reasoning capability \\
        DSBench \cite{jing2025dsbenchfardatascience} & Data science & 20 real-world tasks & End-to-end completion rate \\
        SciCode \cite{tian2024scicoderesearchcodingbenchmark} & Scientific coding & Curated by scientists & Code quality, task completion \\
        MASSW \cite{zhang2024masswnewdatasetbenchmark} & Scientific workflows & Benchmarking tasks & Workflow orchestration quality \\
        MMSci \cite{li2025mmscidatasetgraduatelevelmultidiscipline} & Multimodal science & Graduate-level questions & Cross-modal understanding \\
        TPBench \cite{Theoretical_Physics_Benchmark} & Theoretical physics & Physics reasoning & Problem-solving accuracy \\
        \bottomrule
    \end{tabular}
    \end{adjustbox}
    \scriptsize
    \textit{Note: These benchmarks represent domain-specific evaluation frameworks for specialized research capabilities.}
\end{table}

\subsubsection{Qualitative Assessment Frameworks}

Beyond numeric benchmarks, qualitative evaluation provides insight into practical effectiveness:

\begin{table}[ht]
    \centering
    \caption{Documented Output Characteristics of Deep Research Systems}
    \label{tab:output-characteristics}
    \begin{adjustbox}{width=\textwidth}
    \begin{tabular}{lcccc}
        \toprule
        \textbf{System} & \textbf{Content Organization} & \textbf{Information Diversity} & \textbf{Verification Features} & \textbf{Novel Connection Mechanisms} \\
        \midrule
        \path{OpenAI/Deep Research} \cite{openai2025} & Hierarchical structure with 5+ sections, executive summaries & Cross-domain source integration (reported in \cite{openai2025}) & Statement-level citation linking, contradiction flagging & Cross-domain connection identification \\
        
        \path{Gemini/Deep Research} \cite{deep_research_now_available_gemini} & Multi-level heading organization, standardized formatting & Multi-perspective source inclusion (documented in \cite{deep_research_now_available_gemini}) & Source credibility metrics, confidence indicators & Thematic pattern identification \\
        
        \path{Perplexity/Deep Research} \cite{perplexity2025} & Progressive information disclosure, expandable sections & Real-time source aggregation across platforms & Direct quote attribution, inline source linking & Timeline-based relationship mapping \\
        
        \path{mshumer/OpenDeepResearcher} \cite{mshumer2024} & Template-based document structure, consistent formatting & Topic-based categorization of sources & Basic citation framework, reference listing & Topic cluster visualization \\
        
        \path{grapeot/deep_research_agent} \cite{grapeot2024} & Minimal formatting, content-focused presentation & Source type categorization, domain tracking & Source credibility scoring system based on metadata & Not implemented per repository documentation \\
        
        \path{Agent-RL/ReSearch} \cite{agentrl2024} & Adaptive content organization based on information types & Exploratory search patterns documented in repository & Contradiction detection algorithms & Pattern-based insight generation documented in \cite{agentrl2024} \\
        \bottomrule
    \end{tabular}
    \end{adjustbox}
    \scriptsize
    \textit{Note: Characteristics documented based on system technical documentation, published demonstrations, repository analysis, and official descriptions as of April 2025. Specific feature implementations may vary across system versions.}
\end{table}


Commercial systems generally demonstrate stronger qualitative performance, particularly in output coherence and factual accuracy. \path{OpenAI/Deep Research} \cite{openai2025} produces exceptionally well-structured reports with reliable factual content, while also achieving moderate innovation in connecting disparate sources. \path{Gemini/Deep Research} \cite{deep_research_now_available_gemini} shows similar strengths in coherence and accuracy, with slightly less emphasis on novel insights.

Some open-source implementations show particular strengths in specific dimensions. \path{Agent-RL/ReSearch} \cite{agentrl2024} achieves notable performance in insight novelty through its exploration-focused approach, while 
\path{grapeot/deep_research_agent} \cite{grapeot2024} demonstrates strong factual accuracy through its emphasis on information verification. These specialized capabilities highlight the diversity of approaches within the Deep Research ecosystem.

\subsubsection{Efficiency and Resource Utilization Metrics}

Practical deployment considerations include computational requirements and operational efficiency:

\begin{table}[ht]
    \centering
    \caption{Efficiency and Resource Utilization}
    \label{tab:efficiency}
    \begin{adjustbox}{width=\textwidth}
    \begin{tabular}{lccccc}
        \toprule
        \textbf{System} & \textbf{Response Time*} & \textbf{Compute Requirements} & \textbf{Token Efficiency**} \\
        \midrule
        \path{OpenAI/Deep Research} \cite{openai2025} & 5-30 min & Cloud-only & High (detailed, citation-rich) \\
        \path{Perplexity/Deep Research} \cite{perplexity2025} & 2m59s & Cloud-only & - \\
        \path{Grok 3 Beta} \cite{Grok_3_Beta} & - & Cloud-only & - \\
        \path{Nanobrowser} \cite{nanobrowser2024} & - & User-defined via LLM API key & - \\     
        \path{n8n} \cite{n8n2024} & - & Self-hosted or cloud-based; scalable & - \\     
        \bottomrule
    \end{tabular}
    \end{adjustbox}
    \small{*Typical response time for moderately complex research tasks}\\
    \small{**Efficiency of token utilization relative to output quality}
\end{table}

Commercial cloud-based services offer optimized performance with moderate response times, though with dependency on external infrastructure and associated costs. \path{Perplexity/Deep Research} \cite{perplexity2025} achieves particularly strong efficiency metrics, with relatively quick response times and high token efficiency despite its competitive output quality.

Open-source implementations present greater variability in efficiency metrics. Systems like \path{AutoGLM-Research} \cite{autoglm_research2025} and \path{QwenLM/Qwen-Agent} \cite{qwen2025} require substantial computational resources but can be deployed in local environments, offering greater control and potential cost savings for high-volume usage. Lighter-weight implementations like \path{nickscamara/open-deep-research} \cite{nickscamara2024} can operate with more limited resources but typically demonstrate longer response times and lower token efficiency.

This comparative analysis highlights the diversity of approaches and capabilities across the Deep Research ecosystem. While commercial implementations currently demonstrate leading performance on standard benchmarks, open-source alternatives offer competitive capabilities in specific domains and use cases, with particular advantages in customization, control, and potential cost efficiency for specialized applications. The subsequent sections will build on this analysis to examine implementation technologies, evaluation methodologies, and application domains in greater detail.

\section{Implementation Technologies and Challenges}

The practical realization of Deep Research systems involves numerous technical challenges spanning infrastructure design, system integration, and safeguard implementation. This section examines the key implementation technologies that enable effective Deep Research capabilities and the challenges that must be addressed for reliable, efficient operation.

\begin{figure}[ht]
    \centering
    \includegraphics[width=1.0\linewidth]{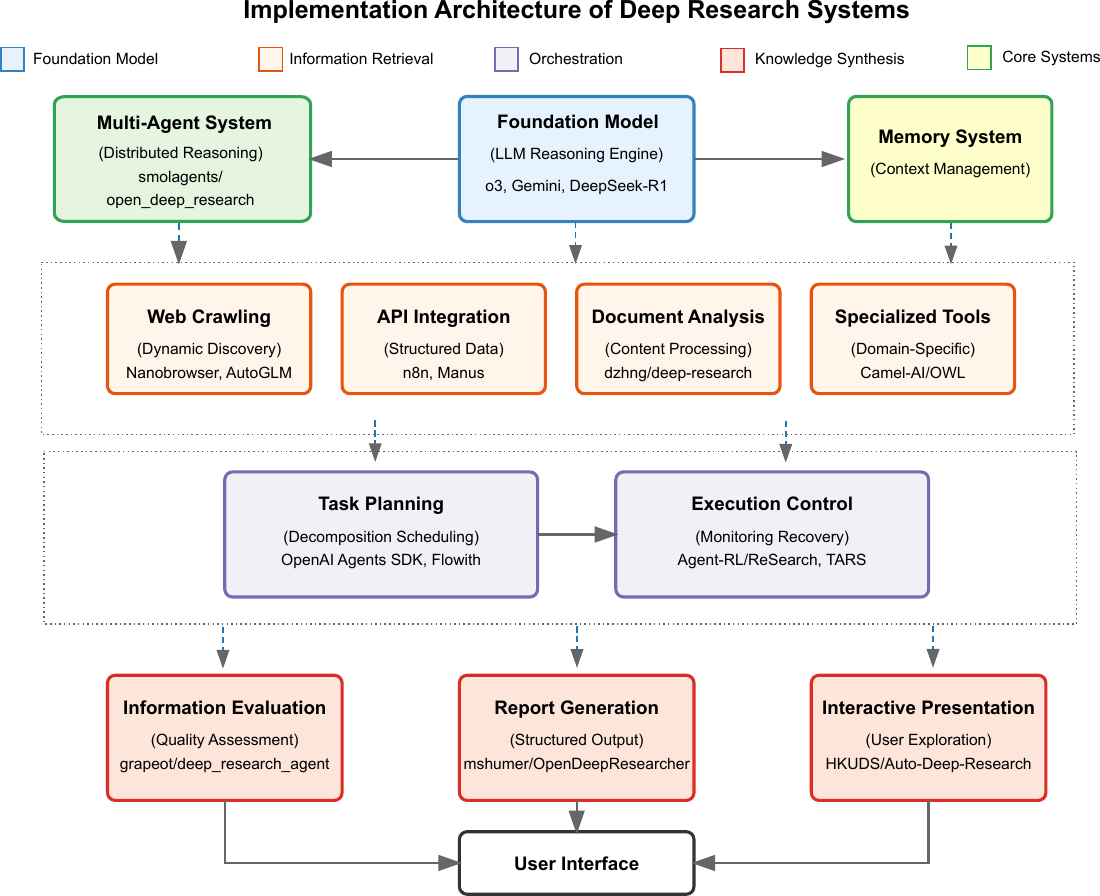}
    \caption{Implementation Architecture of Deep Research Systems}
    \label{fig:implementation}
\end{figure}

\subsection{Architectural Implementation Patterns}
\label{sec:architectural-patterns}

The diverse systems analyzed in this survey reveal several distinct architectural patterns that represent different approaches to implementing Deep Research capabilities. This section examines four fundamental architectural patterns: monolithic, pipeline-based, multi-agent, and hybrid implementations. For each pattern, we analyze the underlying structural principles, component interactions, information flow mechanisms, and representative systems.

\subsubsection{Monolithic Architecture Pattern}

\begin{figure}[ht]
    \centering
    \includegraphics[width=1.0\linewidth]{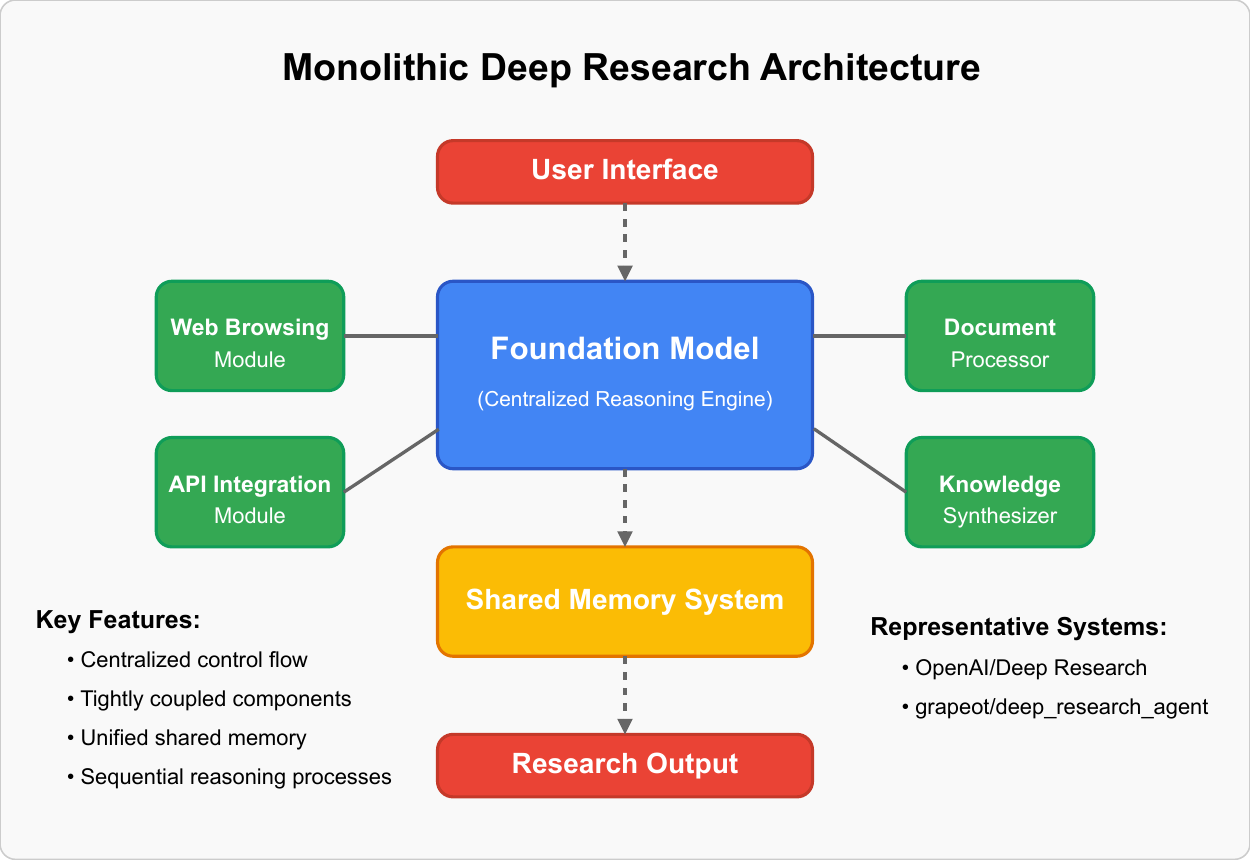}
    \caption{Monolithic Deep Research Architecture}
    \label{fig:monolithic-architecture}
\end{figure}

Monolithic implementations integrate all Deep Research capabilities within a unified architectural framework centered around a core reasoning engine. As illustrated in Figure~\ref{fig:monolithic-architecture}, these systems employ a centralized control mechanism with direct integration of specialized modules.

The defining characteristics of this architecture include:

\begin{itemize}
    \item \textbf{Centralized Control Flow:} All operations route through a primary reasoning engine that maintains global state and execution context
    \item \textbf{Tightly Coupled Integration:} Specialized modules (web browsing, document processing, etc.) are directly integrated with the central controller
    \item \textbf{Shared Memory Architecture:} Information state is maintained in a centralized memory system accessible to all components
    \item \textbf{Sequential Reasoning Processes:} Operations typically follow a structured sequence defined by the central controller
\end{itemize}

This architectural pattern offers strong coherence and reasoning consistency through its unified control structure. However, it presents challenges for extensibility and can struggle with parallelization of complex operations. Representative implementations include \path{OpenAI/Deep Research} \cite{openai2025} and \path{grapeot/deep_research_agent} \cite{grapeot2024}, which demonstrate how this architecture enables coherent reasoning across diverse information sources while maintaining implementation simplicity.

\subsubsection{Pipeline-Based Architecture Pattern}

\begin{figure}[ht]
    \centering
    \includegraphics[width=1.0\linewidth]{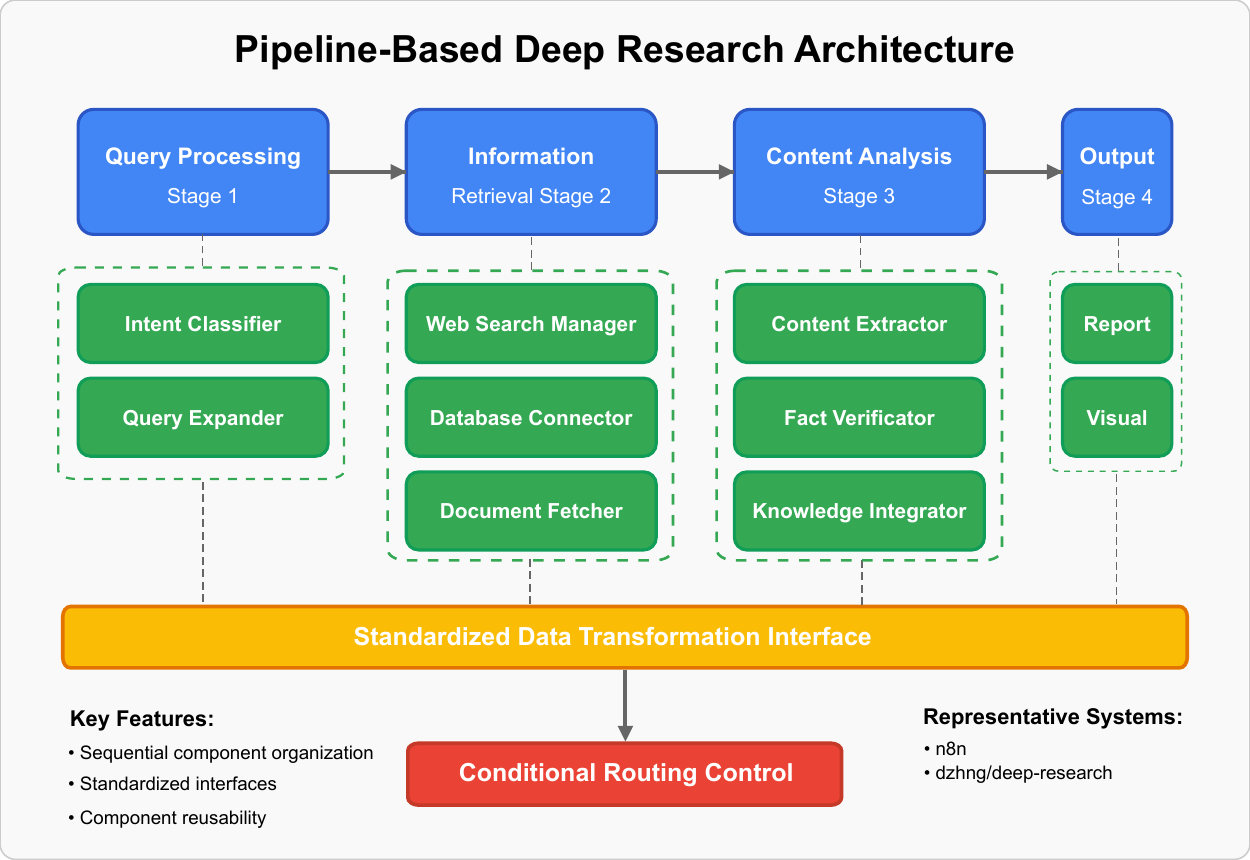}
    \caption{Pipeline-Based Deep Research Architecture}
    \label{fig:pipeline-architecture}
\end{figure}

Pipeline architectures implement Deep Research capabilities through a sequence of specialized processing stages connected through well-defined interfaces. As shown in Figure~\ref{fig:pipeline-architecture}, these systems decompose research workflows into discrete processing components with explicit data transformations between stages.

The key characteristics of pipeline implementations include:

\begin{itemize}
    \item \textbf{Sequential Component Organization:} Research tasks flow through a predefined sequence of specialized processing modules
    \item \textbf{Standardized Interfaces:} Clear data transformation specifications between pipeline stages enable modular component replacement
    \item \textbf{Staged Processing Logic:} Each component implements a specific transformation, with minimal dependence on global state
    \item \textbf{Configurable Workflow Paths:} Advanced implementations enable conditional routing between alternative processing paths based on intermediary results
\end{itemize}

Pipeline architectures excel in workflow customization and component reusability but may struggle with complex reasoning tasks requiring iterative refinement across components. Systems like \path{n8n} \cite{n8n2024} and \path{dzhng/deep-research} \cite{dzhng2024} exemplify this approach, demonstrating how explicit workflow sequencing enables sophisticated research automation through composition of specialized components.

\subsubsection{Multi-Agent Architecture Pattern}

\begin{figure}[ht]
    \centering
    \includegraphics[width=1.0\linewidth]{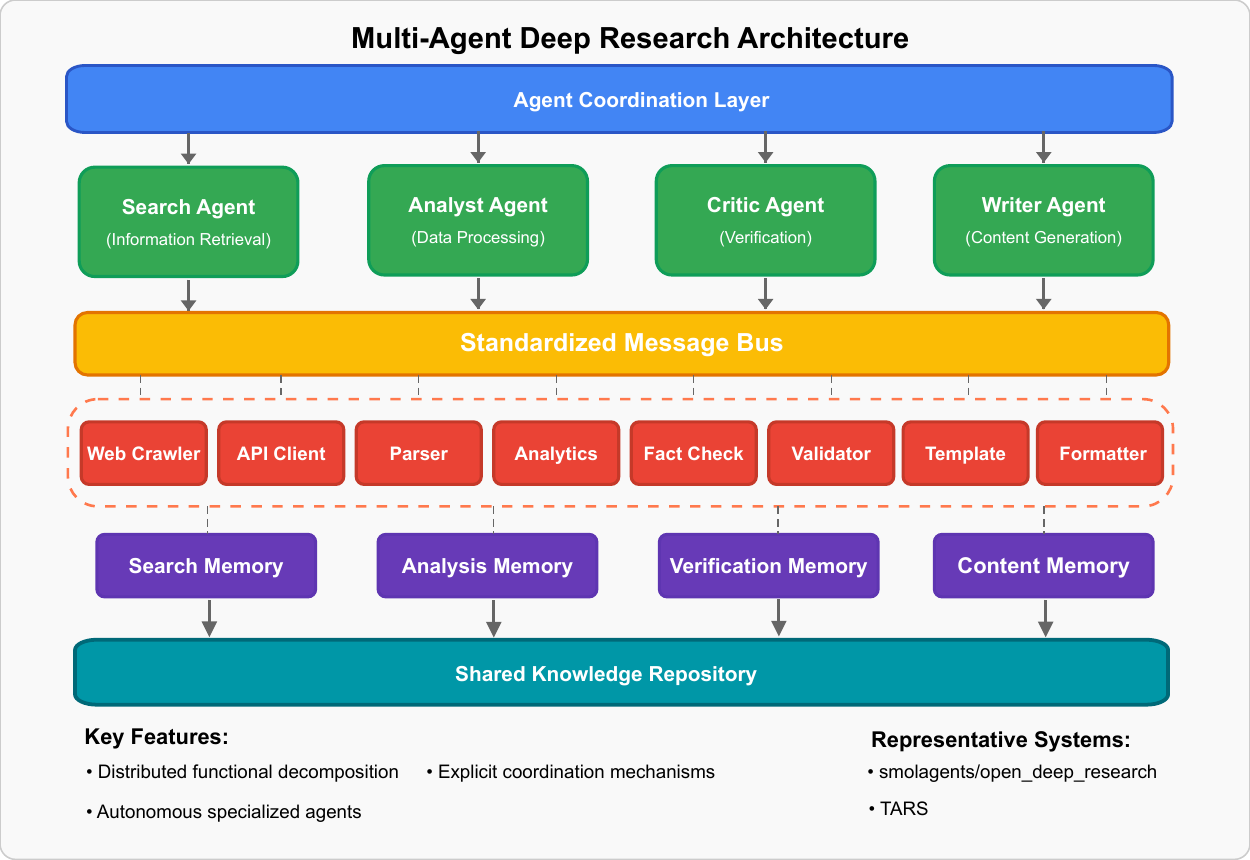}
    \caption{Multi-Agent Deep Research Architecture}
    \label{fig:multi-agent-architecture}
\end{figure}

Multi-agent architectures implement Deep Research capabilities through ecosystems of specialized autonomous agents coordinated through explicit communication protocols. Figure~\ref{fig:multi-agent-architecture} illustrates how these systems distribute research functionality across collaborating agents with differentiated roles and responsibilities.

The defining elements of multi-agent implementations include:

\begin{itemize}
    \item \textbf{Distributed Functional Decomposition:} Research capabilities are distributed across specialized agents with defined roles (searcher, analyst, critic, etc.)
    \item \textbf{Explicit Coordination Mechanisms:} Standardized message passing and task delegation protocols enable inter-agent collaboration
    \item \textbf{Autonomous Decision Logic:} Individual agents maintain independent reasoning capabilities within their designated domains
    \item \textbf{Dynamic Task Allocation:} Advanced implementations employ flexible task assignment based on agent capabilities and current workload
\end{itemize}

Multi-agent architectures excel in complex research tasks requiring diverse specialized capabilities and parallel processing. Their distributed nature enables exceptional scaling for complex research workflows but introduces challenges in maintaining overall coherence and consistent reasoning across agents. Representative implementations include \path{smolagents/open_deep_research} \cite{smolagents2024} and \path{TARS} \cite{tars2025}, which demonstrate how multi-agent coordination enables sophisticated research workflows through specialized agent collaboration.

\subsubsection{Hybrid Architecture Pattern}

\begin{figure}[ht]
    \centering
    \includegraphics[width=1.0\linewidth]{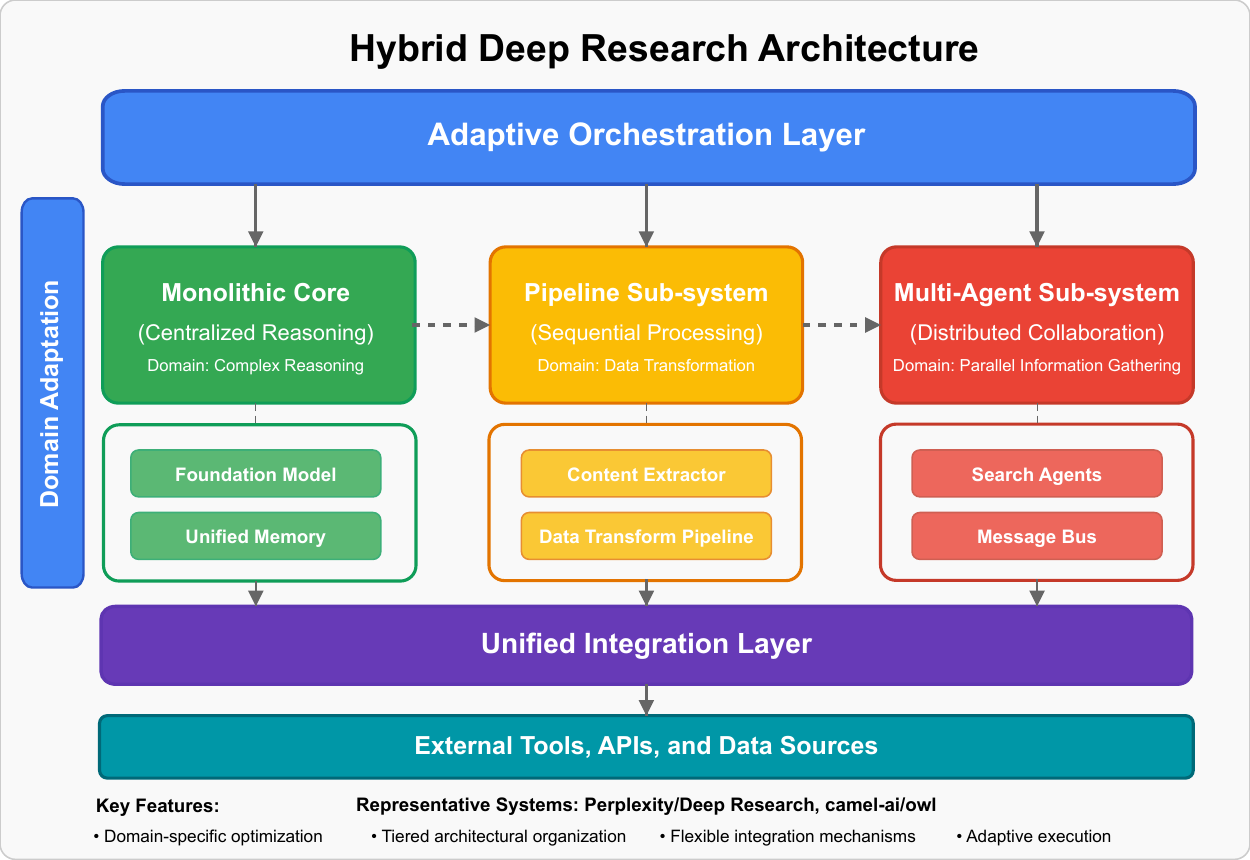}
    \caption{Hybrid Deep Research Architecture}
    \label{fig:hybrid-architecture}
\end{figure}

Hybrid architectures combine elements from multiple architectural patterns to balance their respective advantages within unified implementations. As shown in Figure~\ref{fig:hybrid-architecture}, these systems employ strategic integration of architectural approaches to address specific research requirements.

Key characteristics of hybrid implementations include:

\begin{itemize}
    \item \textbf{Tiered Architectural Organization:} Different architectural patterns are employed at different system levels based on functional requirements
    \item \textbf{Domain-Specific Optimization:} Architectural approaches are selected based on domain-specific processing requirements
    \item \textbf{Flexible Integration Mechanisms:} Standardized interfaces enable communication between components employing different architectural patterns
    \item \textbf{Adaptive Execution Frameworks:} Control mechanisms dynamically adjust processing approaches based on task characteristics
\end{itemize}

Hybrid architectures offer exceptional flexibility and optimization opportunities but introduce implementation complexity and potential integration challenges. Systems like \path{Perplexity/Deep Research} \cite{perplexity2025} and \path{Camel-AI/OWL} \cite{camel2025} exemplify this approach, combining centralized reasoning with distributed information gathering and specialized processing pipelines to achieve sophisticated research capabilities with balanced performance characteristics.

\subsubsection{Emerging Agent Framework Ecosystems}

Beyond the core architectural patterns described above, the Deep Research ecosystem has been significantly enhanced by specialized agent frameworks that provide standardized components for agent development. 
Emerging systems incorporate specialized agent frameworks \cite{xie2023openagentsopenplatformlanguage, li2023camelcommunicativeagentsmind,chen2023agentversefacilitatingmultiagentcollaboration} that structure reasoning in ways particularly suited to complex research tasks requiring both depth and breadth of analysis. 
As detailed in comprehensive analyses of agent frameworks \cite{howtothinkaboutagentframeworks, xu2024theagentcompanybenchmarkingllmagents}, these systems offer varying approaches to agent orchestration, execution control, and reasoning orchestration. 

Key frameworks include LangGraph \cite{LangGraph}, which provides graph-based control flow for language model applications, enabling complex reasoning patterns through explicit state management and transition logic. Google's Agent Development Kit (ADK) \cite{GoogleADK} offers a comprehensive framework for agent development with standardized interfaces for tool integration, planning, and execution monitoring. CrewAI \cite{CrewAI} implements an agent collaboration framework designed specifically for multi-specialist workflows, enabling role-based task distribution with explicit coordination mechanisms. More experimental frameworks like Agno \cite{Agno} explore agentic autonomy through self-improvement and meta-reasoning capabilities.

The TapeAgents framework \cite{bahdanau2024tapeagentsholisticframeworkagent} provides a particularly comprehensive approach to agent development and optimization, with explicit support for iterative refinement through systematic recording and analysis of agent behavior. These frameworks collectively demonstrate an ongoing shift toward standardized agent components that enhance development efficiency while enabling more complex reasoning and execution patterns.

\subsubsection{Architectural Pattern Comparison}

Table~\ref{tab:architecture-comparison} provides a comparative analysis of these architectural patterns across key performance dimensions:


\begin{table}[ht]
    \centering
    \caption{Architectural Pattern Characteristics in Deep Research Systems}
    \label{tab:architecture-comparison}
    \begin{adjustbox}{width=\textwidth}
    \begin{tabular}{lcccc}
        \toprule
        \textbf{Characteristic} & \textbf{Monolithic} & \textbf{Pipeline} & \textbf{Multi-Agent} & \textbf{Hybrid} \\
        \midrule
        Control Structure & Centralized & Sequential & Distributed & Mixed \\
        Component Coupling & Tight & Loose & Moderate & Variable \\
        Failure Propagation & System-wide & Stage-limited & Agent-isolated & Component-dependent \\
        Development Complexity & Minimal & Moderate & Substantial & Maximal \\
        Deployment Flexibility & Limited & Moderate & Moderate & High \\
        Representative Systems & \path{grapeot/deep_research_agent} & \path{n8n}, \path{dzhng/deep-research} & \path{smolagents}, \path{TARS} & \path{Perplexity}, \path{Camel-AI/OWL} \\
        \bottomrule
    \end{tabular}
    \end{adjustbox}
    \scriptsize
    \textit{Note: Characteristics based on architectural analysis of surveyed systems. Quantitative performance comparison requires standardized benchmarking across identical tasks and environments.}
\end{table}

Each architectural pattern presents distinct advantages and limitations that influence its suitability for specific Deep Research applications. Monolithic architectures excel in reasoning coherence and implementation simplicity, making them appropriate for focused research applications with well-defined workflows. Pipeline architectures offer exceptional extensibility and component reusability, enabling customized research workflows through modular composition. Multi-agent architectures provide superior parallelization and fault tolerance, supporting complex research tasks requiring diverse specialized capabilities. Hybrid architectures balance these characteristics through strategic integration, offering flexible optimization for diverse research requirements.

The architectural pattern selection significantly influences system capabilities, performance characteristics, and application suitability. As the Deep Research ecosystem continues to evolve, we anticipate further architectural innovation combining elements from these foundational patterns to address emerging application requirements and technical capabilities.

\subsection{Infrastructure and Computational Optimization}

Deep Research systems require sophisticated infrastructure to support their complex reasoning and information processing capabilities.

\subsubsection{Distributed Reasoning Architectures}

Effective reasoning across expansive information landscapes requires specialized architectural approaches. Frameworks like \path{AutoChain} \cite{AutoChain} and \path{AutoGen} \cite{AutoGen} have pioneered distributed agent paradigms that can be applied to research workflows. Advanced systems employ distributed reasoning architectures that decompose complex queries into parallel processing paths. \path{OpenAI/Deep Research} \cite{openai2025} implements a hierarchical reasoning framework that distributes analytical tasks across multiple execution threads while maintaining coherent central coordination. 

Implementation approaches increasingly leverage specialized frameworks for efficient LLM serving, including LightLLM \cite{lightllm}, Ollama \cite{Ollama}, VLLM \cite{vllm}, and Web-LLM \cite{web-llm} for browser-based deployment. These frameworks enable more efficient utilization of computational resources, particularly important for resource-intensive research workflows requiring extensive model inference. Such optimizations are especially critical for open-source implementations operating with more constrained computational resources compared to commercial cloud-based alternatives.

\paragraph{Parallel Reasoning Pathways}
Advanced systems employ distributed reasoning architectures that decompose complex queries into parallel processing paths. \path{OpenAI/Deep Research} \cite{openai2025} implements a hierarchical reasoning framework that distributes analytical tasks across multiple execution threads while maintaining coherent central coordination. Similar approaches are evident in \path{Gemini/Deep Research} \cite{deep_research_now_available_gemini}, which leverages Google's distributed computing infrastructure to parallelize information analysis while preserving reasoning consistency.

Open-source implementations like \path{HKUDS/Auto-Deep-Research} \cite{hkuds2024} and \path{Agent-RL/ReSearch} \cite{agentrl2024} demonstrate more accessible distributed reasoning approaches, utilizing task decomposition and asynchronous processing to enhance performance within more constrained computational environments. These systems show that effective parallelization can be achieved even without the extensive infrastructure of commercial platforms.

\paragraph{Memory and State Management}
Distributed reasoning introduces significant challenges in memory coherence and state management. 
Commercial systems implement sophisticated state synchronization mechanisms that maintain consistent reasoning contexts across distributed components. OpenAI's implementation utilizes a hierarchical memory architecture with explicit coordination protocols \cite{o3ando4mini}, while Google's approach leverages its existing distributed computing frameworks adapted for reasoning workflows.

Open-source alternatives like \path{Camel-AI/OWL} \cite{camel2025} employ simplified but effective memory management approaches, including centralized knowledge repositories with controlled access patterns. These implementations demonstrate pragmatic solutions to state management challenges within more constrained technical environments.

\subsubsection{Parallel Search and Information Retrieval}

Information acquisition represents a primary bottleneck in Deep Research performance:

\paragraph{Concurrent Query Execution}
Advanced systems implement sophisticated parallel search infrastructures to accelerate information gathering. \path{Perplexity/Deep Research} \cite{perplexity2025} employs a multi-threaded search architecture that dispatches dozens of concurrent queries across different information sources, significantly accelerating the research process. Similar capabilities are evident in \path{dzhng/deep-research} \cite{dzhng2024}, which implements a specialized scheduler for concurrent web queries with adaptive rate limiting to avoid service restrictions.

Infrastructure tools like \path{Nanobrowser} \cite{nanobrowser2024} provide optimized platforms for parallel browsing operations, enabling multiple concurrent page loads with shared resource management. These specialized components enhance the information gathering capabilities of integrated systems like \path{Manus} \cite{manus2025} and \path{Flowith/OracleMode} \cite{flowith2025}, which leverage concurrent browsing to accelerate their research workflows.

\paragraph{Query Coordination and Deduplication}
Effective parallel search requires sophisticated coordination to avoid redundancy and ensure comprehensive coverage. Commercial systems implement advanced query planning that dynamically adapts to intermediate results, adjusting search strategies based on discovered information. OpenAI's implementation includes explicit deduplication mechanisms that identify and consolidate redundant sources, while Perplexity employs source diversification techniques to ensure broad coverage.

Open-source tools like \path{nickscamara/open-deep-research} \cite{nickscamara2024} implement pragmatic approaches to query coordination, including simple but effective caching mechanisms and result fingerprinting to avoid redundant processing. These techniques demonstrate that effective coordination can be achieved with relatively straightforward implementation approaches.

\subsubsection{Resource Allocation and Efficiency Optimization}

Computational efficiency significantly impacts both performance and operational economics:

\paragraph{Adaptive Resource Allocation}
Advanced systems implement dynamic resource allocation based on task characteristics and complexity. \path{Gemini/Deep Research} \cite{deep_research_now_available_gemini} employs sophisticated workload prediction to provision computational resources adaptively, allocating additional capacity for more complex research tasks. Similar approaches are emerging in open-source implementations like \path{QwenLM/Qwen-Agent} \cite{qwen2025}, which incorporates task complexity estimation to guide resource allocation decisions.

\paragraph{Progressive Processing Strategies}
Efficiency-focused implementations employ progressive processing approaches that incrementally refine results based on available information. \path{Perplexity/Deep Research} \cite{perplexity2025} utilizes a staged analysis approach that provides preliminary findings quickly while continuing deeper analysis in the background. This strategy enhances perceived responsiveness while ensuring comprehensive results for complex queries.

Open-source alternatives like \path{mshumer/OpenDeepResearcher} \cite{mshumer2024} implement simpler but effective progressive strategies, including early result previews and incremental report generation. These approaches demonstrate pragmatic solutions to efficiency challenges without requiring sophisticated infrastructure.

\subsection{System Integration and Interoperability}

Deep Research systems must effectively coordinate diverse components and external services to deliver comprehensive capabilities.

\subsubsection{API Design and Standardization}

Consistent interfaces enable modular development and component interoperability:

\paragraph{Component Interface Standardization} 
Current Deep Research implementations employ largely incompatible architectures and interfaces. Future research could build upon emerging standardization efforts like Anthropic's Model Context Protocol (MCP) \cite{mcp} and Google's Agent2Agent Protocol (A2A) \cite{A2A, A2A_a_new_era_of_agent_interoperability} to establish truly universal component interfaces. MCP provides a structured framework for model-tool interaction, enabling consistent integration patterns across diverse LLM applications, while A2A focuses on standardized agent-to-agent communication to facilitate multi-agent systems. These complementary approaches could form the foundation for comprehensive standardization enabling modular development and interchangeable components across implementations. Early steps in this direction appear in frameworks like \path{OpenAI/AgentsSDK} \cite{openaiagents2025}, which provides standardized agent definitions, but more comprehensive standardization would require broader industry adoption of common protocols. 

\paragraph{Workflow Automation}
Several workflow automation platforms like \path{Dify} \cite{dify}, \path{Coze} \cite{coze}, and \path{Flowise} \cite{flowise} have emerged as low-code environments for building LLM-powered applications, potentially offering standardized frameworks for Deep Research components. 
Advanced workflow orchestration platforms including Temporal \cite{Temporal}, Restate \cite{Restate}, and Orkes \cite{Orkes} provide robust infrastructure for complex, stateful workflows with explicit support for long-running processes and reliability patterns crucial for sophisticated research applications. 
Implementation approaches might include defining standard message passing protocols between research components, establishing common data structures for research tasks and results, developing compatibility layers between competing standards, extending existing protocols with research-specific interaction patterns, and establishing common evaluation frameworks for component interoperability. These advances could accelerate ecosystem development by enabling specialized components from diverse developers to work seamlessly within unified frameworks, significantly enhancing the pace of innovation through componentization and reuse.

\paragraph{External Service Integration}
Access to specialized external services significantly enhances research capabilities. Advanced retrieval frameworks like LlamaIndex~\cite{llama_index} provide standardized interfaces for retrieval augmentation, enabling consistent integration patterns across diverse information sources and document formats. 
Systems like \path{n8n} \cite{n8n2024} excel in external service integration through their comprehensive connector library and standardized authentication mechanisms. This capability enables access to specialized information sources and analytical services that extend beyond basic web search.

Open-source frameworks like \path{Jina-AI/node-DeepResearch} \cite{jina2025} implement simplified but effective API integration patterns, providing standardized wrappers for common services while maintaining extensibility for custom integrations. These approaches balance standardization with flexibility for diverse research requirements.

\subsubsection{Tool Integration Frameworks}

Effective orchestration of diverse tools enhances overall system capabilities:

\paragraph{Tool Selection and Composition}
Advanced systems implement sophisticated tool selection based on task requirements and information context. \path{Manus} \cite{manus2025} features an adaptive tool selection framework that identifies appropriate tools for specific research subtasks, dynamically composing workflows based on available capabilities. Similar approaches are emerging in open-source implementations like \path{grapeot/deep_research_agent} \cite{grapeot2024}, which includes basic tool selection heuristics based on task classification.

\paragraph{Tool Execution Monitoring}
Reliable tool usage requires effective execution monitoring and error handling. Commercial systems implement sophisticated monitoring frameworks that track tool execution, detect failures, and implement recovery strategies. OpenAI's implementation includes explicit success criteria verification and fallback mechanisms for tool failures, ensuring reliable operation even with unreliable external components.

Open implementations like \path{Agent-RL/ReSearch} \cite{agentrl2024} demonstrate more accessible monitoring approaches, including simplified execution tracking and basic retry mechanisms for common failure modes. These implementations show that effective monitoring can be achieved with relatively straightforward implementation strategies.

Recent advances in agent collaboration frameworks~\cite{li2023metaagentssimulatinginteractionshuman, qiao2025benchmarkingagenticworkflowgeneration} highlight significant challenges in agent coordination~\cite{cemri2025multiagentllmsystemsfail}, particularly for complex research tasks requiring diverse, specialized capabilities working in concert toward unified research objectives.

\subsubsection{Cross-Platform Compatibility}

Deployment flexibility requires careful attention to environmental dependencies:

\paragraph{Platform Abstraction Layers}
Cross-platform implementations employ abstraction layers to isolate core logic from environmental dependencies. \path{TARS} \cite{tars2025} implements a sophisticated abstraction architecture that separates its core reasoning framework from platform-specific integration components, enabling deployment across diverse environments. Similar approaches are evident in \path{Nanobrowser} \cite{nanobrowser2024}, which provides consistent browsing capabilities across different operating systems.

\paragraph{Containerization and Deployment Standardization}
Modern implementations leverage containerization to ensure consistent deployment across environments. \path{OpenManus} \cite{openmanus2025} provides explicit container configurations that encapsulate all dependencies, enabling reliable deployment across diverse infrastructures. Similar approaches are employed by \path{AutoGLM-Research} \cite{autoglm_research2025}, which provides standardized deployment configurations for different environments. Alongside containerization, modern cloud platforms such as Vercel~\cite{vercel} offer streamlined, standardized deployment workflows for the web-based interfaces of many research applications.

\subsubsection{Research-Oriented Coding Assistance Integration}

The integration of AI-powered coding assistants represents an increasingly important dimension of Deep Research system capabilities, particularly for computational research workflows requiring custom analysis scripts, data processing pipelines\cite{Hiniduma_2024}, and research automation tools.

\paragraph{Coding Assistant Integration Patterns}
Modern research workflows increasingly depend on custom code development for data analysis, visualization, and automation tasks. AI coding assistants have emerged as crucial tools for enhancing researcher productivity in these computational aspects. The landscape of coding assistance tools demonstrates varying approaches to integration with research workflows, from IDE-native completion systems to conversational code generation interfaces. Systems like GitHub Copilot \cite{copilotgithub,bakal2025experiencegithubcopilotdeveloper} provide seamless integration within development environments, enabling context-aware code completion for research scripts and analysis workflows. Complementary approaches like ChatGPT-based code generation \cite{yetiştiren2023evaluatingcodequalityaiassisted} offer conversational interfaces that can translate research requirements into executable implementations. More specialized frameworks like AutoDev \cite{tufano2024autodevautomatedaidrivendevelopment}, \path{DSPy}\cite{DSPy}, and \path{Pydantic-AI}\cite{Pydantic-AI} enable end-to-end automated development workflows particularly suited for research prototype generation and experimental tool creation. Additionally, tools like Bolt \cite{bolt} allow researchers to create web applications directly from text descriptions, handling the coding process while they focus on their vision. Evolutionary coding agents like AlphaEvolve \cite{alphaevolve} further enhance capabilities by iteratively optimizing algorithms using autonomous pipelines of LLMs and evolutionary feedback mechanisms. Recent research explores the synergy between generative AI and software engineering, leveraging techniques like zero-shot prompting to enhance coding assistants and streamline development processes \cite{calegario2023exploringintersectiongenerativeai}. However, research has revealed limitations in these assistants’ capabilities, such as ambiguous beliefs regarding research claims and a lack of credible evidence to support their responses \cite{brown2024exploringevidencebasedbeliefsbehaviors}. A large-scale survey demonstrates that developers frequently decline initial suggestions, citing unmet functional or non-functional requirements and challenges in controlling the tool to generate desired outputs \cite{liang2023largescalesurveyusabilityai}. User resistance behaviors documented in such surveys highlight the need for comprehensive adoption strategies, including providing active support during initial use, clearly communicating system capabilities, and adhering to predefined collaboration rules to mitigate low acceptance rates\cite{simkute2024itthereneedit}. This underscores the need for adaptive hint systems, which can provide personalized support for bug finding and fixing by tailoring to user understanding levels and program representations to improve accuracy in debugging tasks \cite{rawal2024hintshelpfindingfixing}. Pioneering studies employ physiological measurements such as EEG and eye tracking to quantify developers' cognitive load during AI-assisted programming tasks, addressing critical gaps in understanding actual usage patterns and productivity impacts \cite{haque2025decodingdevelopercognitionage}. Furthermore, tools like CodeScribe address challenges in AI-driven code translation for scientific computing by combining prompt engineering with user supervision to automate conversion processes while ensuring correctness \cite{dhruv2025leveraginglargelanguagemodels}. Similarly, CodeCompose's multi-line suggestion feature deployed at Meta demonstrates substantial productivity improvements, saving 17\% of keystrokes through optimized latency solutions despite initial usability challenges \cite{dunay2024multilineaiassistedcodeauthoring}. Moreover, for debugging tasks, ChatDBG \cite{levin2025chatdbgaipowereddebuggingassistant} enhances debugging capabilities by enabling programmers to engage in collaborative dialogues for root cause analysis and bug resolution, leveraging LLMs to provide domain-specific reasoning. Intelligent QA assistants are also being developed to streamline bug resolution processes~\cite{yao2024bugblitzaiintelligentqaassistant}, and grey literature reviews indicate a growing trend in AI-assisted test automation~\cite{A_Multi-Year_Grey_Literature_Review_on_AI-assisted_Test_Automation}. Additionally, benchmarks like CodeMMLU \cite{manh2025codemmlumultitaskbenchmarkassessing} evaluate code understanding and reasoning across diverse tasks, revealing significant comprehension gaps in current models despite advanced generative capabilities. Empirical evaluations of ACATs through controlled development scenarios demonstrate nuanced variations in acceptance patterns, modification reasons, and effectiveness based on task characteristics and user expertise \cite{tan2024faraipoweredprogrammingassistants}. Generative AI tools significantly enhance developer productivity by accelerating learning processes and altering collaborative team workflows through reduced repetitive tasks, fundamentally transforming development paradigms \cite{ulfsnes2024transformingsoftwaredevelopmentgenerative}. To realize the vision of next-generation AI coding assistants, it is crucial to address integration gaps and establish robust design principles such as setting clear usage expectations and employing extendable backend architectures \cite{nghiem2024envisioningnextgenerationaicoding}.

\begin{table}[ht]
    \centering
    \caption{Qualitative Assessment of AI Coding Assistants for Research Applications}
    \label{tab:coding-assistants}
    \begin{adjustbox}{width=\textwidth}
    \begin{tabular}{lcccc}
        \toprule
        \textbf{System} & \textbf{Documented Capabilities} & \textbf{Integration Approach} & \textbf{Evaluation Evidence} & \textbf{Research-Specific Features} \\
        \midrule
        GitHub Copilot \cite{copilotgithub, Zhang_2023} & Code completion, documentation & IDE-native integration & User study on practices \cite{Zhang_2023} & Limited domain specialization \\
        Amazon CodeWhisperer \cite{Github-Copilot-Amazon-Whisperer-ChatGPT} & Security-focused suggestions & AWS ecosystem integration & Comparative evaluation \cite{yetiştiren2023evaluatingcodequalityaiassisted} & Cloud research workflows \\
        ChatGPT Code \cite{yetiştiren2023evaluatingcodequalityaiassisted} & Conversational code generation & API-based interaction & Code quality assessment \cite{yetiştiren2023evaluatingcodequalityaiassisted} & Natural language specification \\
        Cursor \cite{cursor} & Context-aware completion & Codebase integration & No published evaluation & Repository-level understanding \\
        Codeium \cite{ovi2024benchmarkingchatgptcodeiumgithub} & Multi-language support & Editor extensions & Comparative benchmark \cite{ovi2024benchmarkingchatgptcodeiumgithub} & Analysis workflow support \\
        AutoDev \cite{tufano2024autodevautomatedaidrivendevelopment} & Automated development & Task automation pipeline & Empirical evaluation \cite{tufano2024autodevautomatedaidrivendevelopment} & End-to-end implementation \\
        GPT-Pilot \cite{gpt_pilot} & Project scaffolding & Guided development process & Repository demonstrations & Research prototype generation \\
        \bottomrule
    \end{tabular}
    \end{adjustbox}
    \scriptsize
    \textit{Note: Capabilities and evaluations based on published studies and documented features. Comparative performance requires standardized evaluation across identical tasks.}
\end{table}

The diversity of coding assistance approaches highlights the importance of integration flexibility within Deep Research systems. While some implementations benefit from tightly integrated coding assistance that understands research context, others require more flexible interfaces that can accommodate diverse development workflows and programming paradigms. This integration dimension becomes particularly crucial as research increasingly requires custom computational tools and analysis pipelines that extend beyond pre-existing software packages\cite{Natural_Language_Generation_and_Understanding_of_Big_Code,Natural_Language_is_a_Programming_Language,Sergeyuk_2025}. Recent work by Chen et al. \cite{chen2025needhelpdesigningproactive} demonstrates that proactive programming assistants, which automatically provide suggestions to enhance productivity and user experience, represent a key advancement in this domain. Additionally, ChatDev \cite{chatdev} exemplifies how linguistic communication serves as a unifying bridge for multi-agent collaboration in software development, streamlining the entire lifecycle from design to testing. Moreover, research on integrating AI assistants in Agile meetings reveals critical links to team collaboration dynamics and provides roadmaps for facilitating their adoption in development contexts \cite{cabrerodaniel2024exploringhumanaicollaborationagile}. As demonstrated by Talissa Dreossi\cite{Bridging_Logic_Programming_and_Deep_Learning_for_Explainability_through_ILASP}, this hybrid approach bridges the gap between the high performance of deep learning models and the transparency of symbolic reasoning, advancing AI by providing interpretable and trustworthy applications.

\paragraph{Research Workflow Code Generation}
Advanced coding assistants specifically optimized for research contexts demonstrate particular value in translating research methodologies into executable implementations. Systems like GPT-Pilot \cite{gpt_pilot} enable guided development of complete research applications, while domain-specific tools can generate analysis scripts aligned with particular research methodologies or data types. These capabilities enhance research efficiency by reducing the technical barriers between research design and computational implementation.

Implementation patterns typically involve integration with research data management systems, version control workflows, and collaborative development environments that support reproducible research practices. The effectiveness of such integration depends significantly on the coding assistant's understanding of research-specific requirements including documentation standards, reproducibility considerations, and domain-specific libraries and frameworks commonly used in particular research fields\cite{Joshi_2024}.

\subsection{Technical Challenges and Solutions}

Deep Research systems face numerous technical challenges that must be addressed for reliable, trustworthy operation.

\subsubsection{Hallucination Control and Factual Consistency}

Maintaining factual accuracy represents a fundamental challenge for LLM-based research systems:

\paragraph{Source Grounding Techniques}
Advanced implementations employ explicit source grounding to enhance factual reliability. \path{Perplexity/Deep Research} \cite{perplexity2025} implements strict attribution requirements that link all generated content to specific sources, reducing unsupported assertions. Similar approaches are evident in \path{OpenAI/Deep Research} \cite{openai2025}, which maintains explicit provenance tracking throughout the reasoning process.

Open-source implementations like \path{grapeot/deep_research_agent} \cite{grapeot2024} demonstrate more accessible grounding approaches, including simple but effective citation tracking and verification mechanisms. These techniques show that meaningful improvements in factual reliability can be achieved with straightforward implementation strategies.

\paragraph{Contradiction Detection and Resolution}
Effective research requires identification and resolution of contradictory information. Commercial systems implement sophisticated contradiction detection mechanisms that identify inconsistencies between sources and implement resolution strategies \cite{A_Survey_on_LLM-Generated_Text_Detection}. \path{Gemini/Deep Research} \cite{deep_research_now_available_gemini} includes explicit uncertainty modeling and conflicting evidence presentation, enhancing transparency when definitive conclusions cannot be reached.

Open implementations like \path{HKUDS/Auto-Deep-Research} \cite{hkuds2024} employ simpler but useful contradiction identification approaches, flagging potential inconsistencies for user review. These implementations demonstrate that even basic contradiction handling can significantly enhance research reliability.

\subsubsection{Privacy Protection and Security Design}

Research systems must safeguard sensitive information and protect against potential misuse:

\paragraph{Query and Result Isolation}
Secure implementations employ strict isolation between user queries to prevent information leakage. Commercial platforms implement sophisticated tenant isolation that ensures complete separation between different users' research activities. Similar concerns motivate open-source implementations like \path{OpenManus} \cite{openmanus2025}, which enables local deployment for sensitive research applications.

\paragraph{Source Data Protection}
Responsible implementation requires careful handling of source information. Systems like \path{Flowith/OracleMode} \cite{flowith2025} implement controlled data access patterns that respect source restrictions including authentication requirements and access limitations. These approaches enhance compliance with source terms of service while ensuring comprehensive information access. Recent advancements include benchmarking frameworks such as CI-Bench \cite{cheng2024cibenchbenchmarkingcontextualintegrity}, which evaluates how well systems adhere to contextual norms and privacy expectations.

\subsubsection{Explainability and Transparency}

The scientific context places particularly stringent requirements on explanation quality. Mengaldo~\cite{mengaldo2025explainblackboxsake} argues that transparent explanation is not merely a feature but a fundamental requirement for scientific applications, emphasizing that black-box approaches fundamentally contradict scientific methodology's requirement for transparent reasoning and reproducible results. This perspective suggests that explanation capabilities may require different standards in scientific Deep Research applications compared to general AI systems. 
Trustworthy research systems must provide insight into their reasoning processes and sources:

\paragraph{Reasoning Trail Documentation}
Advanced implementations maintain explicit documentation of the reasoning process. \path{OpenAI/Deep Research} \cite{openai2025} includes comprehensive reasoning traces that expose the analytical steps leading to specific conclusions. Similar capabilities are emerging in open-source alternatives like \path{mshumer/OpenDeepResearcher} \cite{mshumer2024}, which includes basic reasoning documentation to enhance result interpretability. 

\paragraph{Source Attribution and Verification}
Transparent systems provide clear attribution for all information and enable verification. \path{Perplexity/Deep Research} \cite{perplexity2025} implements comprehensive citation practices with explicit links to original sources, enabling direct verification of all claims. Similar approaches are employed by \path{dzhng/deep-research} \cite{dzhng2024}, which maintains rigorous source tracking throughout the research process.

These implementation technologies and challenges highlight the complex engineering considerations involved in creating effective Deep Research systems. While commercial platforms benefit from extensive infrastructure and specialized components, open-source implementations demonstrate that effective research capabilities can be achieved through pragmatic approaches to the same fundamental challenges. The diversity of implementation strategies across the ecosystem reflects different priorities in balancing capability, efficiency, reliability, and accessibility.

\section{Evaluation Methodologies and Benchmarks}

Rigorous evaluation of Deep Research systems presents unique challenges due to their complex capabilities and diverse application contexts. This section examines established frameworks for assessment, identifies emerging evaluation standards, and analyzes the strengths and limitations of current approaches.

\begin{figure}[ht]
    \centering
    \includegraphics[width=1.0\linewidth]{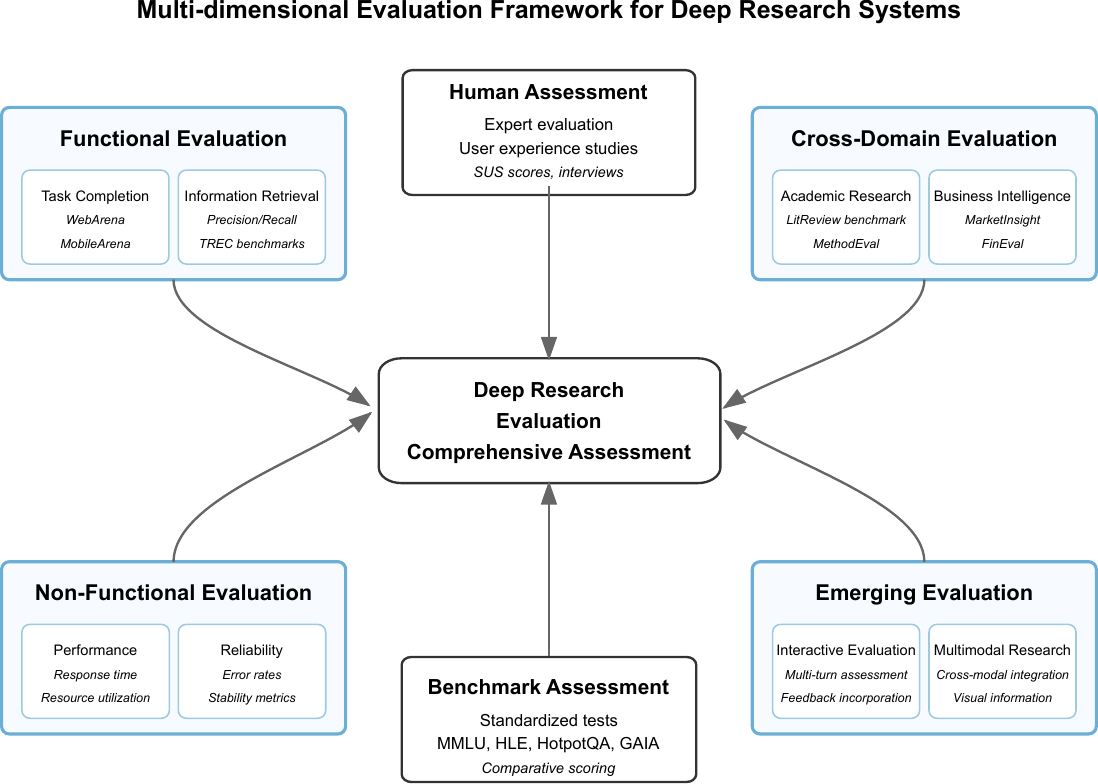}
    \caption{Multi-dimensional Evaluation Framework for Deep Research Systems}
    \label{fig:evaluation-framework}
\end{figure}

\subsection{Functional Evaluation Frameworks}

Functional evaluation assesses core capabilities essential to effective research performance.

\subsubsection{Task Completion Capability Assessment}

The ability to successfully complete research tasks represents a fundamental evaluation dimension:

\paragraph{Task Success Rate Metrics}
Quantitative assessment of task completion provides objective performance measures. Standardized evaluation suites like WebArena \cite{webarena2024} measure successful completion of web-based research tasks. For instance, \path{AutoGLM} \cite{autoglm_research2025} achieves a 55.2\% success rate on VAB-WebArena-Lite (improving to 59.1\% on a second attempt) and 96.2\% on OpenTable evaluation tasks. Similarly, benchmarks like MobileArena evaluate successful completion of mobile interface tasks, where \path{AutoGLM} \cite{autoglm_research2025} demonstrates a 36.2\% success rate on AndroidLab and 89.7\% on common tasks in popular Chinese apps \cite{AutoGLM_Autonomous_Foundation_Agents_for_GUIs}. Domain-specific benchmarks, such as AutoPenBench for generative agents in penetration testing~\cite{gioacchini2024autopenbenchbenchmarkinggenerativeagents}, provide further targeted assessments. These benchmarks provide meaningful comparative metrics, though with limitations in representing real-world research complexity.

These benchmarks provide meaningful comparative metrics, though with limitations in representing real-world research complexity. \path{Perplexity/Deep Research} \cite{perplexity2025} explicitly highlights this distinction, noting that while benchmark performance provides comparative indicators, practical effectiveness depends significantly on task characteristics and domain specifics.

\paragraph{Multi-Attempt Resolution Rates}
Effective research often involves iterative refinement with multiple attempts. Advanced evaluation frameworks incorporate multi-attempt metrics that assess system resilience and adaptability. \path{AutoGLM} \cite{AutoGLM2024} demonstrates significant performance improvement with second attempts (55.2\% to 59.1\% on WebArena-Lite), highlighting the importance of error recovery and adaptive strategies in practical research contexts.

Open-source frameworks like \path{Agent-RL/ReSearch} \cite{agentrl2024} explicitly emphasize iterative improvement through reinforcement learning approaches, demonstrating how evaluation methods that consider adaptability provide more comprehensive assessment than single-attempt metrics alone.

\subsubsection{Information Retrieval Quality Evaluation}

Effective information gathering forms the foundation of successful research:

\paragraph{Search Effectiveness Metrics}
Information retrieval quality significantly impacts overall research performance. Evaluation frameworks employ metrics including precision (relevance of retrieved information), recall (comprehensiveness of coverage), and F1 scores (balanced measure of both). Systems like \path{Perplexity/Deep Research} \cite{perplexity2025} demonstrate particular strength in recall metrics, effectively identifying comprehensive information across diverse sources.

Specialized information retrieval benchmarks like TREC \cite{trec} provide standardized assessment of search effectiveness. However, to the best of our knowledge, there is no specific evidence that the Deep Research systems from OpenAI, Google, Perplexity, or any of the open-source projects listed in this survey have been formally evaluated on TREC benchmarks \cite{trec} . 
This limitation motivates domain-specific evaluation approaches that better reflect particular research requirements.

\paragraph{Source Diversity Assessment}
Comprehensive research requires balanced information from diverse perspectives and sources. Advanced evaluation frameworks incorporate explicit diversity metrics that assess the breadth of source utilization. Commercial systems like \path{Gemini/Deep Research} \cite{deep_research_now_available_gemini} emphasize source diversity as a key performance indicator, while open implementations like \path{dzhng/deep-research} \cite{dzhng2024} incorporate specific mechanisms to ensure balanced source consideration.

Emerging evaluation approaches include explicit source spectra analysis that examines distribution across domains, perspectives, and publication types. These methods provide more nuanced assessment of information gathering quality beyond simple relevance metrics, addressing concerns about potential bias in automated research processes.

\subsubsection{Knowledge Synthesis Accuracy Assessment}

Transforming information into accurate, coherent insights represents a crucial capability:

\paragraph{Factual Consistency Metrics}
Reliable research requires accurate synthesis without introducing errors or misrepresentations. Evaluation frameworks employ fact verification techniques that compare generated content against source materials, identifying potential inaccuracies or unsupported claims. Systems like \path{grapeot/deep_research_agent} \cite{grapeot2024} emphasize factual verification through explicit source linking, enabling direct accuracy assessment. Benchmark suites like TruthfulQA \cite{truthfulQA} assess the truthfulness of language models under challenging conditions. While specific accuracy figures for \path{OpenAI/Deep Research} \cite{openai2025} and \path{Perplexity/Deep Research} \cite{perplexity2025} on TruthfulQA \cite{truthfulQA} are not publicly available, these systems have demonstrated notable performance on other rigorous benchmarks. For instance, \path{OpenAI/Deep Research} \cite{openai2025} achieved a 26.6\% accuracy \cite{openai2025} on Humanity’s Last Exam (HLE) \cite{HLE}. Similarly, \path{Perplexity/Deep Research} \cite{perplexity2025} attained a 21.1\% accuracy \cite{perplexity2025} on the same benchmark. The development of unified, fine-grained, and multi-dimensional evaluation frameworks for summarization further advances the ability to assess the quality of synthesized content from LLMs~\cite{lee2024unisumevalunifiedfinegrainedmultidimensional}. These metrics provide standardized comparison points, though with recognized limitations in representing the complexity of real-world research synthesis.

\paragraph{Logical Coherence Assessment}
Effective research requires logically sound integration of information into coherent analyses. Sophisticated evaluation approaches employ reasoning validity assessment that examines logical structures and inference patterns in research outputs. This dimension proves particularly challenging for automated assessment, often requiring expert human evaluation for reliable scoring.

Commercial systems like \path{OpenAI/Deep Research} \cite{openai2025} and \path{Gemini/Deep Research} \cite{deep_research_now_available_gemini} emphasize logical coherence in their evaluation frameworks, while open-source alternatives like \path{mshumer/OpenDeepResearcher} \cite{mshumer2024} incorporate simplified but useful logical consistency checks. These approaches highlight the importance of sound reasoning in effective research outputs beyond simple factual accuracy.

\subsection{Non-Functional Evaluation Metrics}

Beyond core functionality, practical effectiveness depends on operational characteristics that impact usability and deployment.

\subsubsection{Performance and Efficiency Metrics}

Operational efficiency significantly impacts practical utility:

\paragraph{Response Time Profiling}
Timeliness represents a crucial dimension of research effectiveness. Evaluation frameworks incorporate response time metrics that measure completion duration across standardized tasks. Commercial systems demonstrate varying performance characteristics, with \path{Perplexity/Deep Research} \cite{perplexity2025} achieving relatively quick response times (2-5 minutes for moderate tasks) while \path{OpenAI/Deep Research} \cite{openai2025} typically requires longer processing (5-10 minutes) for similar complexity.

Open-source implementations generally demonstrate longer response times, though with significant variation based on implementation approaches and deployment environments. Systems like \path{nickscamara/open-deep-research} \cite{nickscamara2024} emphasize accessibility over performance optimization, while \path{QwenLM/Qwen-Agent} \cite{qwen2025} incorporates specific optimizations to enhance response times within resource constraints.

\paragraph{Resource Utilization Assessment}
Computational efficiency enables broader deployment and accessibility. Comprehensive evaluation includes resource profiling that measures memory consumption, computational requirements, and energy utilization across standardized workloads. Specialized benchmarks like Minerva assess programmable memory capabilities of language models, offering insights into their efficiency in handling long-context information~\cite{Minerva}. Commercial cloud-based systems obscure some of these metrics due to their managed infrastructure, though with operational costs providing indirect resource indicators. Open implementations like \path{Camel-AI/OWL} \cite{camel2025} and \path{AutoGLM-Research} \cite{autoglm_research2025} provide more transparent resource profiles, enabling direct assessment of deployment requirements and operational economics. These metrics highlight significant variation in efficiency across the ecosystem, with implications for practical deployment scenarios and accessibility.

\subsubsection{Reliability and Stability Metrics}

Consistent performance under diverse conditions ensures practical usability:

\paragraph{Error Rate Analysis}
Reliability under challenging conditions significantly impacts user trust and adoption. Robust evaluation frameworks incorporate error rate metrics that measure failure frequency across diverse scenarios. Commercial systems generally demonstrate lower error rates compared to open-source alternatives, though with remaining challenges in complex or novel research contexts.

Specialized reliability testing employs adversarial scenarios designed to trigger failure modes, providing insight into system robustness. Systems like \path{OpenAI/Deep Research} \cite{openai2025} and \path{Agent-RL/ReSearch} \cite{agentrl2024} incorporate explicit error recovery mechanisms that enhance reliability under challenging conditions, highlighting the importance of resilience in practical research applications.

\paragraph{Long-Term Stability Assessment}
Consistent performance over extended operation provides crucial deployment confidence. Comprehensive evaluation includes stability metrics that measure performance consistency across extended sessions and repeated executions. This dimension proves particularly relevant for open-source implementations that must operate in diverse deployment environments with varying infrastructure stability.

Systems like \path{Flowith/OracleMode} \cite{flowith2025} and \path{TARS} \cite{tars2025} emphasize operational stability through robust error handling and recovery mechanisms, enabling reliable performance in production environments. These capabilities highlight the importance of engineering quality beyond core algorithmic performance in practical research applications.

\subsubsection{User Experience and Usability Metrics}

Effective interaction significantly impacts practical utility:

\paragraph{Interface Usability Assessment}
Intuitive interfaces enhance accessibility and effective utilization. Usability evaluation 
frameworks employ standardized usability metrics including System Usability Scale (SUS) \cite{Lewis03072018} scores and task completion time measurements. Commercial systems typically demonstrate stronger usability characteristics, with \path{Perplexity/Deep Research} \cite{perplexity2025} particularly emphasizing intuitive interaction for non-technical users. Open-source alternatives show greater variability, with implementations like \path{HKUDS/Auto-Deep-Research} \cite{hkuds2024} incorporating specific interface enhancements to improve accessibility.

User studies provide more nuanced usability assessment beyond standardized metrics. Evaluations of systems like \path{Manus} \cite{manus2025} and \path{Flowith/OracleMode} \cite{flowith2025} incorporate explicit user feedback to identify interaction challenges and improvement opportunities. These approaches highlight the importance of human-centered design in practical research applications beyond technical performance. Similarly, frameworks such as \path{AdaptoML-UX} \cite{gomaa2024adaptomluxadaptiveusercenteredguibased} enable HCI researchers to employ automated ML pipelines without specialized expertise, facilitating robust model development and customization.

\paragraph{Learning Curve Assessment}
Approachability for new users significantly impacts adoption and effective utilization. Comprehensive evaluation includes learning curve metrics that measure time-to-proficiency across user segments with varying technical backgrounds. Commercial systems generally demonstrate gentler learning curves, with \path{Perplexity/Deep Research} \cite{perplexity2025} explicitly designed for accessibility to non-technical users.

Open implementations show greater variability, with systems like \path{n8n} \cite{n8n2024} requiring more technical expertise for effective deployment and utilization. More accessible alternatives like \path{nickscamara/open-deep-research} \cite{nickscamara2024} incorporate simplified interfaces designed for broader accessibility, highlighting diverse approaches to the accessibility-sophistication balance across the ecosystem.

\subsection{Cross-Domain Evaluation Benchmarks}

Standardized benchmarks enable objective comparison across systems and domains.

\subsubsection{Academic Research Task Benchmarks}

Specialized benchmarks assess capabilities relevant to scholarly research:

\paragraph{Literature Review Benchmarks}
Comprehensive literature synthesis represents a fundamental academic research task requiring sophisticated information retrieval, critical analysis, and synthesis capabilities. To the best of our knowledge, no benchmark suite is specifically designed to evaluate systems' ability to identify relevant literature, synthesize key findings, and highlight research gaps across scientific domains. We propose leveraging existing high-quality literature reviews published in \textit{Nature Reviews} journals as gold standards. Citation networks from academic knowledge graphs—such as Microsoft Academic Graph, Semantic Scholar Academic Graph, and Open Academic Graph—could provide complementary evaluation data by measuring a system's ability to traverse citation relationships and identify seminal works\cite{Artificial_Intelligence_for_Literature_Reviews,Survey-on-Vision-Language-Action-Models}.

While direct literature review benchmarks remain underdeveloped, several indirect benchmarks offer insight into related capabilities. \path{OpenAI/Deep Research} \cite{openai2025} demonstrates leading performance, achieving 26.6\% accuracy on Humanity's Last Exam (HLE) \cite{HLE} and averaging 72.57\% on the GAIA benchmark \cite{GAIA}, reflecting strong performance in complex reasoning tasks essential for literature synthesis. Similarly, \path{Perplexity/Deep Research} \cite{perplexity2025} achieves 21.1\% accuracy on HLE \cite{HLE} and 93.9\% on SimpleQA \cite{simpleqa}, indicating robust factual retrieval capabilities.

These benchmarks include challenging cases requiring integration across multiple disciplines, identification of methodological limitations, and disambiguation of conflicting findings—all crucial for effective literature review. Such tasks demonstrate the importance of sophisticated reasoning capabilities beyond simple information retrieval. While specific performance metrics for systems like \path{Camel-AI/OWL} \cite{camel2025} are not publicly available, their specialized academic optimization suggests potential effectiveness in handling complex synthesis tasks.

\paragraph{Methodology Evaluation Benchmarks}
Critical assessment of research methodology requires sophisticated analytical capabilities. To the best of our knowledge, no benchmark is specifically designed for quantitative methodology assessment of strengths and limitations.
A comprehensive methodology evaluation benchmark would need to assess a system's ability to identify flaws in research design, statistical approaches, sampling methods, and interpretive limitations across diverse disciplines. An effective benchmark might incorporate multi-layered evaluation criteria including: reproducibility assessment, identification of confounding variables, appropriate statistical power analysis, and proper handling of uncertainty. Future benchmarks could utilize expert-annotated corpora of research papers with methodological strengths and weaknesses clearly marked, creating a gold standard against which systems' analytical capabilities can be measured while minimizing bias through diverse evaluation metrics that reflect methodological best practices across different fields of inquiry.

Beyond standard benchmarks, case study evaluations of complete AI scientist systems provide valuable insights into current capabilities. Beel et al.~\cite{beel2025evaluatingsakanasaiscientist} conduct a detailed assessment of Sakana's AI Scientist for autonomous research, examining whether current implementations represent genuine progress toward ``Artificial Research Intelligence'' or remain limited in fundamental ways, highlighting the gap between current benchmarks and comprehensive research capability evaluation.

\subsubsection{Business Analysis Task Benchmarks}

Standardized evaluation for business intelligence applications:

\paragraph{Market Analysis Benchmarks}
Strategic decision support necessitates a comprehensive understanding of market dynamics. Advanced AI systems, such as \path{OpenAI/Deep Research} \cite{openai2025}, are designed to analyze competitive landscapes, identify market trends, and generate strategic recommendations based on diverse business information. \path{OpenAI/Deep Research} has demonstrated significant capabilities in handling complex, multi-domain data analysis tasks, providing detailed insights and personalized recommendations. Similarly, Google's \path{Gemini/Deep Research} \cite{deep_research_now_available_gemini} offers robust performance in processing extensive datasets, delivering concise and factual reports efficiently.

These benchmarks include challenging scenarios requiring integration of quantitative financial data with qualitative market dynamics and regulatory considerations. Such tasks highlight the importance of both analytical depth and domain knowledge, with systems like \path{Manus} \cite{manus2025} demonstrating strong performance through specialized business intelligence capabilities.

\paragraph{Financial Analysis Benchmarks}
Economic assessment requires sophisticated quantitative reasoning combined with contextual understanding of market dynamics. The FinEval benchmark \cite{fineval} provides a standardized framework for measuring systems' capabilities in analyzing financial statements, evaluating investment opportunities, and assessing economic risk factors across diverse scenarios. To our knowledge, no Deep Research projects have yet published official FinEval benchmark results, though several commercial demonstrations suggest strong performance in this domain. \path{OpenAI/Deep Research} \cite{openai2025} has demonstrated particular strength in quantitative financial analysis through its ability to process complex numerical data while incorporating relevant market context. Meanwhile, open-source implementations show more variable performance, though specialized systems like \path{n8n} \cite{n8n2024} achieve competitive results through strategic integration with financial data sources and analytical tools. These patterns highlight the critical importance of domain-specific integrations and data accessibility in financial analysis applications, extending beyond core language model capabilities to create truly effective analytical systems.

\subsubsection{General Knowledge Management Benchmarks}

Broad applicability assessment across general research domains:

\paragraph{Factual Research Benchmarks}
Accurate information gathering forms the foundation of effective research. The SimpleQA benchmark \cite{simpleqa} evaluates language models' ability to answer short, fact-seeking questions with a single, indisputable answer. \path{Perplexity/Deep Research} \cite{perplexity2025} demonstrates exceptional performance on this benchmark, achieving an accuracy of 93.9\% \cite{perplexity2025}. OpenAI's Deep Research tool, integrated into ChatGPT, offers comprehensive research capabilities, though specific accuracy metrics on SimpleQA \cite{simpleqa} are not publicly disclosed \cite{openai2025}. Similarly, Google's \path{Gemini/Deep Research} provides robust information synthesis features, but detailed performance data on SimpleQA \cite{simpleqa} is not available.

These metrics provide useful baseline performance indicators, though with recognized limitations in representing more complex research workflows. Comparative evaluation highlights the importance of information quality beyond simple factual recall, with sophisticated systems demonstrating more nuanced performance profiles across complex tasks.

\paragraph{Humanities and Social Sciences Benchmarks}
Comprehensive evaluation requires assessment beyond STEM domains. The MMLU benchmark \cite{MMLU} evaluates systems' performance across humanities and social science research tasks, including historical analysis, ethical evaluation, and social trend identification. Performance shows greater variability compared to STEM-focused tasks, with generally lower accuracy across all systems while maintaining similar relative performance patterns.
These benchmarks highlight remaining challenges in domains requiring nuanced contextual understanding and interpretive reasoning. Commercial systems maintain performance leads, though with open alternatives like \path{smolagents/open_deep_research} \cite{smolagents2024} demonstrating competitive capabilities in specific humanities domains through specialized component design.

\subsection{Emerging Evaluation Approaches}

Beyond established benchmarks, novel evaluation methods address unique aspects of Deep Research performance.

\paragraph{Interactive Evaluation Frameworks}
Traditional static benchmarks often fail to capture the dynamic and interactive nature of real-world research workflows. To address this gap, interactive evaluation frameworks have been developed to assess AI systems' abilities to iteratively refine research strategies through multiple interaction rounds. 
Notably, QuestBench \cite{QuestBench} is a novel benchmark which specifically assesses an AI system's ability to identify missing information and ask appropriate clarification questions, a crucial skill for real-world research scenarios where problems are often underspecified.
To the best of our knowledge, no deep research system invested in this survey has yet been publicly evaluated using QuestBench. Nonetheless, these systems have demonstrated strong performance in other interactive evaluations, highlighting their effectiveness in supporting iterative research processes.

\paragraph{Multimodal Research Evaluation}
Comprehensive research increasingly involves diverse content modalities. Advanced evaluation frameworks incorporate multimodal assessment that measures systems' ability to integrate information across text, images, data visualizations, and structured content. Commercial systems generally demonstrate stronger multimodal capabilities, with \path{Gemini/Deep Research} \cite{deep_research_now_available_gemini} particularly excelling in image-inclusive research tasks.

Open implementations show emerging multimodal capabilities, with systems like \path{Jina-AI/node-DeepResearch} \cite{jina2025} incorporating specific components for multimodal content processing. These approaches highlight the growing importance of cross-modal integration in practical research applications beyond text-centric evaluation.

\paragraph{Ethical and Bias Assessment}
Responsible research requires careful attention to ethical considerations and potential biases. Comprehensive evaluation increasingly incorporates explicit assessment of ethical awareness, bias detection, and fairness in information processing. Commercial systems implement sophisticated safeguards, with \path{OpenAI/Deep Research} \cite{openai2025} incorporating explicit ethical guidelines and bias mitigation strategies. Open implementations show varied approaches to these considerations, with systems like \path{grapeot/deep_research_agent} \cite{grapeot2024} emphasizing transparency in source selection and attribution.

These evaluation dimensions highlight the importance of responsibility beyond technical performance, addressing growing concerns about potential amplification of existing information biases through automated research systems. Ongoing development of standardized ethical evaluation frameworks represents an active area of research with significant implications for system design and deployment.

The diverse evaluation approaches outlined in this section highlight both the complexity of comprehensive assessment and the ongoing evolution of evaluation methodologies alongside system capabilities. While standard benchmarks provide useful comparative metrics, practical effectiveness depends on alignment between system capabilities, evaluation criteria, and specific application requirements. This alignment represents a key consideration for both system developers and adopters seeking to integrate Deep Research capabilities into practical workflows.

\subsection{Comparative Evaluation Methodology}
\label{sec:evaluation-methodology}

To ensure systematic and consistent evaluation across diverse Deep Research systems, we have developed a comprehensive evaluation framework. This section outlines our methodological approach, evaluation criteria selection, and application consistency across systems.

\subsubsection{Systems Selection Criteria}

Our evaluation encompasses various Deep Research systems selected based on the following criteria:

\begin{itemize}
    \item \textbf{Functional Completeness:} Systems must implement at least two of the three core dimensions of Deep Research as defined in Section \ref{sec:definition_scope}
    \item \textbf{Public Documentation:} Sufficient technical documentation must be available to enable meaningful analysis
    \item \textbf{Active Development:} Systems must have demonstrated active development or usage within the past 12 months
    \item \textbf{Representational Balance:} Selection ensures balanced representation of commercial, open-source, general-purpose, and domain-specialized implementations
\end{itemize}

\subsubsection{Evaluation Dimensions and Metrics Application}

Our evaluation employs a consistent set of dimensions across all systems, though the specific benchmarks within each dimension vary based on system focus and available performance data. Table~\ref{tab:evaluation-coverage} presents the evaluation coverage across representative systems.

\begin{table}[ht]
    \centering
    \caption{Evaluation Metrics Application Across Systems}
    \label{tab:evaluation-coverage}
    \begin{adjustbox}{width=\textwidth}
    \begin{tabular}{lccccc}
        \toprule
        \textbf{System} & \textbf{Functional} & \textbf{Performance} & \textbf{Efficiency} & \textbf{Domain-Specific} & \textbf{Usability} \\
        & \textbf{Benchmarks} & \textbf{Metrics} & \textbf{Metrics} & \textbf{Benchmarks} & \textbf{Assessment} \\
        \midrule
        \path{OpenAI/Deep Research} & HLE, GAIA & Factual accuracy & Response time & Academic citation & User interface \\
        \path{Gemini/Deep Research} & MMLU & Output coherence & Cloud compute & Market analysis & Mobile support \\
        \path{Perplexity/Deep Research} & HLE, SimpleQA & Source diversity & Response time & Legal search & Multi-device \\
        \path{Grok 3 Beta} & MMLU & Source verification & Cloud efficiency & Financial analysis & Voice interface \\
        \path{Manus} & GAIA & Cross-domain & API latency & Business analysis & Dashboard \\
        \path{Agent-RL/ReSearch} & HotpotQA & Planning efficiency & Local compute & Scientific research & CLI interface \\
        \path{AutoGLM-Research} & WebArena & GUI navigation & Mobile efficiency & Domain adaptation & Accessibility \\
        \path{n8n} & Workflow & API integration & Self-hosted & Enterprise workflow & No-code design \\
        \bottomrule
    \end{tabular}
    \end{adjustbox}
\end{table}

\subsubsection{Data Collection Methods}

Our evaluation data comes from four primary sources:

\begin{enumerate}
    \item \textbf{Published Benchmarks:} Performance metrics reported in peer-reviewed literature or official system documentation
    \item \textbf{Technical Documentation Analysis:} Capabilities and limitations outlined in official documentation, APIs, and technical specifications
    \item \textbf{Repository Examination:} Analysis of open-source code repositories for architectural patterns and implementation approaches
    \item \textbf{Experimental Verification:} Where inconsistencies exist, we conducted direct testing of publicly available systems to verify capabilities
\end{enumerate}

When benchmark results are unavailable for specific systems, we indicate this gap explicitly rather than extrapolating performance. This approach ensures transparency regarding the limits of our comparative analysis while maintaining the integrity of available evaluation data.

\subsubsection{Cross-System Comparison Challenges}

Several methodological challenges exist in comparing Deep Research systems:

\begin{itemize}
    \item \textbf{Benchmark Diversity:} Different systems emphasize different benchmarks based on their focus areas
    \item \textbf{Implementation Transparency:} Commercial systems often provide limited details about internal architectures
    \item \textbf{Rapid Evolution:} Systems undergo frequent updates, potentially rendering specific benchmark results obsolete
    \item \textbf{Domain Specialization:} Domain-specific systems excel on targeted benchmarks but may perform poorly on general evaluations
\end{itemize}

We address these challenges through qualitative architectural analysis alongside quantitative benchmarks, enabling meaningful comparison despite data limitations. Section \ref{Sec3.3:Performance} presents the resulting comparative analysis, highlighting both performance differentials and the limitations of direct comparison across heterogeneous implementations.

\section{Applications and Use Cases}

The technical capabilities of Deep Research systems enable transformative applications across diverse domains. This section examines implementation patterns, domain-specific adaptations, and representative use cases that demonstrate the practical impact of these technologies.

\begin{figure}[ht]
    \centering
    \includegraphics[width=1.0\linewidth]{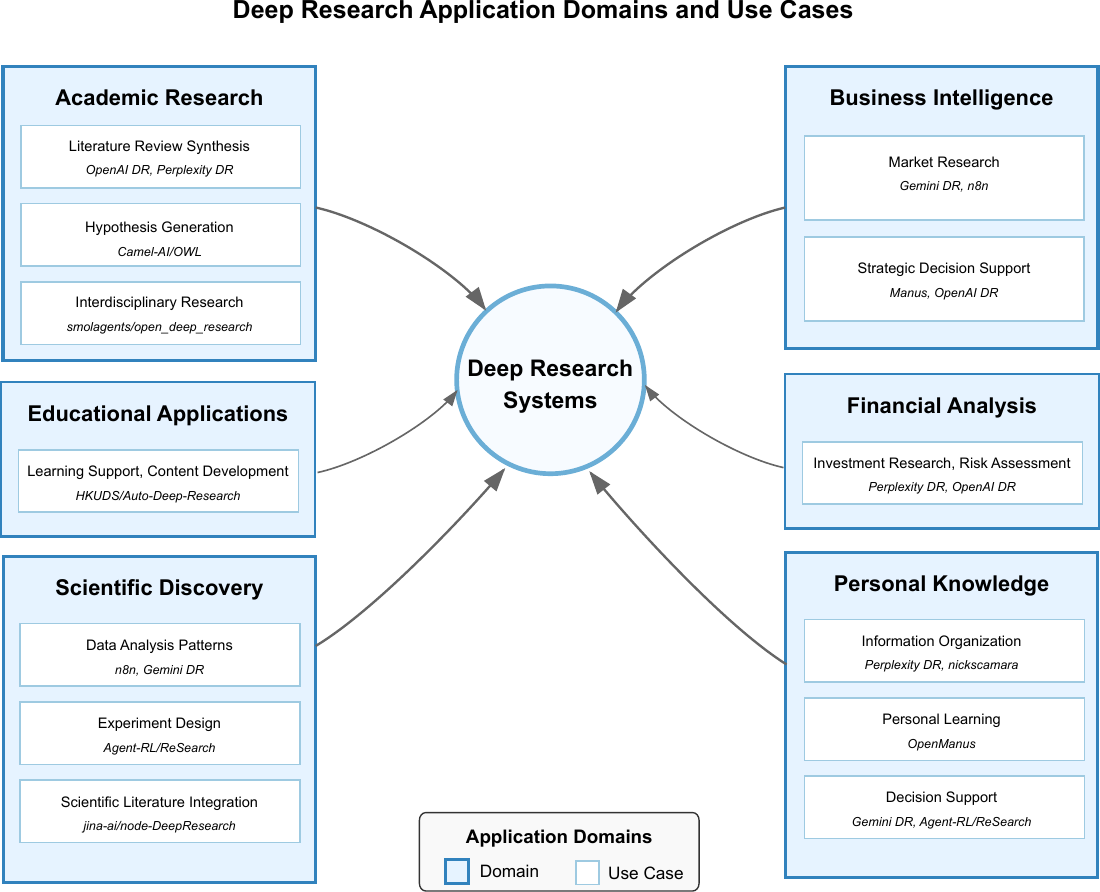}
    \caption{Deep Research Application Domains and Use Cases}
    \label{fig:applications}
\end{figure}

\subsection{Academic Research Applications}

Deep Research systems offer significant enhancements to scholarly research workflows.

\subsubsection{Literature Review and Synthesis}

Comprehensive literature analysis forms the foundation of effective research:

\paragraph{Systematic Review Automation}
Deep Research systems demonstrate particular effectiveness for systematic literature reviews requiring exhaustive coverage of existing research. Systems like Google's \path{Gemini/Deep Research} \cite{deep_research_now_available_gemini} can efficiently analyze thousands of research papers, a capability that has significant implications for fields like biomedicine where the volume of literature makes comprehensive manual review increasingly challenging \cite{biomedical}. \path{OpenAI/Deep Research} \cite{openai2025} has been successfully deployed for medical research reviews, analyzing thousands of publications to identify intervention efficacy patterns with significantly reduced human effort compared to traditional methods.
Similar capabilities are evident in \path{Perplexity/Deep Research} \cite{perplexity2025} and \path{Gemini/Deep Research} \cite{deep_research_now_available_gemini}, which enables rapid synthesis of research findings across disciplinary boundaries. Generative AI frameworks integrating retrieval-augmented generation further automate systematic reviews by expanding user queries to retrieve relevant scholarly articles and reduce time and resource burdens \cite{roy2023gearupgenerativeaiexternal}.

Open-source implementations like \path{dzhng/deep-research} \cite{dzhng2024} have found adoption in academic settings where local deployment and customization are prioritized. Specialized scientific implementations like \path{AI-Researcher} \cite{AI-Researcher} extend these capabilities with domain-specific optimizations for academic literature processing and analysis. These systems enable literature review automation with greater control over search scope and synthesis methods, particularly valuable for specialized research domains with unique requirements. Implementation patterns typically involve customization of search strategies, source weightings, and output formats to align with disciplinary conventions. 

\paragraph{Research Gap Identification}
Beyond simple synthesis, advanced systems effectively identify unexplored areas and research opportunities. \path{Gemini/Deep Research} \cite{deep_research_now_available_gemini} has demonstrated this capability in interdisciplinary contexts, identifying connection opportunities between distinct research domains that might otherwise remain undiscovered. This application leverages the system's ability to process extensive literature across fields while identifying patterns and absences in existing research coverage.

Open implementations like \path{HKUDS/Auto-Deep-Research} \cite{hkuds2024} incorporate specific mechanisms for gap analysis, including explicit detection of methodological limitations and underexplored variables across research corpora. These capabilities highlight the potential for automated systems to not only synthesize existing knowledge but actively contribute to research direction through systematic gap identification.

\subsubsection{Hypothesis Generation and Testing}

AI-assisted hypothesis development enhances research creativity and validation:

\paragraph{Hypothesis Formulation Support}
Deep Research systems effectively generate testable hypotheses based on existing literature and theoretical frameworks. \path{OpenAI/Deep Research} \cite{openai2025} provides explicit hypothesis generation capabilities, identifying potential causal relationships and testable predictions derived from literature synthesis. These features enable researchers to explore broader possibility spaces than might be practical through manual review alone.

Specialized frameworks like \path{Camel-AI/OWL} \cite{camel2025} implement domain-specific hypothesis generation for scientific applications, incorporating field-specific constraints and validation criteria. These approaches highlight how domain adaptation enhances the practical utility of hypothesis generation capabilities beyond generic formulation. Implementation patterns typically involve iterative refinement with researcher feedback to align generated hypotheses with specific research objectives.

\paragraph{Preliminary Validation Assessment}
Advanced systems support hypothesis validation through evidence assessment and methodological planning. \path{Gemini/Deep Research} \cite{deep_research_now_available_gemini} enables preliminary hypothesis testing through automated data source identification, statistical power analysis, and potential confound identification. These capabilities streamline the transition from hypothesis formulation to empirical testing, reducing manual effort in research design.

Open implementations like \path{Agent-RL/ReSearch} \cite{agentrl2024} incorporate specific validation planning components, guiding researchers through experimental design considerations based on hypothesis characteristics. These approaches demonstrate how Deep Research capabilities extend beyond information gathering to actively support the complete research workflow from conception through validation planning.

\subsubsection{Interdisciplinary Research Support}

Cross-domain integration represents a particular strength of automated research systems:

\paragraph{Cross-Domain Knowledge Translation}
Deep Research systems effectively bridge terminological and conceptual gaps between disciplines. \path{Perplexity/Deep Research} \cite{perplexity2025} demonstrates this capability through explicit concept mapping between fields, enabling researchers from diverse backgrounds to explore unfamiliar domains with reduced onboarding barriers. This application leverages the system's broad knowledge base to identify conceptual parallels across disciplinary boundaries.

Open frameworks like \path{smolagents/open_deep_research} \cite{smolagents2024} implement specialized agents for disciplinary translation, with explicit focus on terminological mapping and concept alignment. These approaches highlight how multi-agent architectures can effectively address the challenges of interdisciplinary communication through specialized component design\cite{CATER}.

\paragraph{Methodology Transfer Facilitation}
Advanced systems enable effective adaptation of research methods across domains. \path{OpenAI/Deep Research} \cite{openai2025} supports methodology transfer through explicit identification of adaptation requirements and implementation guidance when applying techniques from one field to another. This capability accelerates methodological innovation by facilitating cross-pollination between research traditions. Implementation patterns typically involve specialized methodological components like those in \path{QwenLM/Qwen-Agent} \cite{qwen2025}, which incorporates explicit methodology modeling to identify transfer opportunities and adaptation requirements. This is particularly relevant in fields like engineering, where AI is beginning to impact established design procedures for complex dynamical systems~\cite{depayrebrune2024impactaiengineeringdesign}. These approaches demonstrate how Deep Research systems can actively contribute to methodological innovation beyond simple information retrieval and synthesis.

\subsection{Scientific Discovery Applications}

Deep Research technologies enable enhanced scientific investigation across disciplines.

\subsubsection{Data Analysis and Pattern Recognition}

Automated analysis enhances insight extraction from complex scientific data:

\paragraph{Large-Scale Data Synthesis}
Deep Research systems effectively integrate findings across extensive datasets to identify broader patterns. \path{Gemini/Deep Research} \cite{deep_research_now_available_gemini} has been applied to climate science research, synthesizing findings across hundreds of climate models and observational datasets to identify consistent patterns and outliers. This application leverages the system's ability to process and integrate diverse data formats while maintaining analytical coherence. Open implementations like \path{n8n} \cite{n8n2024} enable similar capabilities through workflow automation that coordinates specialized analytical tools across complex data processing pipelines. Furthermore, SqlCompose \cite{maddila2024aiassistedsqlauthoringindustry} enhances analytical workflows by automating SQL authoring to reduce syntax barriers and improve efficiency in large-scale data operations, as demonstrated through enterprise deployment and user feedback. Systems like DataInquirer quantitatively measure workflow patterns and task execution consistency, revealing significant variations across practitioners while also assessing AI tool impacts on aligning novice approaches with expert practices \cite{zhao2024quantifyingdatascienceworkflows}. AI assistants specifically designed for data wrangling tasks can provide semi-automated support in transforming and cleaning data through interactive recommendations, thereby enhancing workflow efficiency \cite{petricek2022aiassistantsframeworksemiautomated}. Other systems assist domain experts in making sense of multi-modal personal tracking data through visualization and human-in-the-loop LLM agents~\cite{VitalInsight}. Additionally, no-code machine-readable documentation frameworks support responsible dataset evaluation by facilitating quality assessment and accuracy verification during large-scale data synthesis \cite{roman2024opendatasheetsmachinereadabledocumentation}. These approaches demonstrate how tool integration capabilities extend analytical reach beyond the core language model's native capabilities, particularly valuable for quantitative scientific applications.

\paragraph{Anomaly Detection and Investigation}
Advanced systems effectively identify unexpected patterns and facilitate targeted investigation. \path{OpenAI/Deep Research} \cite{openai2025} demonstrates this capability in pharmacological contexts, identifying unexpected drug interaction patterns across clinical literature and proposing mechanistic explanations for further investigation. This application combines pattern recognition with explanatory hypothesis generation to enhance scientific discovery.

Specialized tools like \path{grapeot/deep_research_agent} \cite{grapeot2024} implement focused anomaly detection capabilities, with particular emphasis on statistical outlier identification and contextual explanation. These approaches highlight how targeted optimization can enhance specific scientific workflows beyond general-purpose research capabilities\cite{Kazemitabaar_2024}.

\subsubsection{Experiment Design and Simulation}

AI assistance enhances experimental planning and virtual testing:

\paragraph{Experimental Protocol Optimization}
Deep Research systems support experimental design through comprehensive protocol development and optimization. \path{Gemini/Deep Research} \cite{deep_research_now_available_gemini} provides explicit protocol generation capabilities, incorporating existing methodological best practices while identifying potential confounds and control strategies. These features streamline experimental planning while enhancing methodological rigor.

Open implementations like \path{Agent-RL/ReSearch} \cite{agentrl2024} incorporate specialized experimental design components with particular emphasis on statistical power optimization and confound control. These approaches demonstrate how focused optimization can enhance specific scientific workflows through specialized component design targeting critical research phases.

Despite these capabilities, significant gaps remain between current systems and truly autonomous scientific discovery. Yu et al.~\cite{yu2025unlockingpotentialairesearchers} identify critical missing elements in current AI research systems, particularly highlighting limitations in open-ended exploration, creative hypothesis generation, and experimental design optimization that constrain their effectiveness in leading scientific discovery processes.

\paragraph{Theoretical Model Testing}
Advanced systems enable accelerated testing of theoretical models through simulation and virtual experimentation. \path{OpenAI/Deep Research} \cite{openai2025} supports this application through integration with computational modeling tools, enabling rapid assessment of theoretical predictions against existing evidence. This capability accelerates theory refinement by identifying empirical constraints and validation opportunities more efficiently than manual methods.

Implementation patterns typically involve specialized tool integration like that found in \path{Manus} \cite{manus2025}, which provides sophisticated orchestration of computational modeling and simulation tools within research workflows. Systems like \path{AgentLaboratory} \cite{AgentLaboratory} further enhance these capabilities through specialized experimental design components that generate statistically rigorous protocols based on research objectives and methodological best practices. 
These approaches highlight how tool integration capabilities significantly enhance scientific applications beyond the language model's native capabilities.

\subsubsection{Scientific Literature Integration}

Comprehensive knowledge integration enhances scientific understanding:

\paragraph{Cross-Modal Scientific Content Analysis}
Deep Research systems effectively integrate information across text, data, and visualizations prevalent in scientific literature. \path{Gemini/Deep Research} \cite{deep_research_now_available_gemini} demonstrates particular strength in this application, extracting and synthesizing information from scientific figures, tables, and text into cohesive analyses. This capability enables more comprehensive literature utilization than text-only approaches.

Open implementations like \path{Jina-AI/node-DeepResearch} \cite{jina2025} incorporate specialized components for multimodal scientific content processing, enabling similar capabilities in customizable frameworks. These approaches highlight the growing importance of multimodal processing in scientific applications, reflecting the diverse information formats prevalent in scientific communication.

\paragraph{Conflicting Evidence Resolution}
Advanced systems help navigate contradictory findings common in scientific literature. \path{Perplexity/Deep Research} \cite{perplexity2025} provides explicit conflict identification and resolution guidance, identifying methodological differences, contextual factors, and potential reconciliation approaches when faced with contradictory evidence. This capability enhances scientific understanding by providing structured approaches to evidence integration rather than simple aggregation.

Implementation patterns typically involve sophisticated evidence modeling like that found in \path{HKUDS/Auto-Deep-Research} \cite{hkuds2024}, which implements explicit evidence weighting and confidence estimation mechanisms. These approaches demonstrate how specialized components for scientific evidence handling enhance the practical utility of Deep Research systems in complex scientific contexts.

\subsubsection{Autonomous Scientific Discovery}

Fully autonomous research systems represent an emerging direction that extends current Deep Research capabilities toward greater autonomy. Recent work in this area includes the AI Scientist system~\cite{lu2024aiscientistfullyautomated} that implements an automated discovery loop with hypothesis generation, experimentation, and theory revision capacities. Similarly, the Dolphin system~\cite{yuan2025dolphinmovingclosedloopautoresearch} demonstrates how closed-loop auto-research can integrate thinking, practice, and feedback mechanisms to implement systematic scientific discovery processes.

This evolution toward more autonomous operation represents a significant advancement beyond traditional tool-based approaches, enabling continuous research cycles with minimal human intervention while maintaining scientific rigor through structured validation processes. Systems like CycleResearcher~\cite{weng2025cycleresearcherimprovingautomatedresearch} further enhance this approach by incorporating automated peer review mechanisms \cite{Lin_2023} that improve output quality through systematic feedback loops mimicking scientific review processes.

Practical implementation of these concepts appears in systems like AgentLaboratory \cite{schmidgall2025agentlaboratoryusingllm}, which demonstrates how LLM agents can function as effective research assistants within structured laboratory environments. Complementing these approaches, the concept of self-maintainability (SeM) addresses critical gaps in laboratory automation by enabling systems to autonomously adapt to disturbances and maintain operational readiness \cite{ochiai2025automatingcareselfmaintainabilitylaboratory}. In addition, strategies such as BOLAA \cite{liu2023bolaabenchmarkingorchestratingllmaugmented} orchestrate multiple specialized agents by employing a controller to manage communication among them, enhancing the resolution of complex tasks. Moreover, Automated Capability Discovery (ACD) \cite{lu2025automatedcapabilitydiscoverymodel} automates the evaluation of foundation models by designating one model as a scientist to propose open-ended tasks that systematically uncover unexpected capabilities and failures. Similarly, SeqMate \cite{mondal2024seqmatenovellargelanguage} utilizes large language models to automate RNA sequencing data preparation and analysis, enabling user-friendly one-click analytics and report generation for biologists. The FutureHouse Platform \cite{FutureHousePlatform} broadens accessibility by delivering the first publicly available superintelligent AI agents for scientific discovery through web interfaces and APIs. These implementations highlight both the significant potential and current limitations of autonomous scientific discovery systems, suggesting an evolutionary path toward increasingly capable research automation while maintaining appropriate human oversight and validation.

\subsection{Business Intelligence Applications}

Deep Research technologies enable enhanced strategic decision support in commercial contexts.

\subsubsection{Market Research and Competitive Analysis}

Comprehensive market understanding supports strategic planning:

\paragraph{Competitor Landscape Mapping}
Deep Research systems effectively synthesize comprehensive competitive intelligence across diverse sources. \path{Gemini/Deep Research} \cite{deep_research_now_available_gemini} enables detailed competitor analysis across financial disclosures, product announcements, market reception, and strategic positioning to identify competitive dynamics and market opportunities. This application leverages the system's ability to integrate information across public and specialized business sources with current market context.

Open implementations like \path{n8n} \cite{n8n2024} support similar capabilities through workflow automation that integrates specialized business intelligence data sources. These approaches demonstrate how effective tool integration can create sophisticated business intelligence applications by coordinating specialized components within consistent analytical frameworks.

\paragraph{Emerging Trend Identification}
Advanced systems effectively identify early-stage market trends and potential disruptions. \path{OpenAI/Deep Research} \cite{openai2025} demonstrates this capability through temporal pattern analysis across industry publications, startup activity, and technology development indicators. This application combines historical pattern recognition with current signal detection to anticipate market evolution with greater lead time than manual methods alone.

Implementation patterns typically involve specialized analytical components like those in \path{Flowith/OracleMode} \cite{flowith2025}, which incorporates explicit trend modeling and weak signal amplification techniques. These approaches highlight how specialized optimization enhances business intelligence applications through components targeting specific analytical requirements.

\subsubsection{Strategic Decision Support}

AI-enhanced analysis informs high-stakes business decisions:

\paragraph{Investment Opportunity Assessment}
Deep Research systems support investment analysis through comprehensive opportunity evaluation. \path{Perplexity/Deep Research} \cite{perplexity2025} enables detailed investment analysis incorporating financial metrics, market positioning, competitive dynamics, and growth indicators within unified analytical frameworks. This application integrates quantitative financial assessment with qualitative market understanding to support more comprehensive investment evaluation.

Open frameworks like \path{mshumer/OpenDeepResearcher} \cite{mshumer2024} implement investment analysis components with particular emphasis on structured evaluation frameworks and comprehensive source integration. These approaches demonstrate how domain-specific optimization enhances practical utility for specialized business applications beyond generic research capabilities.

\paragraph{Risk Factor Identification}
Advanced systems support risk management through comprehensive threat identification and assessment. \path{Gemini/Deep Research} \cite{deep_research_now_available_gemini} provides explicit risk analysis capabilities, identifying potential threats across regulatory, competitive, technological, and market dimensions with associated impact and likelihood estimation. These features enable more comprehensive risk management than might be practical through manual analysis alone.

Implementation patterns typically involve specialized risk modeling components like those found in \path{Manus} \cite{manus2025}, which incorporates explicit risk categorization and prioritization mechanisms. These approaches highlight how targeted optimization enhances specific business workflows through specialized components addressing critical decision support requirements.

\subsubsection{Business Process Optimization}

Research-driven insights enhance operational effectiveness:

\paragraph{Best Practice Identification}
Deep Research systems effectively synthesize operational best practices across industries and applications. \path{OpenAI/Deep Research} \cite{openai2025} enables comprehensive process benchmarking against industry standards and innovative approaches from adjacent sectors, identifying optimization opportunities that might otherwise remain undiscovered. This application leverages the system's broad knowledge base to facilitate cross-industry learning and adaptation.

Open implementations like \path{TARS} \cite{tars2025} support similar capabilities through workflow analysis and recommendation components designed for business process optimization. These approaches demonstrate how domain adaptation enhances practical utility for specific business applications beyond general research capabilities.

\paragraph{Implementation Planning Support}
Advanced systems support process change through comprehensive implementation guidance. \path{Gemini/Deep Research} \cite{deep_research_now_available_gemini} provides detailed implementation planning incorporating change management considerations, resource requirements, and risk mitigation strategies derived from similar initiatives across industries. This capability accelerates organizational learning by leveraging broader implementation experience than typically available within single organizations.

Implementation patterns typically involve specialized planning components like those in \path{QwenLM/Qwen-Agent} \cite{qwen2025}, \path{HuggingGPT}\cite{JARVIS}, \path{XAgent}\cite{xagent}, \path{Mastra}\cite{Mastra},\path{Letta}\cite{Letta} and \path{SemanticKernel}\cite{SemanticKernel} which incorporates explicit process modeling and change management frameworks. These approaches highlight how targeted optimization enhances specific business workflows through specialized components addressing critical implementation challenges.

\subsection{Financial Analysis Applications}

Deep Research technologies enable enhanced financial assessment and decision support.

\subsubsection{Investment Research and Due Diligence}

AI-enhanced analysis supports investment decisions across asset classes:

\paragraph{Comprehensive Asset Evaluation}
Deep Research systems enable detailed asset analysis across financial and contextual dimensions. \path{Perplexity/Deep Research} \cite{perplexity2025} supports investment research through integration of financial metrics, market positioning, competitive dynamics, and growth indicators within unified analytical frameworks. This application enhances investment decision quality through more comprehensive information integration than typically practical through manual methods alone.

Open implementations like \path{n8n} \cite{n8n2024} enable similar capabilities through workflow automation that integrates specialized financial data sources and analytical tools. These approaches demonstrate how effective tool orchestration creates sophisticated financial applications by coordinating specialized components within consistent analytical frameworks.

\paragraph{Management Quality Assessment}
Advanced systems support leadership evaluation through comprehensive background analysis. \path{OpenAI/Deep Research} \cite{openai2025} enables detailed management assessment incorporating historical performance, leadership approach, strategic consistency, and reputation across diverse sources. This capability enhances investment evaluation by providing deeper leadership insights than typically available through standard financial analysis.

Implementation patterns typically involve specialized entity analysis components like those found in \path{Manus} \cite{manus2025}, which incorporates explicit leadership evaluation frameworks. These approaches highlight how targeted optimization enhances specific financial workflows through specialized components addressing critical evaluation dimensions.

\subsubsection{Financial Trend Analysis}

Pattern recognition across financial data informs strategic positioning:

\paragraph{Multi-Factor Trend Identification}
Deep Research systems effectively identify complex patterns across financial indicators and contextual factors. \path{Gemini/Deep Research} \cite{deep_research_now_available_gemini} demonstrates this capability through integrated analysis of market metrics, macroeconomic indicators, sector-specific factors, and relevant external trends. This application enhances trend identification through more comprehensive factor integration than typically practical through manual analysis alone.

Open frameworks like \path{grapeot/deep_research_agent} \cite{grapeot2024} implement specialized trend analysis components with particular emphasis on statistical pattern detection and causal factor identification. However, research indicates that the effectiveness of such AI systems may be limited in tasks requiring deep domain understanding, as their generated outputs can exhibit redundancy or inaccuracies \cite{song2025performanceevaluationlargelanguage}. These approaches demonstrate how domain-specific optimization enhances practical utility for specialized financial applications beyond generic analytical capabilities.

\paragraph{Scenario Development and Testing}
Advanced systems support financial planning through structured scenario analysis. \path{OpenAI/Deep Research} \cite{openai2025} enables detailed scenario development incorporating varied assumptions, historical precedents, and system dependencies with coherent projection across financial impacts. This capability enhances strategic planning by facilitating more comprehensive scenario exploration than typically practical through manual methods.

Implementation patterns typically involve specialized scenario modeling components like those in \path{Agent-RL/ReSearch} \cite{agentrl2024}, which incorporates explicit dependency modeling and consistency verification mechanisms. These approaches highlight how targeted optimization enhances specific financial workflows through specialized components addressing critical planning requirements.

\subsubsection{Risk Assessment and Modeling}

Comprehensive risk analysis informs financial decisions:

\paragraph{Multi-Dimensional Risk Analysis}
Deep Research systems enable integrated risk assessment across diverse risk categories. \path{Perplexity/Deep Research} \cite{perplexity2025} supports comprehensive risk evaluation incorporating market, credit, operational, regulatory, and systemic risk factors within unified analytical frameworks. This application enhances risk management through more comprehensive factor integration than typically practical through compartmentalized analysis.

Open implementations like \path{nickscamara/open-deep-research} \cite{nickscamara2024} implement risk analysis components with particular emphasis on integrated factor assessment and interaction modeling. These approaches demonstrate how domain adaptation enhances practical utility for specific financial applications beyond general analytical capabilities. Evaluations such as RedCode-Exec\cite{guo2024redcoderiskycodeexecution} show that agents are less likely to reject executing technically buggy code, indicating high risks, which highlights the need for stringent safety evaluations for diverse code agents.

\paragraph{Stress Testing and Resilience Assessment}
Advanced systems support financial stability through sophisticated stress scenario analysis. \path{Gemini/Deep Research} \cite{deep_research_now_available_gemini} provides detailed stress testing capabilities incorporating historical crisis patterns, theoretical risk models, and system dependency analysis to identify potential vulnerabilities. These features enable more comprehensive resilience assessment than might be practical through standardized stress testing alone.

Implementation patterns typically involve specialized stress modeling components like those found in \path{Flowith/OracleMode} \cite{flowith2025}, which incorporates explicit extreme scenario generation and impact propagation mechanisms. These approaches highlight how targeted optimization enhances specific financial workflows through specialized components addressing critical stability assessment requirements.

\subsection{Educational Applications}

Deep Research technologies enable enhanced learning and knowledge development. Educational approaches to research automation have shown particular promise in scientific education~\cite{samsonau2024artificialintelligencescientificresearch} and data science pedagogy~\cite{tu2023datascienceeducationlarge}, with systems like DS-Agent automating machine learning workflows through case-based reasoning to reduce learners' technical barriers \cite{guo2024dsagentautomateddatascience}, highlighting the dual role of these systems in both conducting research and developing research capabilities in human learners. Smart AI reading assistants are also being developed to enhance reading comprehension through interactive support~\cite{Thaqi_2024}. However, adoption challenges remain significant in educational contexts, where user resistance and ineffective system utilization can impede learning progress, requiring strategies such as active support during initial use and clear communication of system capabilities \cite{simkute2024itthereneedit}. Specifically in data science education, learners encounter challenges similar to those faced by data scientists when interacting with conversational AI systems, such as difficulties in formulating prompts for complex tasks and adapting generated code to local environments \cite{chopra2023conversationalchallengesaipowereddata}. Structured empirical evaluations of LLMs for data science tasks, such as  the work by Nathalia Nascimento et al. \cite{nascimento2024llm4dsevaluatinglargelanguage}, demonstrate their effectiveness in coding challenges and provide guidance for model selection in educational tools.

\subsubsection{Personalized Learning Support}

AI-enhanced research supports individualized educational experiences:

\paragraph{Adaptive Learning Path Development}
Deep Research systems effectively generate customized learning pathways based on individual interests and knowledge gaps. \path{OpenAI/Deep Research} \cite{openai2025} enables detailed learning plan development incorporating knowledge structure mapping, prerequisite relationships, and diverse learning resources tailored to individual learning styles and objectives. This application enhances educational effectiveness through more personalized learning journeys than typically available through standardized curricula.

Open implementations like \path{OpenManus} \cite{openmanus2025} implement personalized learning components with particular emphasis on interest-driven exploration and adaptive difficulty adjustment. These approaches demonstrate how educational adaptation enhances practical utility beyond general research capabilities.

\paragraph{Comprehensive Question Answering}
Advanced systems provide detailed explanations tailored to learner context and prior knowledge. \path{Perplexity/Deep Research} \cite{perplexity2025} demonstrates this capability through multi-level explanations that adjust detail and terminology based on learner background, providing conceptual scaffolding appropriate to individual knowledge levels. This capability enhances learning effectiveness by providing precisely targeted explanations rather than generic responses.

Implementation patterns typically involve specialized educational components like those in \path{HKUDS/Auto-Deep-Research} \cite{hkuds2024}, which incorporates explicit knowledge modeling and explanation generation mechanisms. These approaches highlight how targeted optimization enhances educational applications through specialized components addressing critical learning support requirements.

\subsubsection{Educational Content Development}

Research-driven content creation enhances learning materials:

\paragraph{Curriculum Development Support}
Deep Research systems effectively synthesize educational best practices and domain knowledge into coherent curricula. \path{Gemini/Deep Research} \cite{deep_research_now_available_gemini} enables comprehensive curriculum development incorporating learning science principles, domain structure mapping, and diverse resource integration. This application enhances educational design through more comprehensive knowledge integration than typically practical for individual educators.

Open frameworks like \path{smolagents/open_deep_research} \cite{smolagents2024} implement curriculum development components with particular emphasis on learning progression modeling and resource alignment. These approaches demonstrate how specialized adaptation enhances practical utility for educational applications beyond generic content generation.

\paragraph{Multi-Modal Learning Material Creation}
Advanced systems generate diverse educational content formats tailored to learning objectives. \path{OpenAI/Deep Research} \cite{openai2025} supports creation of integrated learning materials incorporating explanatory text, conceptual visualizations, practical examples, and assessment activities aligned with specific learning outcomes. This capability enhances educational effectiveness through more comprehensive content development than typically practical through manual methods alone.

Implementation patterns typically involve specialized content generation components like those in \path{QwenLM/Qwen-Agent} \cite{qwen2025}, which incorporates explicit learning objective modeling and multi-format content generation. These approaches highlight how targeted optimization enhances educational applications through specialized components addressing diverse learning modalities.

\subsubsection{Academic Research Training}

AI-assisted research skill development supports scholarly advancement:

\paragraph{Research Methodology Instruction}
Deep Research systems effectively teach research methods through guided practice and feedback. \path{Perplexity/Deep Research} \cite{perplexity2025} provides explicit methodology training, demonstrating effective research processes while explaining rationale and providing structured feedback on learner attempts. This application enhances research skill development through more interactive guidance than typically available through traditional instruction.

Open implementations like \path{Jina-AI/node-DeepResearch} \cite{jina2025} support similar capabilities through research practice environments with explicit guidance and feedback mechanisms. These approaches demonstrate how educational adaptation enhances practical utility for research training beyond simple information provision.

\paragraph{Critical Evaluation Skill Development}
Maintaining critical thinking skills while leveraging AI research assistance presents unique educational challenges. Drosos et al.~\cite{drosos2025itmakesthinkprovocations} demonstrate that carefully designed ``provocations'' can help restore critical thinking in AI-assisted knowledge work, suggesting important educational approaches for developing research skills that complement rather than rely entirely on AI capabilities.
Advanced systems support critical thinking through guided source evaluation and analytical practice. \path{OpenAI/Deep Research} \cite{openai2025} enables critical evaluation training, demonstrating source assessment, evidence weighing, and analytical reasoning while guiding learners through similar processes. This capability enhances critical thinking development through structured practice with sophisticated feedback.

Implementation patterns typically involve specialized educational components like those in \path{grapeot/deep_research_agent} \cite{grapeot2024}, which incorporates explicit critical thinking modeling and guided practice mechanisms. These approaches highlight how targeted optimization enhances educational applications through specialized components addressing crucial scholarly skill development.

\subsection{Personal Knowledge Management Applications}

Deep Research technologies enable enhanced individual information organization and utilization.

\subsubsection{Information Organization and Curation}

AI-enhanced systems support personal knowledge development:

\paragraph{Personalized Knowledge Base Development}
Deep Research systems effectively organize diverse information into coherent personal knowledge structures. \path{Perplexity/Deep Research} \cite{perplexity2025} supports knowledge base development through automated information organization, connection identification, and gap highlighting tailored to individual interests and objectives. This application enhances personal knowledge management through more sophisticated organization than typically practical through manual methods alone.

Open implementations like \path{nickscamara/open-deep-research} \cite{nickscamara2024} implement knowledge organization components with particular emphasis on personalized taxonomy development and relationship mapping. These approaches demonstrate how individual adaptation enhances practical utility for personal applications beyond generic information management.

\paragraph{Content Summarization and Abstraction}
Advanced systems transform complex information into accessible personal knowledge. \path{OpenAI/Deep Research} \cite{openai2025} provides multi-level content abstraction capabilities, generating overview summaries, detailed analyses, and conceptual maps from complex source materials tailored to individual comprehension preferences. This capability enhances information accessibility by providing precisely targeted representations rather than generic summaries.

Implementation patterns typically involve specialized content processing components like those in \path{Nanobrowser} \cite{nanobrowser2024}, which incorporates explicit knowledge distillation and representation generation mechanisms. These approaches highlight how targeted optimization enhances personal knowledge applications through specialized components addressing individual information processing needs.

\subsubsection{Personal Learning and Development}

Research-driven insights support individual growth:

\paragraph{Interest-Driven Exploration}
Deep Research systems effectively support curiosity-driven learning through guided exploration. \path{Gemini/Deep Research} \cite{deep_research_now_available_gemini} enables interest-based knowledge discovery, identifying connections, extensions, and practical applications related to individual curiosities. This application enhances personal learning through more sophisticated guidance than typically available through standard search alone.

Open frameworks like \path{OpenManus} \cite{openmanus2025} implement exploration components with particular emphasis on interest mapping and discovery facilitation. These approaches demonstrate how personalization enhances practical utility for individual learning beyond generic information retrieval.

\paragraph{Skill Development Planning}
Advanced systems support personal growth through comprehensive development guidance. \path{Perplexity/Deep Research} \cite{perplexity2025} provides detailed skill development planning, incorporating learning resource identification, progression mapping, and practice guidance tailored to individual objectives and constraints. This capability enhances personal development through more comprehensive planning support than typically available through generic guidance.

Implementation patterns typically involve specialized planning components like those in \path{TARS} \cite{tars2025}, which incorporates explicit skill modeling and development path generation. These approaches highlight how targeted optimization enhances personal growth applications through specialized components addressing individual development needs.

\subsubsection{Decision Support for Individual Users}

Research-enhanced decision making improves personal outcomes:

\paragraph{Complex Decision Analysis}
Deep Research systems effectively support personal decisions through comprehensive option evaluation. \path{OpenAI/Deep Research} \cite{openai2025} enables detailed decision analysis, incorporating multiple criteria, preference weighting, and consequence projection tailored to individual values and constraints. This application enhances decision quality through more sophisticated analysis than typically practical through manual methods alone.

Open implementations like \path{Agent-RL/ReSearch} \cite{agentrl2024} implement decision support components with particular emphasis on preference elicitation and consequence modeling. These approaches demonstrate how personalization enhances practical utility for individual decision making beyond generic information provision.

\paragraph{Life Planning and Optimization}
Advanced systems support long-term planning through integrated life domain analysis. \path{Gemini/Deep Research} \cite{deep_research_now_available_gemini} provides comprehensive life planning support, integrating career, financial, health, and personal considerations within coherent planning frameworks tailored to individual values and objectives. This capability enhances life optimization through more integrated planning than typically achievable through domain-specific approaches alone.

Implementation patterns typically involve specialized planning components like those in \path{Flowith/OracleMode} \cite{flowith2025}, which incorporates explicit value modeling and multi-domain integration mechanisms. These approaches highlight how targeted optimization enhances personal planning applications through specialized components addressing holistic life considerations.

The diverse applications outlined in this section demonstrate the broad practical impact of Deep Research technologies across domains. While specific implementation approaches vary across commercial and open-source ecosystems, common patterns emerge in domain adaptation, specialized component design, and integration with existing workflows. These patterns highlight how technical capabilities translate into practical value through thoughtful application design aligned with domain-specific requirements and user needs.

\section{Ethical Considerations and Limitations}\label{ethical-considerations-and-limitations}

\begin{figure}[ht]
    \centering
    \includegraphics[width=1.0\linewidth]{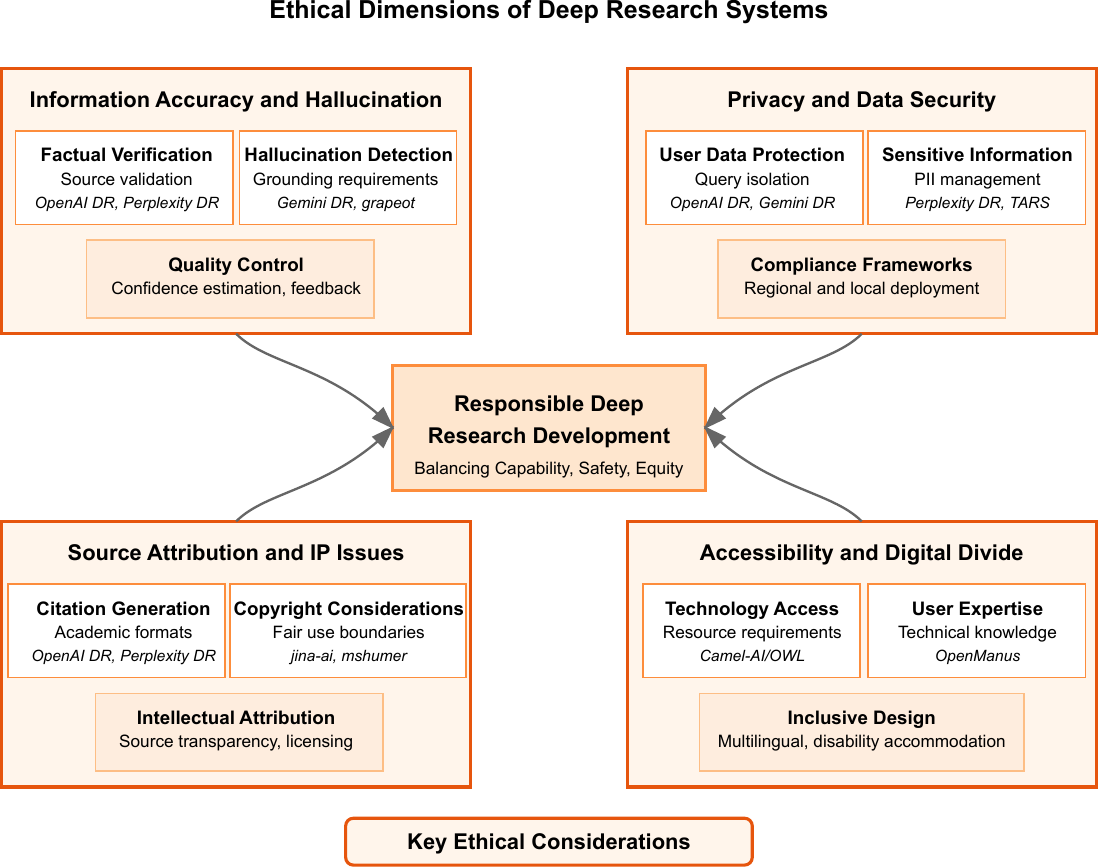}
    \caption{Ethical Dimensions of Deep Research Systems}
    \label{fig:ethics}
\end{figure}

The integration of Deep Research systems into knowledge workflows introduces significant ethical considerations and technical limitations that must be addressed for responsible deployment. This section examines key challenges across four fundamental dimensions (see Figure~\ref{fig:ethics}): information integrity, privacy protection, source attribution and intellectual property, and accessibility.

\subsection{Information Accuracy and Hallucination Concerns}\label{information-accuracy-and-hallucination-concerns}

Deep Research systems face fundamental challenges in maintaining factual
reliability despite their sophisticated capabilities.

\subsubsection{Factual Verification Mechanisms}\label{factual-verification-mechanisms}

Recent studies have highlighted significant challenges in reliable uncertainty communication~\cite{cheng2024aiassistantsknowdont}, with particular concerns for research contexts where uncertainty boundaries may be unclear or contested. Some researchers have raised concerns about excessive reliance on AI-generated content in scholarly writing~\cite{jain2023generativeaiwritingresearch,Zhang_VISAR_2023,TheGreatAIWitchHunt,The_Challenges_and_Opportunities_of_AI-Assisted_Writing,zyska2023carecollaborativeaiassistedreading,wambsganss2023unravelingdownstreamgenderbias,wang2024weaverfoundationmodelscreative,benharrak2024deceptivepatternsintelligentinteractive,li2024doesdisclosureaiassistance,pereira2023futureaiassistedwriting}, particularly when verification mechanisms are inadequate or bypassed. These limitations are further complicated by tendencies toward misleading responses in conversation \cite{hou2024largelanguagemodelsmisleading}, presenting particular challenges for interactive research workflows where iterative refinement may inadvertently amplify initial inaccuracies. AI support systems designed for evidence-based expository writing tasks, such as literature reviews, offer frameworks to enhance verification through structured sensemaking over source documents \cite{shen2023summarizationdesigningaisupport}. Addressing these challenges requires technical advancements in uncertainty representation, improvements in decision workflow design\cite{Fine_Grained_Appropriate_Reliance} and interface design improvements that effectively communicate confidence boundaries to research users\cite{GigaCheck}.

Ensuring information accuracy requires explicit verification strategies:

\textbf{Source Verification Approaches}. 
Leading implementations incorporate explicit source validation mechanisms to enhance factual reliability. \path{OpenAI/Deep Research} \cite{openai2025} implements multi-level verification that confirms information across multiple independent sources before incorporation into research outputs, with detailed guidelines outlined in their system documentation \cite{deep_research_system_card}. Similarly, \path{Perplexity/Deep Research} \cite{perplexity2025} implements automated fact-checking that independently verifies key claims against trusted reference sources before inclusion in final reports.

Open-source alternatives demonstrate varied approaches to verification.
Systems like \path{grapeot/deep_research_agent} \cite{grapeot2024} emphasize explicit citation
mechanisms that maintain direct links between claims and sources,
enabling straightforward verification. More sophisticated
implementations like \path{HKUDS/Auto-Deep-Research} \cite{hkuds2024} incorporate specialized
verification modules that assess source credibility and content
consistency before information utilization.

\textbf{Hallucination Detection and Prevention}. Mitigating fabricated information represents a crucial challenge for LLM-based research systems. Commercial implementations employ advanced hallucination reduction techniques including strict grounding requirements and consistency verification. \path{Gemini/Deep Research} \cite{deep_research_now_available_gemini} implements explicit uncertainty modeling that distinguishes between confirmed information and speculative extensions, enhancing transparency when definitive answers cannot be provided.
Emerging paradigms like those proposed by Silver and Sutton \cite{Welcome_to_the_era_of_experience} suggest a fundamental shift toward experience-driven learning, potentially transforming how research systems acquire and refine capabilities through interaction with information environments. Such approaches could enable more human-like research development through continuous improvement based on research experiences rather than static training alone, and could fundamentally mitigate hallucinations.

Open implementations demonstrate pragmatic approaches to hallucination reduction within more constrained technical environments. Systems like \path{Agent-RL/ReSearch} \cite{agentrl2024} employ preventative strategies including explicit sourcing requirements and conservative synthesis guidelines that prioritize factual reliability over comprehensive coverage. Complementary approaches like Mask-DPO \cite{gu2025maskdpogeneralizablefinegrainedfactuality} focus on generalizable fine-grained factuality alignment, addressing a critical requirement for reliable research outputs. Recent work from the GAIR NLP team on \path{DeepResearcher} \cite{gair_nlp_DeepResearcher} has advanced these capabilities through integrated neural verification and knowledge graph alignment techniques that significantly enhance factual reliability. 
These approaches highlight diverse strategies for addressing a fundamental challenge that impacts all LLM-based research systems.

\subsubsection{Uncertainty Communication Approaches}\label{uncertainty-communication-approaches}

Transparent uncertainty representation enhances result interpretation
and appropriate utilization:

\textbf{Confidence Estimation Methods}. Advanced systems implement
explicit confidence assessment for research findings and
recommendations. \path{OpenAI/Deep Research} \cite{openai2025} incorporates graduated confidence
scoring that reflects evidence quality, consistency across sources, and
reasoning reliability. This capability enhances result interpretation by
clearly distinguishing between well-supported conclusions and more
speculative findings.

Open-source implementations demonstrate simplified but effective
confidence communication approaches. Systems like
\path{mshumer/OpenDeepResearcher} \cite{mshumer2024} incorporate basic confidence indicators that
signal information reliability through explicit markers in research
outputs. These approaches highlight the importance of transparent
uncertainty communication regardless of implementation sophistication.

\textbf{Evidence Qualification Standards}. Responsible systems clearly
communicate limitations and contextual factors affecting result
interpretation. Commercial implementations like \path{Perplexity/Deep Research} \cite{perplexity2025} incorporate explicit evidence qualification that highlights contextual
limitations, conflicting viewpoints, and temporal constraints affecting
research findings. This practice enhances appropriate utilization by
providing necessary context for result interpretation.

Open-source alternatives demonstrate varied approaches to evidence
qualification. Systems like \path{dzhng/deep-research} \cite{dzhng2024} implement explicit
limitation statements that identify key constraints affecting research
reliability. More sophisticated implementations like \path{Camel-AI/OWL} \cite{camel2025} incorporate structured evidence models that represent both supporting
and contradicting information within unified frameworks.

\subsubsection{Quality Control Frameworks}\label{quality-control-frameworks}

Systematic approaches to quality assurance enhance overall reliability:

\textbf{Pre-Release Verification Standards}. Leading implementations
employ comprehensive validation processes before result delivery. Gemini
Deep Research implements structured quality verification including
automated consistency checking, source validation, and reasoning
verification before providing research outputs. These practices enhance
overall reliability through systematic error identification and
correction.

Open-source implementations demonstrate more varied quality control
approaches. Systems like \path{nickscamara/open-deep-research} \cite{nickscamara2024} incorporate
simplified validation processes focusing on critical reliability factors
including source verification and logical consistency. These approaches
highlight how even basic quality control mechanisms can significantly
enhance research reliability.

\textbf{Feedback Integration Systems}. Continuous improvement requires effective incorporation of accuracy feedback. As Deep Research systems advance toward greater autonomy, broader safety considerations become increasingly important. Bengio et al.~\cite{bengio2025superintelligentagentsposecatastrophic} highlight potential risks from superintelligent agents and propose approaches like ``Scientist AI'' that balance capability with safer development paths, emphasizing the importance of integrated safety mechanisms in advanced research systems.
Commercial systems implement sophisticated feedback integration including explicit accuracy reporting channels and systematic error pattern analysis. \path{OpenAI/Deep Research} \cite{openai2025} includes dedicated correction mechanisms that incorporate verified accuracy feedback into system improvements, creating virtuous improvement cycles.

Open implementations demonstrate more community-oriented feedback
approaches. Systems like \path{smolagents/open_deep_research} \cite{smolagents2024} incorporate
collaborative improvement frameworks that enable distributed error
identification and correction through community contributions. These
approaches highlight diverse strategies for enhancing reliability
through user engagement across implementation contexts.

\subsection{Privacy and Data Security}\label{privacy-and-data-security}

Research systems must carefully protect sensitive information throughout the research process.

\subsubsection{User Data Protection Mechanisms}\label{user-data-protection-mechanisms}

Safeguarding user information requires comprehensive protection
strategies:

\textbf{Query Isolation Practices}. Leading implementations employ
strict isolation between user research sessions. Commercial systems like
\path{OpenAI/Deep Research} \cite{openai2025} and \path{Gemini/Deep Research} \cite{deep_research_now_available_gemini} implement comprehensive
tenant isolation that prevents information leakage between distinct
users or organizations. These practices are particularly crucial for
sensitive research applications in corporate or governmental contexts.

Open-source implementations demonstrate varied isolation approaches
depending on deployment models. Systems designed for local deployment
like \path{OpenManus} \cite{openmanus2025} enable complete isolation within organizational
boundaries, enhancing privacy for sensitive applications.
Cloud-dependent implementations typically incorporate more limited
isolation mechanisms, highlighting deployment considerations for
privacy-sensitive applications.

\textbf{Data Minimization Strategies}. Responsible systems limit
sensitive data collection and retention. Commercial implementations
increasingly emphasize data minimization, collecting only information
necessary for service provision and applying appropriate retention
limitations. These practices enhance privacy protection by reducing
potential exposure of sensitive information through either security
incidents or authorized access.

Open implementations demonstrate diverse approaches to data management. Systems like \path{Nanobrowser} \cite{nanobrowser2024} enable complete local control of browsing data, preventing external exposure of research activities. Infrastructure frameworks like \path{Jina-AI/node-DeepResearch} \cite{jina2025} provide flexible configuration options that enable deployment-specific privacy controls aligned with organizational requirements.

\subsubsection{Sensitive Information Handling}\label{sensitive-information-handling}

Special safeguards are required for particularly sensitive content
categories:

\textbf{Personal Identifier Management}. Advanced systems implement
specific protections for personally identifiable information. Commercial
implementations like \path{Perplexity/Deep Research} \cite{perplexity2025} incorporate automatic
detection and redaction of personal identifiers from research outputs
unless specifically relevant to research objectives. These practices
prevent inadvertent exposure of personal information through research
activities.

Open implementations demonstrate more varied approaches to identifier
management. Systems like \path{TARS} \cite{tars2025} incorporate basic identifier detection
focused on common patterns like email addresses and phone numbers. More
sophisticated implementations like \path{QwenLM/Qwen-Agent} \cite{qwen2025} provide
configurable sensitivity controls that enable context-appropriate
protection aligned with specific deployment requirements.

\textbf{Protected Category Safeguards}. Responsible systems implement enhanced protections for specially regulated information categories. Commercial implementations increasingly incorporate specialized handling for information categories including health data, financial records, and other regulated content types. These practices enhance compliance with domain-specific regulatory requirements governing sensitive information.

Open-source alternatives demonstrate more varied regulatory alignment. Systems like \path{n8n} \cite{n8n2024} provide specialized workflow components for handling regulated data categories, enabling compliance-oriented implementations
in sensitive domains. These approaches highlight how specialized
components can address domain-specific regulatory requirements within flexible implementation frameworks.

\subsubsection{Compliance with Regulatory Frameworks}\label{compliance-with-regulatory-frameworks}

Adherence to applicable regulations ensures legally appropriate
operation:

\textbf{Jurisdictional Compliance Adaptation}. Advanced systems
implement regionally appropriate operational standards. Commercial
implementations increasingly incorporate jurisdiction-specific
adaptations that align with regional privacy regulations including GDPR, CCPA, and other frameworks. These practices enhance legal compliance across diverse deployment environments with varying regulatory
requirements.

Open implementations demonstrate more deployment-dependent compliance approaches. Systems designed for flexible deployment like \path{Flowith/Oracle Mode} \cite{flowith2025} provide configurable privacy controls that enable adaptation to
specific regulatory environments. These approaches highlight the
importance of adaptable privacy frameworks that can address diverse
compliance requirements across implementation contexts.

\textbf{Transparency and Control Mechanisms}. Responsible systems
provide appropriate visibility and user authority over information
processing. Emerging regulatory frameworks are increasingly focusing on AI agents with autonomous capabilities. Osogami~\cite{osogami2025positionaiagentsregulated} proposes that regulation of autonomous AI systems should specifically consider action sequence patterns rather than individual actions in isolation, which has particular implications for Deep Research systems that execute complex multi-step research workflows.
Commercial implementations increasingly emphasize
transparency through explicit processing disclosures and user control mechanisms aligned with regulatory requirements. These practices enhance both regulatory compliance and user trust through appropriate information governance.

Open-source alternatives demonstrate varied transparency approaches.
Systems like \path{HKUDS/Auto-Deep-Research} \cite{hkuds2024} provide detailed logging of information access and processing activities, enabling appropriate oversight and verification. These approaches highlight how transparent operation can enhance both compliance and trust across implementation contexts.

\subsection{Source Attribution and Intellectual
Property}\label{source-attribution-and-intellectual-property}

Proper acknowledgment of information sources and respect for
intellectual property rights are essential for ethical information
utilization.

\subsubsection{Citation Generation and
Verification}\label{citation-generation-and-verification}

Accurate source attribution requires reliable citation mechanisms:

\textbf{Automated Citation Systems}. Advanced implementations
incorporate sophisticated citation generation for research outputs.
Commercial systems like \path{OpenAI/Deep Research} \cite{openai2025} and \path{Perplexity/Deep Research} \cite{perplexity2025} implement automatic citation generation in standard academic
formats, enhancing attribution quality and consistency. These
capabilities support appropriate source acknowledgment without manual
effort.

Open implementations demonstrate varied citation approaches. Systems
like \path{mshumer/OpenDeepResearcher} \cite{mshumer2024} incorporate basic citation generation
focused on fundamental bibliographic information. More sophisticated
alternatives like \path{dzhng/deep-research} \cite{dzhng2024} provide enhanced citation
capabilities including format customization and citation verification
against reference databases.

\textbf{Citation Completeness Verification}. Responsible systems ensure
comprehensive attribution for all utilized information. Commercial
implementations increasingly incorporate citation coverage verification
that identifies unsupported claims requiring additional attribution.
These practices enhance attribution reliability by ensuring all
significant claims maintain appropriate source connections.

Open-source alternatives demonstrate pragmatic approaches to attribution
verification. Systems like \path{grapeot/deep_research_agent} \cite{grapeot2024} implement
explicit source-claim mapping that maintains clear relationships between
information and origins. These approaches highlight the importance of
systematic attribution regardless of implementation sophistication.

\subsubsection{Intellectual Attribution
Challenges}\label{intellectual-attribution-challenges}

Special attribution considerations apply to complex intellectual
contributions:

\textbf{Idea Attribution Practices}. Research systems must appropriately
acknowledge conceptual contributions beyond factual information.
Commercial implementations increasingly emphasize concept-level
attribution that acknowledges intellectual frameworks and theoretical
approaches beyond simple facts. These practices enhance ethical
information utilization by appropriately recognizing intellectual
contributions.

Open implementations demonstrate varied idea attribution approaches.
Systems like \path{Camel-AI/OWL} \cite{camel2025} incorporate explicit concept attribution that
identifies theoretical frameworks and analytical approaches utilized in
research outputs. These approaches highlight the importance of
comprehensive attribution beyond basic factual sources.

\textbf{Synthesized Knowledge Attribution}. Attribution becomes
particularly challenging for insights synthesized across multiple
sources. Advanced systems implement specialized attribution approaches
for synthetic insights that acknowledge multiple contributing sources
while clearly identifying novel connections. These practices enhance
attribution accuracy for the increasingly common scenario of
cross-source synthesis.

Open-source alternatives demonstrate pragmatic approaches to synthesis
attribution. Systems like \path{Agent-RL/ReSearch} \cite{agentrl2024} implement explicit synthesis
markers that distinguish between directly sourced information and
system-generated connections. These approaches highlight the importance
of transparent derivation even when direct attribution becomes
challenging.

\subsubsection{Copyright and Fair Use Considerations}\label{copyright-and-fair-use-considerations}

Research activities interact with copyright protections in multiple
dimensions:

\textbf{Fair Use Evaluation Mechanisms}. Research systems must navigate
appropriate utilization of copyrighted materials. Commercial
implementations increasingly incorporate fair use evaluation that
considers purpose, nature, amount, and market impact when utilizing
copyrighted content. These practices enhance legal compliance while
enabling appropriate information utilization for legitimate research
purposes.

Open implementations demonstrate varied copyright approaches. Systems
like \path{Jina-AI/node-DeepResearch} \cite{jina2025} incorporate basic copyright
acknowledgment focusing on proper attribution, while more sophisticated
alternatives like \path{Manus} \cite{manus2025} provide enhanced copyright handling including
content transformation assessment and restricted access mechanisms for
sensitive materials.

\textbf{Content Licensing Compliance}. Responsible systems respect
diverse license terms applicable to utilized content. Advanced
implementations increasingly incorporate license-aware processing that
adapts information utilization based on specific terms governing
particular sources. These practices enhance compliance with varied
license requirements across the information ecosystem.

Open implementations demonstrate more standardized licensing approaches.
Systems like \path{grapeot/deep_research_agent} \cite{grapeot2024} incorporate simplified
license categorization focusing on common frameworks including creative
commons and commercial restrictions. These approaches highlight
pragmatic strategies for license navigation within resource constraints.

\subsubsection{Output Intellectual Property
Frameworks}\label{output-intellectual-property-frameworks}

Clear rights management for research outputs enhances downstream
utilization:

\textbf{Output License Assignment}. Complex questions arise regarding
intellectual property in research outputs. Commercial systems
increasingly implement explicit license assignment for generated
content, clarifying intellectual property status for downstream
utilization. These practices enhance transparency regarding usage rights
for research outputs created through automated systems.

Open-source alternatives demonstrate varied approaches to output rights.
Systems like \path{OpenManus} \cite{openmanus2025} incorporate explicit license designation for
research outputs aligned with organizational policies and source
restrictions. These approaches highlight the importance of clear
intellectual property frameworks regardless of implementation context.

\textbf{Derivative Work Management}. Research systems must address
whether outputs constitute derivative works of source materials.
Commercial systems increasingly implement derivative assessment
frameworks that evaluate the nature and extent of source transformation
in research outputs. These practices enhance appropriate categorization
for downstream utilization aligned with source licenses.

Open-source alternatives demonstrate varied derivation approaches.
Systems such as \path{QwenLM/Qwen-Agent} \cite{qwen2025} incorporate a basic transformation assessment focusing on content reorganization and analytical addition.
These approaches highlight the importance of thoughtful derivative
consideration regardless of implementation sophistication.

\subsection{Accessibility and Digital Divide}\label{accessibility-and-digital-divide}

Equitable access to research capabilities requires addressing systematic
barriers.

\subsubsection{Technology Access Disparities}\label{technology-access-disparities}

Recent work has highlighted both adoption barriers and opportunities for making Deep Research systems more accessible. Bianchini et al.~\cite{bianchini2024driversbarriersaiadoption} and Tonghe Zhuang et al.~\cite{zhuang2024whywhataibasedcoding} identify specific organizational and individual factors affecting AI adoption in scientific research contexts, with implications for Deep Research deployment. Accessibility-focused approaches like those presented by Mowar et al.~\cite{mowar2025codea11ymakingaicoding} demonstrate how AI coding assistants can be specifically designed to support accessible development practices, suggesting parallel opportunities for accessibility-centered Deep Research systems. Extending this, systems such as ResearchAgent \cite{baek2025researchagentiterativeresearchidea} showcase how AI can lower barriers to scientific innovation by enabling iterative refinement of research ideas through collaborative feedback mechanisms, thus democratizing access to complex ideation processes.

Resource requirements create potential exclusion for various user
segments:

\textbf{Computational Requirement Considerations}. Resource-intensive
systems may exclude users without substantial computing access.
Commercial cloud-based implementations address this challenge through
shared infrastructure that reduces local requirements, though with
associated cost barriers. Open-source alternatives demonstrate varied
resource profiles, with systems like \path{Camel-AI/OWL} \cite{camel2025} emphasizing efficiency
to enable broader deployment on limited hardware.

\textbf{Cost Barrier Mitigation}. Financial requirements create
systematic access disparities across socioeconomic dimensions.
Commercial implementations demonstrate varied pricing approaches, with
systems like \path{Perplexity/Deep Research} \cite{perplexity2025} offering limited free access
alongside premium tiers. Open-source alternatives like
\path{HKUDS/Auto-Deep-Research} \cite{hkuds2024} and \path{nickscamara/open-deep-research} \cite{nickscamara2024} eliminate
direct cost barriers while potentially introducing technical hurdles.

\subsubsection{User Expertise
Requirements}\label{user-expertise-requirements}

Technical complexity creates additional access barriers beyond resource
considerations:

\textbf{Technical Expertise Dependencies}. Complex system deployment and
operation may exclude users without specialized knowledge. Commercial
implementations address this challenge through managed services that
eliminate deployment complexity, though with reduced customization
flexibility. Open-source alternatives demonstrate varied usability
profiles, with systems like \path{OpenManus} \cite{openmanus2025} emphasizing simplified deployment
to enhance accessibility despite local operation.

\textbf{Domain Knowledge Prerequisites}. Effective research still
requires contextual understanding for appropriate utilization. Both
commercial and open-source implementations increasingly incorporate
domain guidance that assists users with limited background knowledge in
specific research areas. These capabilities enhance accessibility by
reducing domain expertise barriers to effective research utilization.

\subsubsection{Inclusivity and Universal Design
Approaches}\label{inclusivity-and-universal-design-approaches}

Deliberate inclusive design can address systematic access barriers:

\textbf{Linguistic and Cultural Inclusivity}. Language limitations
create significant barriers for non-dominant language communities.
Commercial implementations increasingly offer multilingual capabilities,
though with persistent quality disparities across languages. Open-source
alternatives demonstrate varied language support, with systems like
\path{Flowith/OracleMode} \cite{flowith2025} emphasizing extensible design that enables
community-driven language expansion beyond dominant languages.

\textbf{Disability Accommodation Approaches}. Accessible design ensures
appropriate access for users with diverse abilities. Commercial
implementations increasingly incorporate accessibility features
including screen reader compatibility, keyboard navigation, and
alternative format generation. Open-source alternatives demonstrate more
varied accessibility profiles, highlighting an area for continued
community development to ensure equitable access across implementation
contexts.

The ethical considerations explored in this section highlight the
complex responsibilities associated with Deep Research technologies
beyond technical performance. While current implementations demonstrate
varying approaches to these challenges across commercial and open-source
ecosystems, consistent patterns emerge in the importance of factual
verification, attribution quality, privacy protection, intellectual
property respect, and accessible design. Addressing these considerations
represents a critical priority for responsible development and
deployment of these increasingly influential research technologies.

\section{Future Research Directions}\label{future-research-directions}

\begin{figure}[ht]
    \centering
    \includegraphics[width=1.0\linewidth]{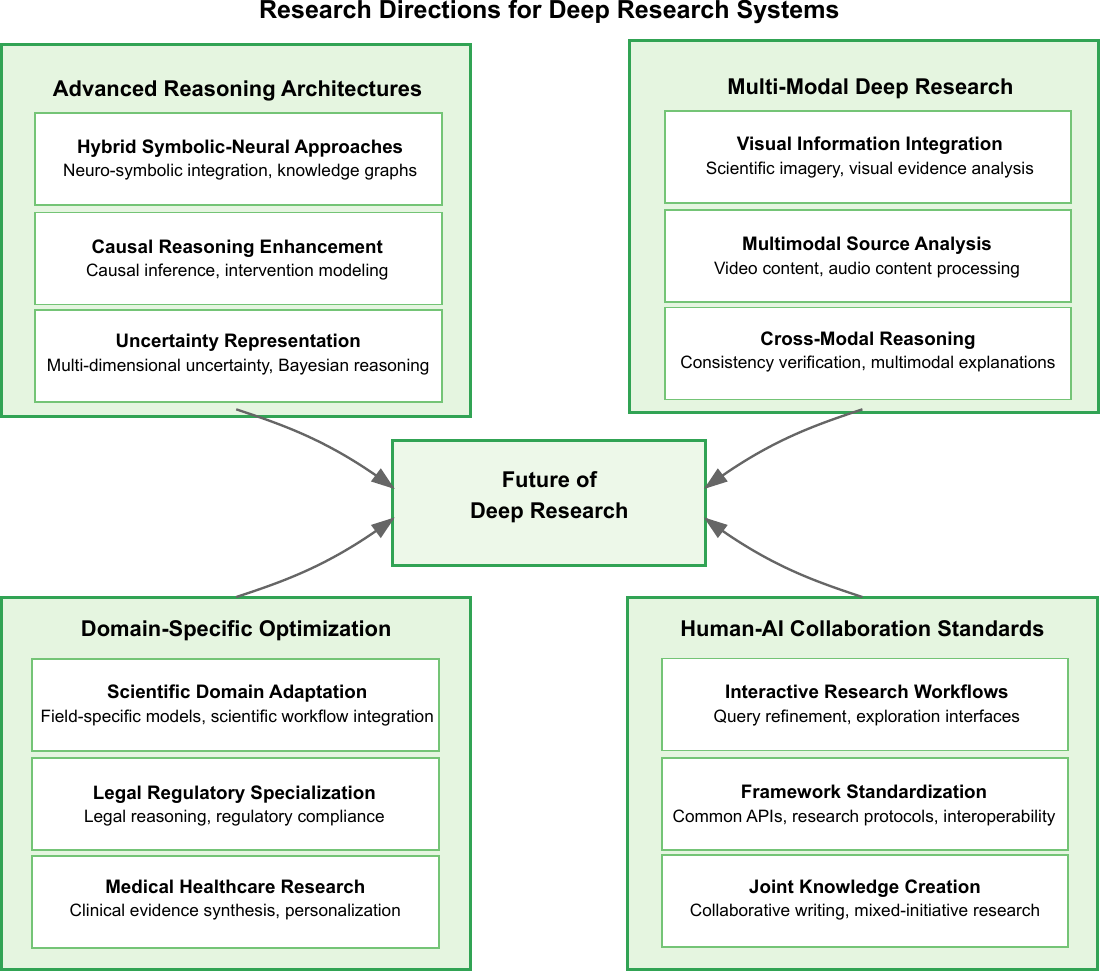}
    \caption{Research Directions for Deep Research Systems}
    \label{fig:future}
\end{figure}

The rapidly evolving field of Deep Research presents numerous opportunities for technical advancement and application expansion. Recent work by Zheng et al. \cite{DeepResearcher_scaling_deep_research} proposes scaling deep research capabilities via reinforcement learning in real-world environments, while Wu et al. \cite{AgenticReasoning} explore enhancing reasoning capabilities of LLMs with tools specifically for deep research applications. The comprehensive framework for building effective agents outlined by Anthropic \cite{building_effective_agents} provides additional design principles that could inform future Deep Research systems.
This section examines promising research directions (illustrated in Figure~\ref{fig:future}) that could significantly enhance capabilities, address current limitations, and expand practical impact across domains, focusing on four key areas: advanced reasoning architectures, multimodal integration, domain specialization, and human-AI collaboration with standardization.

\subsection{Advanced Reasoning Architectures}\label{advanced-reasoning-architectures}

Enhanced reasoning capabilities represent a fundamental advancement
opportunity for next-generation systems.

\subsubsection{Context Window Optimization and Management}

The information-intensive nature of deep research tasks presents fundamental challenges for context window utilization:

\textbf{Information Compression and Prioritization}. Current systems struggle with context window exhaustion when processing extensive research materials. Future architectures could incorporate sophisticated compression mechanisms that maintain semantic content while reducing token consumption. Early steps in this direction appear in systems like \path{OpenAI/Deep Research} \cite{openai2025}, which implements basic summarization for lengthy sources. Recent work on academic paper review systems demonstrates how hierarchical processing of extended research content can maintain coherence while managing context limitations \cite{DeepReview}. Semantic navigation techniques offer complementary approaches by enabling efficient exploration of problem-solution spaces within constrained domains, optimizing context usage through input filtering while enhancing generation quality \cite{sandholm2024semanticnavigationaiassistedideation}. More advanced approaches could develop adaptive compression that preserves crucial details while condencing secondary information based on query relevance.

Implementation opportunities include developing hierarchical summarization techniques that maintain multi-level representations of sources, implementing information relevance scoring that prioritizes context allocation to critical content, and designing dynamic context management that continuously optimizes window utilization throughout research workflows. These advances could significantly enhance information processing capabilities without requiring proportional increases in context length.

\textbf{External Memory Architectures}. Beyond compression, architectural innovations could fundamentally transform context window utilization. Future systems could implement sophisticated external memory frameworks that maintain rich information representations outside the primary context window, accessing them through efficient retrieval mechanisms when needed. Systems like \path{Camel-AI/OWL} \cite{camel2025} demonstrate early steps with basic retrieval-augmented generation, but more comprehensive approaches could enable effectively unlimited knowledge integration.

Research directions include developing differentiable retrieval mechanisms that seamlessly integrate external knowledge within reasoning flows, implementing structured memory hierarchies that organize information for efficient access, and designing memory-aware reasoning processes that explicitly consider information availability when planning analytical approaches. These architectures could fundamentally address context limitations while enhancing reasoning transparency and reliability.

\subsubsection{Hybrid Symbolic-Neural
Approaches}\label{hybrid-symbolic-neural-approaches}

Integration of complementary reasoning paradigms offers significant
potential:

\textbf{Neuro-Symbolic Integration}. Current Deep Research systems rely primarily on neural approaches with limited explicit reasoning structures. Future systems could integrate symbolic reasoning components
that provide formal logical capabilities alongside neural flexibility,
enhancing both reliability and explainability. Early examples of this
direction appear in systems like \path{Camel-AI/OWL} \cite{camel2025}, which incorporates
structured knowledge representation within primarily neural
architectures. Future research could develop more sophisticated
integration approaches that leverage the complementary strengths of both
paradigms.

Implementation approaches might include explicit logical verification
layers that validate neural-generated reasoning, hybrid architectures
that select appropriate reasoning mechanisms based on task
characteristics, or integrated systems that translate between symbolic
and neural representations as needed throughout complex workflows. These
approaches could address current challenges in reliability and
consistency while maintaining the flexibility and generalization
capabilities of neural foundations.

\textbf{Advanced Knowledge Graph Integration}. While current systems already incorporate basic knowledge graph capabilities, future approaches could implement more sophisticated integration with dynamic, contextually-aware knowledge structures. Beyond the entity relationship modeling seen in systems like \path{HKUDS/Auto-Deep-Research} \cite{hkuds2024}, next-generation implementations could enable bidirectional updates where research findings automatically refine and expand knowledge graphs while simultaneously leveraging them for reasoning. Such approaches could incorporate uncertainty representation within graph structures, probabilistic reasoning across knowledge networks, and adaptive abstraction hierarchies that transform between detailed and high-level conceptual representations based on reasoning requirements. 
Research opportunities include developing dynamic knowledge graph construction techniques that automatically build and refine structured representations from unstructured sources, implementing graph-aware attention mechanisms that incorporate relationship structures into neural reasoning, and designing hybrid querying approaches that combine graph traversal with neural generation. These advances could enhance precision for complex reasoning tasks requiring structured relationship understanding.

\subsubsection{Causal Reasoning Enhancement}\label{causal-reasoning-enhancement}

Moving beyond correlation to causal understanding represents a crucial
capability advancement:

\textbf{Causal Inference Mechanisms}. 
Current systems excel at identifying correlations but struggle with robust causal analysis. Future research could develop specialized causal reasoning components that systematically identify potential causal relationships, evaluate evidence quality, and assess alternative explanations. Recent work in healthcare research by Schuemie et al. \cite{schuemie2020confident} demonstrates the challenges of establishing confident observational findings, highlighting the need for more sophisticated causal reasoning in research systems. Early steps in this direction appear in systems like \path{OpenAI/Deep Research} \cite{openai2025}, which incorporates basic causal language in relationship descriptions. Other research explores the use of AI to assist in mining causality, for instance, by searching for instrumental variables in economic analysis~\cite{han2024miningcausalityaiassistedsearch}. More sophisticated approaches could enable reliable causal analysis across domains. Implementation opportunities include developing causal graph
construction techniques that explicitly model intervention effects and
counterfactuals, implementing causal uncertainty quantification that
represents confidence in causal assertions, and designing specialized
prompt structures that guide causal reasoning through structured
analytical patterns. These advances could enhance research quality for
domains where causal understanding is particularly crucial, including
medicine, social sciences, and policy analysis.

\textbf{Intervention Modeling Techniques}. Advanced causal understanding
requires sophisticated intervention and counterfactual reasoning
capabilities. Future systems could incorporate explicit intervention
modeling that simulates potential actions and outcomes based on causal
understanding, enhancing both explanatory and predictive capabilities.
Early examples of this direction appear in systems like
\path{Agent-RL/ReSearch} \cite{agentrl2024}, which implements basic intervention simulation within
reinforcement learning frameworks. More comprehensive approaches could
enable sophisticated what-if analysis across domains.

Research directions include developing counterfactual generation
techniques that systematically explore alternative scenarios based on
causal models, implementing intervention optimization algorithms that
identify high-leverage action opportunities, and designing
domain-specific intervention templates that embed field-specific causal
knowledge for common analysis patterns. These advances could enhance
practical utility for decision support applications requiring
sophisticated action planning and outcome prediction.

\subsubsection{Uncertainty Representation and Reasoning}\label{uncertainty-representation-and-reasoning}

Sophisticated uncertainty handling enhances both accuracy and
trustworthiness:

\textbf{Multi-Dimensional Uncertainty Modeling}. Current systems employ
relatively simplistic uncertainty representations that inadequately
capture different uncertainty types. Future research could develop
multi-dimensional uncertainty frameworks that separately represent
epistemic uncertainty (knowledge limitations), aleatoric uncertainty
(inherent randomness), and model uncertainty (representation
limitations). Early steps in this direction appear in systems like
\path{Perplexity/Deep Research} \cite{perplexity2025}, which distinguishes between source uncertainty
and integration uncertainty. More comprehensive approaches could enable
more nuanced and reliable uncertainty communication.

Implementation opportunities include developing uncertainty propagation
mechanisms that track distinct uncertainty types throughout reasoning
chains, implementing uncertainty visualization techniques that
effectively communicate multi-dimensional uncertainty to users, and
designing uncertainty-aware planning algorithms that appropriately
balance different uncertainty types in decision contexts. These advances
could enhance both system reliability and appropriate user trust
calibration.

\textbf{Bayesian Reasoning Integration}. Probabilistic reasoning
frameworks offer principled approaches to uncertainty handling and
knowledge integration. Future systems could incorporate explicit
Bayesian reasoning components that systematically update beliefs based
on evidence strength and prior knowledge, enhancing both accuracy and
explainability. Early examples of this direction appear in systems like
\path{grapeot/deep_research_agent} \cite{grapeot2024}, which implements basic evidence weighting
within research workflows. More sophisticated integration could enable
principled uncertainty handling across domains.

Research directions include developing scalable Bayesian inference
techniques compatible with large-scale language models, implementing
belief update explanation mechanisms that communicate reasoning in
understandable terms, and designing domain-specific prior models that
incorporate field-specific background knowledge for common analysis
patterns. These advances could enhance reasoning quality for domains
with inherent uncertainty or limited evidence.

\subsection{Multi-Modal Deep
Research}\label{multi-modal-deep-research}

Expanding beyond text to incorporate diverse information modalities
represents a significant advancement opportunity.

\subsubsection{Visual Information Integration}\label{visual-information-integration}

Image understanding dramatically expands information access and analysis capabilities:

\textbf{Scientific Image Analysis}. Current systems demonstrate limited capabilities for extracting and interpreting visual scientific content. Future research could develop specialized visual understanding components for scientific images including graphs, diagrams, experimental images, and visualizations across domains. Early steps in this direction appear in systems like \path{Gemini/Deep Research} \cite{deep_research_now_available_gemini}, which incorporates basic chart extraction capabilities. Frameworks such as ChartCitor \cite{goswami2025chartcitormultiagentframeworkfinegrained} provide fine-grained bounding box citations to enhance explainability for complex chart understanding, improving user trust and productivity. Specialized models like LHRS-Bot \cite{muhtar2024lhrsbotempoweringremotesensing} demonstrate sophisticated reasoning capabilities for remote sensing imagery by leveraging geographic information and multimodal learning. The development of large-scale, domain-specific multimodal datasets for areas like entomology~\cite{Insect_Foundation} and seafloor geology~\cite{nguyen2024seafloorailargescalevisionlanguagedataset} is crucial for training more capable models. More comprehensive approaches could enable sophisticated analysis of visual scientific communication. Implementation opportunities include developing specialized scientific visualization parsers that extract quantitative data from diverse chart types, implementing diagram understanding systems that interpret complex scientific illustrations across domains, and designing domain-specific visual analysis components optimized for field-specific imagery like medical scans or astronomical observations. These advances could dramatically expand information access beyond text-centric sources.

\textbf{Visual Evidence Integration}. Effective research increasingly requires integration of visual evidence alongside textual sources. Future systems could implement sophisticated multimodal reasoning that incorporates visual evidence within comprehensive analytical frameworks, enabling true multimodal research synthesis. Recent analyses have identified multi-modal integration as a key missing capability in current AI research systems \cite{unlocking_ai_researchers}, highlighting the critical importance of cross-modal reasoning for scientific applications. 
Early examples of this direction appear in systems like \path{Gemini/Deep Research} \cite{deep_research_now_available_gemini}, which provides basic integration of image-derived information. More sophisticated approaches could enable balanced evidence integration across modalities.

Research directions include developing evidence alignment techniques
that match textual and visual information addressing common questions,
implementing cross-modal consistency verification that identifies
conflicts between textual claims and visual evidence, and designing
multimodal synthesis mechanisms that generate integrated understanding
across information types. These advances could enhance research quality
for domains with significant visual information components.

\subsubsection{Multimodal Source
Analysis}\label{multimodal-source-analysis}

Comprehensive understanding requires integrated analysis across diverse
information formats:

\textbf{Video Content Processing}. Video represents an increasingly
important but currently underutilized information source. Future
research could develop specialized video understanding components that
extract and interpret temporal visual information, including
presentations, interviews, demonstrations, and dynamic processes.
Initial steps in this direction are emerging in systems like OpenAI's
DALL-E 3, though not yet integrated into Deep Research workflows.
Comprehensive integration could enable access to the extensive knowledge
embedded in video content.

Implementation opportunities include developing lecture understanding
systems that extract structured knowledge from educational videos,
implementing process analysis components that interpret demonstrations
and procedures, and designing integrated audio-visual analysis that
combines visual information with spoken content for comprehensive
understanding. These advances could expand information access to the
rapidly growing corpus of video knowledge.

\textbf{Audio Content Integration}. Spoken information in podcasts,
lectures, interviews, and discussions represents a valuable knowledge
source. Future systems could incorporate sophisticated audio processing
that extracts, interprets, and integrates spoken information within
research workflows. Early examples of speech processing appear in
transcription services, but comprehensive research integration remains
limited. Advanced approaches could enable seamless incorporation of
spoken knowledge alongside traditional text sources.

Research directions include developing speaker identification and
attribution systems that maintain appropriate source tracking for spoken
content, implementing domain-specific terminology extraction that
accurately captures specialized vocabulary in varied acoustic
conditions, and designing temporal alignment techniques that connect
spoken information with related textual or visual content. These
advances could expand information access while maintaining appropriate
attribution and context.

\subsubsection{Cross-Modal Reasoning
Techniques}\label{cross-modal-reasoning-techniques}

Effective multimodal research requires specialized reasoning approaches
across information types:

\textbf{Multi-Modal Chain of Thought Reasoning}. Current reasoning processes typically operate primarily within single modalities despite handling diverse information types. Future systems could implement true multi-modal reasoning chains that explicitly incorporate diverse information types throughout the analytical process, not just in final outputs. Early steps appear in systems like \path{Gemini/Deep Research} \cite{deep_research_now_available_gemini}, which demonstrates basic visual incorporation in reasoning steps. More sophisticated approaches could enable reasoning flows that seamlessly transition between textual analysis, visual processing, numerical computation, and spatial reasoning based on task requirements.

Research opportunities include developing explicit multi-modal reasoning protocols that formalize information transfer between modalities, implementing cross-modal verification techniques that leverage complementary information types throughout reasoning chains, and designing unified representation frameworks that enable coherent reasoning across diverse information formats. These advances could significantly enhance reasoning quality for complex research tasks requiring integrated understanding across modalities, moving beyond the current text-centric reasoning paradigms to more human-like analytical processes that naturally leverage the most appropriate modality for each reasoning component.

\textbf{Cross-Modal Consistency Verification}. Integrating diverse
information modalities introduces new consistency challenges. Future
research could develop specialized verification mechanisms that assess
consistency across textual, visual, numerical, and temporal information,
enhancing overall reliability. Early steps in this direction appear in
systems like \path{Gemini/Deep Research} \cite{deep_research_now_available_gemini}, which implements basic cross-format
validation. More sophisticated approaches could enable reliable
integration of increasingly diverse information types.

Implementation opportunities include developing cross-modal
contradiction detection algorithms that identify conflicts between
information expressed in different formats, implementing uncertainty
alignment techniques that reconcile confidence estimates across
modalities, and designing multimodal fact verification systems that
leverage complementary evidence types for enhanced reliability. These
advances could address emerging challenges in multimodal information
integration.

\textbf{Multimodal Explanation Generation}. Effective communication
often requires coordinated explanation across modalities. Future systems
could generate truly multimodal research outputs that combine textual,
visual, and interactive components to enhance understanding and
persuasiveness. Early examples of this direction appear in systems like
\path{mshumer/OpenDeepResearcher} \cite{mshumer2024}, which implements basic report visualization.
More comprehensive approaches could enable sophisticated multimodal
communication tailored to content requirements.

Research directions include developing coordinated generation
architectures that produce aligned content across modalities,
implementing adaptive format selection algorithms that identify optimal
representation formats for different content types, and designing
multimodal narrative structures that effectively combine diverse formats
within coherent explanatory frameworks. These advances could enhance
communication effectiveness across application domains.

\subsection{Domain-Specific Optimization}\label{domain-specific-optimization}

Tailored enhancement for particular fields offers significant
performance improvements for specialized applications.

\subsubsection{Scientific Domain Adaptation}\label{scientific-domain-adaptation}

Scientific research presents unique requirements and opportunities for
specialization:

\textbf{Field-Specific Model Adaptation}. Current systems employ
relatively general architectures across scientific domains. Future
research could develop specialized adaptation techniques that optimize
performance for particular scientific fields including physics,
chemistry, biology, and others with distinct knowledge structures and
reasoning patterns. Early steps in this direction appear in systems like
\path{AutoGLM-Research} \cite{autoglm_research2025}, which implements domain-specific prompting. Domain-specialized research agents have demonstrated particular promise in physics~\cite{UGPhysics}, chemistry~\cite{bran2023chemcrowaugmentinglargelanguagemodels,chen2024autonomouslargelanguagemodel,alampara2025probinglimitationsmultimodallanguage,Zhao_2024}, materials science~\cite{ni2024matpilotllmenabledaimaterials}, oceanography~\cite{bi2024oceangptlargelanguagemodel}, geospatial analysis~\cite{manvi2024geollmextractinggeospatialknowledge}, patent research~\cite{Ren_2025, wang2024evopatmultillmbasedpatentssummarization}, and broader scientific discovery workflows~\cite{ghafarollahi2024sciagentsautomatingscientificdiscovery}. These specialized implementations highlight the value of domain adaptation beyond general research capabilities. More comprehensive adaptation could enable significant performance improvements for scientific applications.

Implementation approaches might include domain-specific fine-tuning
regimes that emphasize field-relevant reasoning patterns, specialized
architectural modifications that enhance performance for
domain-characteristic tasks, or hybrid systems that incorporate symbolic
components for domain-specific formal reasoning. These approaches could
address current limitations in scientific reasoning while maintaining
general capabilities for cross-domain research.

\textbf{Scientific Workflow Integration}. Effective scientific
application requires integration with existing research methodologies
and tools. Future systems could implement specialized interfaces for
scientific workflows including experimental design, data analysis,
literature integration, and theory development. Early examples of this
direction appear in systems like \path{n8n} \cite{n8n2024}, which provides workflow automation
for data processing. Platforms designed to support machine learning development in fundamental science also illustrate this trend, enabling research in federated cloud environments~\cite{Supporting_the_development_of_Machine_Learning}. More comprehensive integration could enable
seamless incorporation within scientific research processes. Research assistant tools employing prompt-based templates demonstrate domain-agnostic support for tasks such as enhanced literature search queries and preliminary peer review, facilitating standardized assistance across diverse scientific fields \cite{shamsabadi2024fairfreepromptbasedresearch}. User studies highlight varying automation needs across DS/ML workflows, suggesting targeted rather than complete end-to-end automation aligns with researcher preferences \cite{wang2021automationdoesdatascientist}. Research opportunities include developing experimental design assistants
that generate and refine research protocols based on literature and
objectives, implementing integrated analysis pipelines that combine
automated and human analytical components, and designing theory
development frameworks that link empirical findings with formal
theoretical structures. These advances could enhance practical
scientific impact beyond general information access \cite{EAIRA,Enabling_AI_Scientists_to_Recognize_Innovation}.

\subsubsection{Legal and Regulatory Domain
Specialization}\label{legal-and-regulatory-domain-specialization}

Legal applications present distinct challenges requiring specialized
adaptation:

\textbf{Legal Reasoning Enhancement}. Current systems struggle with the
precision and structure of legal analysis. Future research could develop
specialized legal reasoning components that incorporate case-based
reasoning, statutory interpretation, and doctrinal analysis within
coherent legal frameworks. Early steps in this direction appear in
systems like \path{OpenAI/Deep Research} \cite{openai2025}, which incorporates basic legal
language handling. More comprehensive specialization could enable
sophisticated legal applications across practice areas.

Implementation opportunities include developing case analysis systems
that extract and apply relevant precedent principles, implementing
statutory interpretation frameworks that apply established analytical
methodologies to legislative text, and designing multi-jurisdictional
reasoning approaches that navigate conflicts of law across legal
boundaries. These advances could enhance practical utility for legal
research and analysis applications.

\textbf{Regulatory Compliance Specialization}. Compliance applications
require comprehensive coverage with exceptional precision. Future
systems could implement specialized compliance components that ensure
complete regulatory coverage, systematic obligation identification, and
reliable guidance across complex regulatory landscapes. Early examples
of this direction appear in general information retrieval, but true
compliance optimization remains limited. Advanced approaches could
enable reliable automation of currently labor-intensive compliance
processes.

Research directions include developing regulatory change tracking
systems that monitor and interpret evolving requirements, implementing
obligation extraction techniques that identify and classify compliance
requirements across regulatory texts, and designing responsibility
mapping approaches that connect regulatory obligations with
organizational functions and processes. These advances could enhance
practical utility for compliance-intensive industries facing complex
regulatory environments.

\subsubsection{Medical and Healthcare Research
Support}\label{medical-and-healthcare-research-support}

Healthcare applications present unique requirements and ethical
considerations:

\textbf{Clinical Evidence Synthesis}. 
Medical applications require exceptional precision and comprehensive evidence integration. Future research could develop specialized medical components that synthesize clinical evidence across studies, guidelines, and practice observations while maintaining rigorous evaluation standards. Recent efforts such as Google's co-scientist project \cite{co-scientist} demonstrate the potential for AI to assist in scientific research including medical domains. Early steps in this direction appear in systems like \path{Perplexity/Deep Research} \cite{perplexity2025}, which implements enhanced citation for medical claims.
More comprehensive specialization could enable reliable clinical decision support.

Implementation approaches might include evidence grading systems that
apply established frameworks like GRADE \cite{Grade} to clinical research,
meta-analysis components that systematically integrate quantitative
findings across studies, and guideline alignment techniques that map
evidence to established clinical recommendations. These advances could
enhance practical utility for evidence-based medicine while maintaining
appropriate caution for this high-stakes domain.

\textbf{Patient-Specific Research Adaptation}. Personalized medicine
requires adapting general knowledge to individual patient contexts.
Future systems could implement specialized personalization components
that adapt research findings based on patient characteristics,
comorbidities, preferences, and other individual factors. Early examples
of this direction appear in basic filtering of contraindications, but
comprehensive personalization remains limited. Advanced approaches could
enable truly personalized evidence synthesis for clinical applications.

Research opportunities include developing comorbidity reasoning systems
that adjust recommendations based on condition interactions,
implementing preference integration frameworks that incorporate patient
values in evidence synthesis, and designing personalized risk-benefit
analysis approaches that quantify individual trade-offs for treatment
options. These advances could enhance clinical utility while respecting
the complexity of individual patient contexts.

\subsection{Human-AI Collaboration and Standardization}\label{human-ai-collaboration-and-standardization}

Enhancing human-AI partnership and establishing common standards
represent crucial directions for practical research impact and ecosystem
development.

\subsubsection{Interactive Research Workflows}\label{interactive-research-workflows}

Effective collaboration requires sophisticated interaction throughout the research process:

\textbf{Adaptive Query Refinement}. 
Current systems offer limited interaction during query formulation and refinement. Future research could develop sophisticated refinement interfaces that collaboratively develop research questions through iterative clarification, expansion, and focusing based on initial results and user feedback. Early steps in this direction appear in systems like \path{HKUDS/Auto-Deep-Research} \cite{hkuds2024}, which implements basic clarification dialogues, and benchmarks such as QuestBench \cite{QuestBench}, which evaluates AI systems' ability to identify missing information and formulate appropriate clarification questions in underspecified reasoning tasks. More comprehensive approaches could enable truly collaborative question development. Frameworks like \path{AutoAgent} \cite{autoagent} demonstrate how zero-code interfaces can enable non-technical users to effectively guide deep research processes through intuitive interaction patterns, while other systems are exploring methods that go beyond standard retrieval-augmented generation to better handle question identification in real-time conversations~\cite{agrawal2024beyondragquestionidentificationanswer}. Implementation opportunities include developing intent clarification systems that identify potential ambiguities and alternatives in research questions, implementing scope adjustment interfaces that dynamically expand or narrow research focus based on initial findings, and designing perspective diversification tools that suggest alternative viewpoints relevant to research objectives. These advances could enhance research quality by improving question formulation through human-AI collaboration.

\textbf{Interactive Exploration Interfaces}. Current systems typically present relatively static research outputs. Future research could develop sophisticated exploration interfaces that enable dynamic navigation, drilling down, and expansion across research findings based on evolving interests. Early examples of this direction appear in systems like \path{OpenManus} \cite{openmanus2025}, which provides basic exploration capabilities. Advanced approaches could enable truly interactive research experiences tailored to discovery patterns.

Research directions include developing information visualization
techniques specifically designed for research navigation, implementing adaptive detail management that expands or collapses content areas based on user interest signals, and designing seamless source transition mechanisms that enable smooth movement between synthesis and original sources. These advances could enhance discovery by enabling more exploratory and serendipitous research experiences.

\subsubsection{Expertise Augmentation Models}\label{expertise-augmentation-models}

Effective augmentation requires adaptation to user expertise and
objectives:

\textbf{Expertise-Adaptive Interaction}. Current systems offer limited
adaptation to user knowledge levels and expertise. Future research could
develop sophisticated adaptation mechanisms that tailor research
approaches, explanations, and outputs based on user domain knowledge and
research sophistication. Early steps in this direction appear in systems
like \path{Perplexity/Deep Research} \cite{perplexity2025}, which implements basic terminology
adjustment. More comprehensive adaptation could enable truly
personalized research assistance aligned with individual expertise.

Implementation approaches might include expertise inference systems that
dynamically assess user knowledge through interaction patterns,
explanation adaptation mechanisms that adjust detail and terminology
based on expertise models, and knowledge gap identification tools that
highlight potentially unfamiliar concepts within research contexts. Furthermore, mechanisms that learn to strategically request expert assistance when encountering gaps exceeding autonomous capability - as formalized in the Learning to Yield and Request Control (YRC) coordination problem \cite{danesh2025learningcoordinateexperts} - are crucial for optimizing intervention timing and resolution effectiveness. These advances could enhance research effectiveness across diverse user
populations with varying domain familiarity.

\textbf{Complementary Capability Design}. Optimal augmentation leverages complementary human and AI strengths. Future systems could implement specialized interfaces designed around capability complementarity, emphasizing AI contributions in information processing while prioritizing human judgment for subjective evaluation and contextual understanding. Early examples of this direction appear in systems like \path{Agent-RL/ReSearch} \cite{agentrl2024}, which implements basic division of analytical responsibilities. More sophisticated approaches could enable truly synergistic human-AI research partnerships.

Research opportunities include developing explanation components
specifically designed to facilitate human judgment rather than replace it, implementing confidence signaling mechanisms that highlight areas particularly requiring human evaluation, and designing interactive critique frameworks that enable efficient human feedback on system reasoning. Feng Xiong et al. ~\cite{xiong2024aiempoweredhumanresearchintegrating} redefine the collaborative dynamics between human researchers and AI systems. These advances could enhance collaborative effectiveness by optimizing around natural capability distributions.

\subsubsection{Framework Standardization Efforts}\label{framework-standardization-efforts}

Common architectures enable modular development and component
interoperability:

\textbf{Component Interface Standardization}. Advanced implementations employ standardized interfaces between major system components. The \path{OpenAI/AgentsSDK} \cite{openaiagents2025} defines explicit interface standards for agent components, enabling modular development and component substitution. Emerging industry standards like Anthropic's Model Context Protocol (MCP) \cite{mcp} provide standardized interaction frameworks for large language models and tools, enabling consistent integration patterns across implementations. Similarly, Google's Agent2Agent Protocol (A2A) \cite{A2A, A2A_a_new_era_of_agent_interoperability} establishes standardized communication patterns between autonomous agents, facilitating reliable multi-agent coordination. Open-source alternatives like \path{smolagents/open_deep_research} \cite{smolagents2024} implement comparable messaging protocols between agent components, highlighting industry convergence toward standardized interaction patterns.
Projects like \path{Open_deep_search} \cite{open_deep_research} further demonstrate how standardized protocols enable effective collaboration between specialized research agents. Integration of diverse API interactions, as explored in \path{Toolllm} \cite{toolllm}, provides additional standardization opportunities for managing external tool usage within research workflows.

\textbf{Evaluation Metric Standardization}. Current evaluation practices
vary widely across implementations. Future research could establish
standardized evaluation frameworks that enable consistent assessment and
comparison across systems and components. Early examples of this
direction appear in benchmarks like HLE \cite{HLE} and MMLU \cite{MMLU}, but comprehensive
standardization remains limited. Advanced standardization could enable
more efficient development through reliable quality signals and clear
improvement metrics.

Research opportunities include developing standardized benchmark suites
targeting specific research capabilities, implementing common evaluation
methodologies across research domains and applications, and designing
multi-dimensional assessment frameworks that provide nuanced performance
profiles beyond simple accuracy metrics. These advances could enhance
ecosystem quality by establishing clear standards and highlighting
genuine improvements.

\subsubsection{Cross-Platform Research Protocols}\label{cross-platform-research-protocols}

Interoperability across diverse systems enhances collective
capabilities:

\textbf{Research Result Exchange Formats}. Current systems typically
produce outputs in incompatible formats. Future research could develop
standardized exchange formats that enable seamless sharing of research
results across platforms and systems, enhancing collective capabilities.
Early steps in this direction appear in basic document formats, but true
research-specific standardization remains limited. Comprehensive
standardization could enable research workflows spanning multiple
specialized systems.

Implementation opportunities include defining standard structures for
research findings with appropriate attribution and confidence metadata,
establishing common formats for evidence representation across systems,
and developing shared schemas for research questions and objectives to
enable distributed processing. These advances could enhance capability
through specialization and complementary system utilization.

\textbf{Distributed Research Coordination}. Advanced interoperability
enables coordinated research across systems with complementary
capabilities. Future research could develop sophisticated coordination
frameworks that enable multi-system research workflows with appropriate
task allocation, result integration, and process management. Early
examples of this direction appear in workflows like those enabled by
\path{n8n} \cite{n8n2024}, but comprehensive research-specific coordination remains limited.
Advanced approaches could enable truly distributed research ecosystems
with specialized components addressing distinct process elements.

Research directions include developing distributed search coordination
protocols that efficiently leverage specialized search capabilities,
implementing cross-system result verification techniques that ensure
consistency across distributed findings, and designing efficient
coordination protocols that minimize communication overhead in
distributed research workflows. These advances could enhance collective
capability through specialization and parallelization across the
ecosystem.

\subsubsection{Joint Human-AI Knowledge Creation}\label{joint-human-ai-knowledge-creation}

Moving beyond information retrieval to collaborative insight generation:

\textbf{Collaborative Creation Environments}. Advanced collaboration
requires sophisticated content co-creation capabilities. Future research
could develop specialized collaborative environments that enable fluid
transition between human and AI contributions within unified document
development. Early steps in this direction appear in systems like
\path{mshumer/OpenDeepResearcher}, which implements basic collaborative
document generation.
Advanced interfaces like those explored in Self-Explanation in Social AI Agents \cite{Self-Explanation_in_Social_AI_Agents} demonstrate how explanation capabilities can enhance collaborative research through more transparent reasoning processes. Similarly, innovative interaction paradigms like AI-Instruments \cite{ai_instruments} show how prompts can be embodied as instruments to abstract and reflect commands as general-purpose tools, suggesting novel approaches to research interface design that enhance collaborative capabilities through intuitive interaction patterns. Approaches where AI agents learn to assist other agents by observing them also show promise for developing more effective collaborative behaviors~\cite{keurulainen2021learningassistagentsobserving}. Effidit demonstrates comprehensive writing support through multifunctional capabilities including text polishing and context-aware phrase refinement, extending collaborative editing beyond basic generation \cite{shi2022effiditaiwritingassistant}. More comprehensive approaches could enable truly integrated co-creation experiences. 

Implementation opportunities include developing section suggestion
systems that propose potential content expansions based on document
context, implementing stylistic adaptation mechanisms that align
AI-generated content with established document voice and approach, and incorporating implicit feedback mechanisms that interpret rejected suggestions as negative signals to refine outputs while preserving original intent \cite{towle2024enhancingaiassistedwriting}, and
designing seamless revision interfaces that enable efficient editing
across human and AI contributions, like iterative human-AI co-editing as demonstrated by REVISE~\cite{xie2023interactiveeditingtextsummarization} -- a framework allowing writers to dynamically modify summary segments through fill-in-the-middle generation. These advances could enhance
collaborative productivity by reducing friction in joint content
development \cite{Huq2025}.

\textbf{Mixed-Initiative Research Design}. Sophisticated collaboration
includes shared determination of research direction and approach. Future
systems could implement mixed-initiative frameworks that dynamically
balance direction setting between human preferences and AI-identified
opportunities throughout the research process. Early examples of this
direction appear in systems like \path{smolagents/open_deep_research} \cite{smolagents2024}, which
implements basic suggestion mechanisms. Advanced approaches could enable
truly collaborative research planning with balanced initiative
distribution.

Research directions include developing opportunity identification
systems that highlight promising but unexplored research directions,
implementing trade-off visualization techniques that communicate
potential research path alternatives and implications, and designing
preference elicitation frameworks that efficiently capture evolving
research priorities throughout the process, and integrating explainable reward function mechanisms to enhance human understanding of AI's decision logic, thereby improving collaborative efficiency in value alignment contexts \cite{sanneman2021explainingrewardfunctionshumans}. These advances could enhance
discovery by combining human insight with AI-identified opportunities in
balanced partnerships.

The future research directions outlined in this section highlight both
the significant potential for advancement and the multi-faceted nature
of Deep Research development. Progress will likely emerge through
complementary advances across reasoning architectures, multimodal
capabilities, domain specialization, human-AI collaboration, and
ecosystem standardization. While commercial implementations like \path{OpenAI/Deep Research} \cite{openai2025}, \path{Gemini/Deep Research} \cite{deep_research_now_available_gemini}, and \path{Perplexity/Deep Research} \cite{perplexity2025} will
undoubtedly drive significant innovation, open-source alternatives and
academic research will play crucial roles in expanding the boundaries of
what's possible and ensuring broad participation in this rapidly
evolving field.

\section{Conclusion}

This survey has examined the rapidly evolving domain of Deep Research systems, tracing their development from initial implementations in 2023 through the sophisticated ecosystem emerging in 2025. Through comprehensive analysis of commercial offerings like \path{OpenAI/Deep Research} \cite{openai2025}, \path{Gemini/Deep Research} \cite{deep_research_now_available_gemini}, and \path{Perplexity/Deep Research} \cite{perplexity2025}, alongside open-source alternatives including \path{HKUDS/Auto-Deep-Research} \cite{hkuds2024}, \path{dzhng/deep-research} \cite{dzhng2024}, and numerous others, we have identified key technical patterns, implementation approaches, and application opportunities that characterize this transformative technology domain.

\subsection{Key Findings and Contributions}

Our analysis reveals several fundamental insights about the current state and trajectory of Deep Research systems:

\paragraph{Technical Architecture Patterns}
Effective Deep Research implementations demonstrate consistent architectural patterns across foundation models, environmental interaction, task planning, and knowledge synthesis dimensions. Commercial implementations like \path{OpenAI/Deep Research} \cite{openai2025} and \path{Gemini/Deep Research} \cite{deep_research_now_available_gemini} typically leverage proprietary foundation models with extensive context lengths and sophisticated reasoning capabilities, while open-source alternatives like \path{Camel-AI/OWL} \cite{camel2025} and \path{QwenLM/Qwen-Agent} \cite{qwen2025} demonstrate how effective research capabilities can be achieved with more accessible models through specialized optimization.

Environmental interaction capabilities show greater diversity, with specialized tools like \path{Nanobrowser} \cite{nanobrowser2024} and \path{dzhng/deep-research} \cite{dzhng2024} demonstrating exceptional effectiveness in web navigation and content extraction, while comprehensive platforms like \path{Manus} \cite{manus2025} and \path{AutoGLM-Search} \cite{autoglm_research2025} offer broader interaction capabilities across multiple environments. These patterns highlight both the value of specialization and the importance of comprehensive environmental access for effective research.

Task planning and execution approaches reveal similar diversity, with frameworks like \path{OpenAI/AgentsSDK} \cite{openaiagents2025} and \path{Flowith/OracleMode} \cite{flowith2025} providing sophisticated planning capabilities, while systems like \path{Agent-RL/ReSearch} \cite{agentrl2024} and \path{smolagents/open_deep_research} \cite{smolagents2024} emphasize execution reliability and collaborative approaches respectively. Knowledge synthesis capabilities demonstrate consistent emphasis on information evaluation, though with varied approaches to presentation and interactivity across implementations like \path{HKUDS/Auto-Deep-Research} \cite{hkuds2024} and \path{mshumer/OpenDeepResearcher} \cite{mshumer2024}.

\paragraph{Implementation Approach Distinctions}
Our analysis highlights meaningful distinctions between commercial and open-source implementation approaches. Commercial platforms typically offer optimized performance, sophisticated interfaces, and comprehensive capabilities, though with associated costs and customization limitations. Systems like \path{OpenAI/Deep Research} \cite{openai2025} and \path{Perplexity/Deep Research} \cite{perplexity2025} demonstrate exceptional performance on standard benchmarks, though with significant variation in application focus and interaction models.

Open-source implementations demonstrate greater architectural diversity and customization flexibility, though typically with increased deployment complexity and more limited performance on standard benchmarks. Projects like \path{dzhng/deep-research} \cite{dzhng2024}, \path{nickscamara/open-deep-research} \cite{nickscamara2024}, and \path{HKUDS/Auto-Deep-Research} \cite{hkuds2024} offer complete research pipelines with varied architectural approaches, while specialized components like \path{Jina-AI/node-DeepResearch} \cite{jina2025} and \path{Nanobrowser} \cite{nanobrowser2024} enable customized workflows addressing specific requirements. Frameworks such as AutoChain \cite{AutoChain} provide lightweight tools to simplify the creation and evaluation of custom generative agents, enabling rapid iteration for specialized applications.

These distinctions highlight complementary roles within the ecosystem, with commercial implementations offering accessibility and performance for general users, while open-source alternatives enable customization, control, and potentially lower operational costs for specialized applications and high-volume usage. This diversity enhances overall ecosystem health through competition, specialization, and diverse innovation paths.

\paragraph{Application Domain Adaptations}
Our examination of application patterns reveals meaningful adaptations across domains including academic research\cite{jafari2024streamliningselectionphasesystematic,tu2024augmentingauthorexploringpotential,turobov2024usingchatgptthematicanalysis}, scientific discovery\cite{liu2024aigsgeneratingscienceaipowered,qi2024metascientisthumanaisynergisticframework,sourati2021acceleratingsciencehumanversus,su2025headsbetteroneimproved,taniguchi2024collectivepredictivecodingmodel,tiukova2024genesisautomationsystemsbiology,yin2024turingtestsaiscientist,yoon2025knowledgesynthesisphotosynthesisresearch,Aligning_AI_driven_discovery_with_human_intuition,The_impact_of_AI_and_peer_feedback_on_research_writing_skill,alampara2025probinglimitationsmultimodallanguage,ansari2023agentbasedlearningmaterialsdatasets,behandish2022airesearchassociateearlystage,chamoun2024automatedfocusedfeedbackgeneration,frança2023aiempoweringresearch10,gao2024empoweringbiomedicaldiscoveryai,gower2024useairoboticsystemsscientific,gu2025llmsrealizecombinatorialcreativity,hogan2024aiscivisionframeworkspecializinglarge,kramer2023automatedscientificdiscoveryequation,kuznetsov2024naturallanguageprocessingpeer,laverick2024multiagentcosmologicalparameteranalysis,markowitz2024complexityclarityaienhances,mathur2024visionmodularaiassistant}, business intelligence\cite{nguyen2024docmasterunifiedplatformannotation}, financial analysis, education\cite{Beyond_the_Hype,aryan2024llmsdebatepartnersutilizing,yuan2024boostingscientificconceptsunderstanding,qi2024knowledgecomponentbasedmethodologyevaluatingai}, and personal knowledge management\cite{lee2024ethicalpersonalaiapplications,coplu2024prompttimesymbolicknowledgecapture}. Academic applications exemplified by systems like \path{OpenAI/Deep Research} \cite{openai2025} and \path{Camel-AI/OWL} \cite{camel2025} demonstrate particular emphasis on comprehensive literature coverage, methodological understanding, and citation quality. Scientific implementations like \path{Gemini/Deep Research} \cite{deep_research_now_available_gemini} and \path{Agent-RL/ReSearch} \cite{agentrl2024} emphasize experimental design, data analysis, and theory development capabilities.

Business applications leveraging systems like \path{Manus} \cite{manus2025} and \path{n8n} \cite{n8n2024} show stronger focus on information currency, competitive analysis, and actionable insight generation. Educational implementations demonstrate adaptations for learning support, content development, and research skill training across systems like \path{Perplexity/Deep Research} \cite{perplexity2025} and \path{OpenManus} \cite{openmanus2025}. These patterns highlight how general deep research capabilities translate into domain value through specialized adaptation addressing field-specific requirements and workflows.

\paragraph{Ethical Consideration Approaches}
Our analysis reveals both common patterns and implementation diversity in addressing crucial ethical dimensions including information accuracy, privacy protection, intellectual property respect, and accessibility. Commercial implementations typically demonstrate sophisticated approaches to factual verification, with systems like \path{OpenAI/Deep Research} \cite{openai2025} and \path{Perplexity/Deep Research} \cite{perplexity2025} implementing multi-level verification and explicit attribution, while open-source alternatives like \path{grapeot/deep_research_agent} \cite{grapeot2024} and \path{HKUDS/Auto-Deep-Research} \cite{hkuds2024} demonstrate pragmatic approaches within more constrained technical environments.

Privacy protection shows similar patterns, with commercial systems implementing comprehensive safeguards appropriate to their cloud-based operation, while open-source alternatives like \path{OpenManus} \cite{openmanus2025} emphasize local deployment for sensitive applications. Attribution and intellectual property approaches demonstrate consistent emphasis on source transparency and appropriate utilization boundaries, though with varied implementation sophistication across the ecosystem.

These patterns highlight both shared ethical priorities across the ecosystem and implementation diversity reflecting different technical constraints, deployment models, and user requirements. This diversity represents a strength in addressing multi-faceted ethical challenges through complementary approaches and continuous innovation.

\subsection{Limitations and Outlook}

While this survey provides comprehensive analysis of current Deep Research systems and emerging trends, several limitations warrant acknowledgment:

\paragraph{Rapidly Evolving Landscape}
The accelerating pace of development in this domain presents inherent challenges for comprehensive analysis. New systems and capabilities continue to emerge, with commercial offerings like \path{OpenAI/Deep Research} \cite{openai2025}, \path{Gemini/Deep Research} \cite{deep_research_now_available_gemini}, and \path{Perplexity/Deep Research} \cite{perplexity2025} receiving frequent updates, while the open-source ecosystem continuously expands through new projects and enhancements to existing frameworks like \path{dzhng/deep-research} \cite{dzhng2024} and \path{HKUDS/Auto-Deep-Research} \cite{hkuds2024}.

This survey captures the state of the art as of early 2025, but both technical capabilities and implementation approaches will continue to evolve rapidly. The classification framework and analysis methodology provided here offer a structural foundation for continued assessment as the field progresses through subsequent development phases.

\paragraph{Implementation Detail Limitations}
Comprehensive technical analysis faces challenges due to limited implementation transparency, particularly for commercial systems. While open-source implementations like \path{nickscamara/open-deep-research} \cite{nickscamara2024} and \path{Agent-RL/ReSearch} \cite{agentrl2024} enable detailed architectural examination, commercial systems like \path{OpenAI/Deep Research} \cite{openai2025} and \path{Gemini/Deep Research} \cite{deep_research_now_available_gemini} reveal limited internal details, restricting comprehensive comparative analysis of certain technical dimensions.

Our approach addresses this limitation through behavioral analysis, publicly available documentation examination, and consistent evaluation across standardized benchmarks and qualitative assessment frameworks. These methods enable meaningful comparison despite transparency variations, though complete architectural analysis remains challenging for proprietary implementations.

\paragraph{Application Impact Assessment}
Evaluating real-world impact presents persistent challenges given the early deployment stage of many Deep Research systems. While initial applications demonstrate promising capabilities across domains including academic research\cite{wei2023academicgptempoweringacademicresearch,aytar2024retrievalaugmentedgenerationframeworkacademic,perkins2024generativeaitoolsacademic,ramirezmedina2025acceleratingscientificresearchmultillm}, business intelligence, and education\cite{Beyond_the_Hype,aryan2024llmsdebatepartnersutilizing,yuan2024boostingscientificconceptsunderstanding}, a comprehensive long-term impact assessment requires extended observation beyond the scope of this survey. Potential transformative effects on research methodologies, knowledge work, and information access patterns remain partially speculative despite encouraging early indications.

Future research should incorporate longitudinal analysis of deployment patterns, usage evolution, and organizational integration to assess realized impact beyond technical capabilities and early applications. Such analysis would complement the technical and architectural focus of the current survey with valuable perspectives on practical significance and societal implications.

\subsection{Broader Implications}

Beyond specific findings, this survey highlights several broader implications for the future of knowledge work and information access:

\paragraph{Research Methodology Transformation}
Deep Research systems demonstrate potential to fundamentally transform research methodologies across domains. The comprehensive information access, advanced reasoning capabilities, and efficient knowledge synthesis demonstrated by systems like \path{OpenAI/Deep Research} \cite{openai2025}, \path{Gemini/Deep Research} \cite{deep_research_now_available_gemini}, and their open-source alternatives suggest significant opportunities to accelerate discovery, enhance comprehensiveness, and enable novel cross-domain connections beyond traditional research approaches.

Rather than simply automating existing processes, these systems enable fundamentally new research approaches leveraging capabilities exceeding human information processing in scale while complementing human insight, creativity, and contextual understanding. This complementarity suggests evolution toward collaborative research models rather than replacement of human researchers, with significant potential for productivity enhancement and discovery acceleration. However, Ashktorab et al. \cite{ashktorab2024emergingreliancebehaviorshumanai} highlight that in human-AI collaboration, users may exhibit overreliance behaviors, appending AI-generated responses even when conflicting, which can compromise data quality.

\paragraph{Knowledge Access Democratization}
The emergence of accessible Deep Research implementations across commercial and open-source ecosystems demonstrates potential for broader knowledge democratization. Systems like \path{Perplexity/Deep Research} \cite{perplexity2025} with free access tiers and open-source alternatives like \path{nickscamara/open-deep-research} \cite{nickscamara2024} and \path{HKUDS/Auto-Deep-Research} \cite{hkuds2024} enable sophisticated research capabilities previously requiring specialized expertise and substantial resources, potentially reducing barriers to high-quality information access and analysis.

This democratization carries significant implications for education, entrepreneurship, civic participation, and individual knowledge development. While accessibility challenges remain, particularly regarding technical expertise requirements and computational resources, the overall trajectory suggests broadening access to advanced research capabilities with potential positive impacts on knowledge equity across society.

\paragraph{Collective Intelligence Enhancement}
Beyond individual applications, Deep Research systems demonstrate potential for collective intelligence enhancement through improved knowledge integration, insight sharing, and collaborative discovery. The capabilities demonstrated by systems like \path{Manus} \cite{manus2025}, \path{Flowith/OracleMode} \cite{flowith2025}, and \path{smolagents/open_deep_research} \cite{smolagents2024} suggest opportunities for enhanced knowledge synthesis across organizational and disciplinary boundaries, potentially addressing fragmentation challenges in increasingly complex knowledge domains.

Rather than viewing these systems as isolated tools, their integration into collaborative knowledge ecosystems highlights potential for systemic enhancement of collective sense-making, evidence-based decision making, and shared understanding development. This perspective emphasizes the social and organizational dimensions of Deep Research impact beyond technical capabilities and individual productivity enhancement.

\subsection{Final Thoughts}

The rapid emergence and evolution of Deep Research systems represent a significant advancement in the application of artificial intelligence to knowledge discovery and utilization. While technical implementations will continue to evolve and specific systems will emerge and recede, the fundamental capability shift enabled by these technologies appears likely to persist and expand.

The diverse ecosystem spanning commercial platforms like \path{OpenAI/Deep Research} \cite{openai2025}, \path{Gemini/Deep Research} \cite{deep_research_now_available_gemini}, and \path{Perplexity/Deep Research} \cite{perplexity2025}, alongside open-source alternatives like \path{dzhng/deep-research} \cite{dzhng2024}, \path{HKUDS/Auto-Deep-Research} \cite{hkuds2024}, and numerous specialized components, demonstrates innovation across multiple technical dimensions, implementation approaches, and application domains. This diversity enhances overall ecosystem health through competition, specialization, and complementary development trajectories.

As research continues across advanced reasoning architectures, multimodal capabilities, domain specialization, human-AI collaboration, and ecosystem standardization, we anticipate continued rapid advancement building on the foundation established by current implementations. This evolution will likely yield increasingly sophisticated research capabilities with significant implications for knowledge work across domains, potentially transforming how information is discovered, validated, synthesized, and utilized throughout society.

The responsible development of these powerful capabilities requires continued attention to ethical considerations including information accuracy, privacy protection, intellectual property respect, and accessibility. By addressing these considerations alongside technical advancement, the Deep Research ecosystem can fulfill its potential for positive impact on knowledge discovery and utilization while minimizing potential harms or misuse.

In conclusion, Deep Research represents both a fascinating technical domain for continued research and a potentially transformative capability for practical knowledge work across society. The frameworks, analysis, and directions presented in this survey provide a foundation for continued examination of this rapidly evolving field with significant implications for the future of information access, knowledge synthesis, and discovery processes.

\bibliographystyle{ACM-Reference-Format}

\end{document}